\documentclass[10pt,twocolumn,letterpaper]{article}

\usepackage{iccv}
\usepackage{times}
\usepackage{epsfig}
\usepackage{graphicx}
\usepackage{amsmath}
\usepackage{amssymb}
\usepackage{graphicx}
\usepackage{amsmath}
\usepackage{amssymb}
\usepackage{booktabs}
\usepackage{duckuments}
\usepackage{algorithm}
\usepackage{algpseudocode}
\usepackage{amsmath}
\usepackage{amsfonts}
\usepackage{amssymb}
\usepackage{mathtools}
\usepackage{float}
\usepackage[10pt]{moresize}
\DeclareMathOperator*{\argmin}{arg\,min}

\usepackage{color}
\usepackage{epsfig}
\usepackage{graphicx}
\usepackage{times}
\usepackage{algorithm}
\usepackage{comment}

\usepackage{subcaption}
\usepackage{booktabs}

\usepackage{pifont}

\usepackage{microtype}
\usepackage[font=smaller]{caption}
\usepackage{arydshln}
\usepackage{tabularx}
\usepackage{adjustbox}
\usepackage{array}
\usepackage{booktabs}
\usepackage{colortbl}
\usepackage{float,wrapfig}
\usepackage{makecell}
\usepackage{hhline}
\usepackage{multirow}
\usepackage{subcaption} %
\usepackage[percent]{overpic}
\usepackage{graphbox}
\usepackage{diagbox}
\usepackage{stmaryrd}

\usepackage[dvipsnames]{xcolor}

\usepackage{amsmath,amsfonts,amssymb}
\usepackage{bm}
\usepackage{nicefrac}
\usepackage{microtype}
\usepackage{dsfont}
\usepackage{changepage}
\usepackage{extramarks}
\usepackage{fancyhdr}
\usepackage{setspace}
\usepackage{soul}
\usepackage{xspace}
\usepackage{hhline}

\usepackage[pagebackref,breaklinks,colorlinks]{hyperref}
\usepackage{url}

\usepackage{enumitem}

\iccvfinalcopy 
\def\iccvPaperID{11321} 

\ificcvfinal\fi

\begin{document}

\title{DataDAM: Efficient Dataset Distillation with Attention Matching}

\author{Ahmad Sajedi$^{1}$\thanks{Equal contribution}, ~Samir Khaki$^{1*}$, ~Ehsan Amjadian$^{2, 3}$, ~Lucy Z. Liu$^{2}$, ~Yuri A. Lawryshyn$^{1}$,\\ and ~Konstantinos N. Plataniotis$^{1}$\\
	 $^{1}$University of Toronto  ~~~~~~~ $^{2}$Royal Bank of Canada (RBC) ~~~~~~~ $^{3}$University of Waterloo\\
	{\tt\small \{ahmad.sajedi,samir.khaki\}@mail.utoronto.ca}\\
\tt\small Code: \href{https://github.com/DataDistillation/DataDAM}{https://github.com/DataDistillation/DataDAM}
}

\maketitle
        \ificcvfinal\thispagestyle{empty}\fi

\begin{abstract}

\end{abstract}

Researchers have long tried to minimize training costs in deep learning while maintaining strong generalization across diverse datasets. Emerging research on dataset distillation aims to reduce training costs by creating a small synthetic set that contains the information of a larger real dataset and ultimately achieves test accuracy equivalent to a model trained on the whole dataset. Unfortunately, the synthetic data generated by previous methods are not guaranteed to distribute and discriminate as well as the original training data, and they incur significant computational costs. Despite promising results, there still exists a significant performance gap between models trained on condensed synthetic sets and those trained on the whole dataset. In this paper, we address these challenges using efficient Dataset Distillation with Attention Matching (DataDAM), achieving state-of-the-art performance while reducing training costs. Specifically, we learn synthetic images by matching the spatial attention maps of real and synthetic data generated by different layers within a family of randomly initialized neural networks. Our method outperforms the prior methods on several datasets, including CIFAR10/100, TinyImageNet, ImageNet-1K, and subsets of ImageNet-1K     across most of the settings, and achieves improvements of up to 6.5\% and 4.1\% on CIFAR100 and ImageNet-1K, respectively. We also show that our high-quality distilled images have practical benefits for downstream applications, such as continual learning and neural architecture search.
\section{Introduction}


Deep learning has been highly successful in various fields, including computer vision and natural language processing, due to the use of large-scale datasets and modern Deep Neural Networks (DNNs) \cite{deng2009imagenet, he2016deep, dosovitskiyimage, kenton2019bert, sajedi2023end}. 
\begin{figure}[H]
    \centering
    \setlength{\abovecaptionskip}{0.1cm}
    \includegraphics[width=0.47\textwidth]{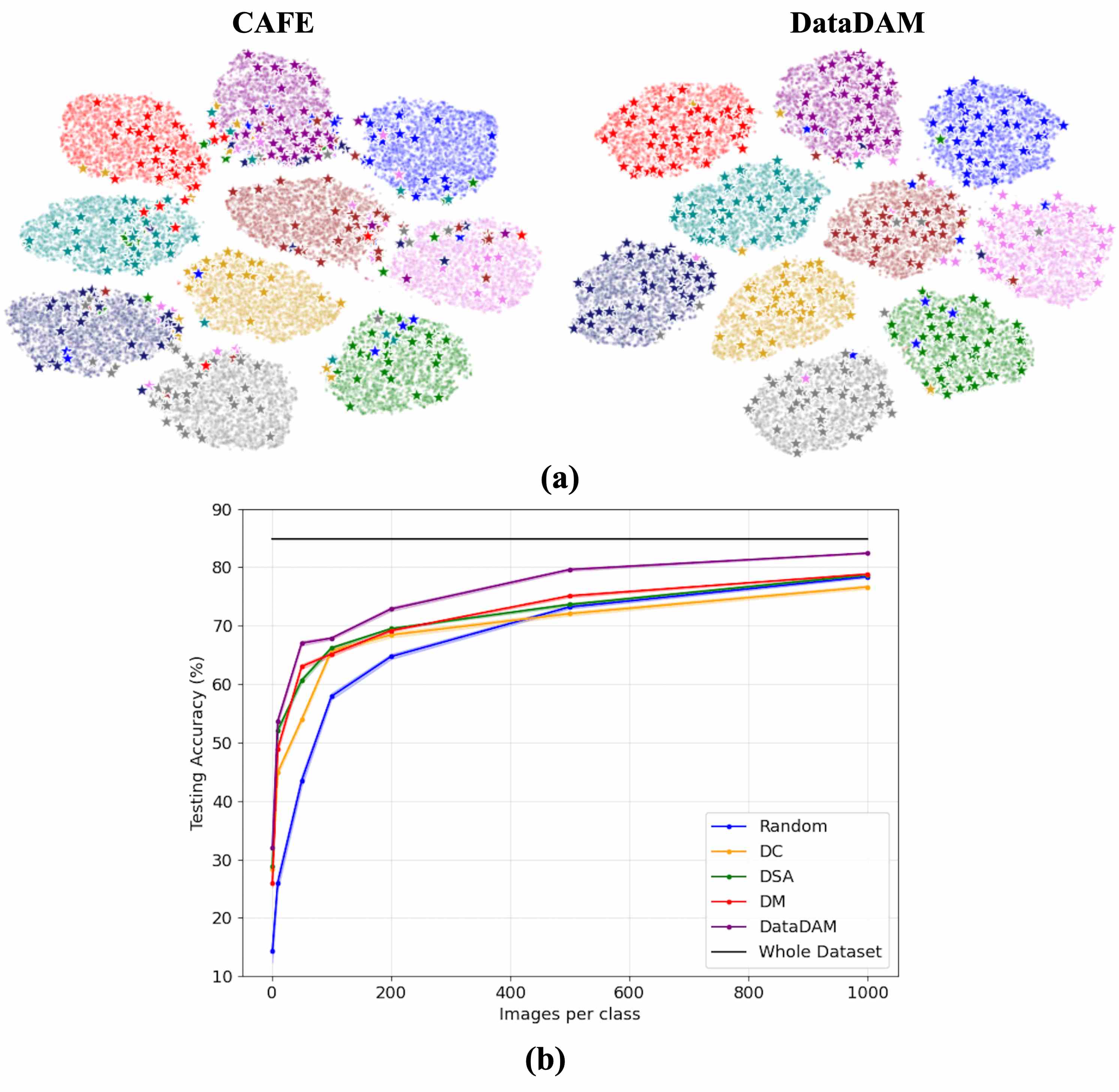}
    \caption{(a) Data distribution of the distilled images on the CIFAR10 dataset with 50 images per class (IPC50) for CAFE \cite{wang2022cafe} and DataDAM. (b) Performance comparison with state-of-the-art methods on the CIFAR10 dataset for varying IPCs.}
    \label{fig:my_label}
\end{figure}
However, extensive infrastructure resources for training, hyperparameter tuning, and architectural searches make it challenging to reduce computational costs while maintaining comparable performance. Two primary approaches to address this issue are model-centric and data-centric. Model-centric methods involve model compression techniques \cite{hinton2015distilling, wu2016quantized, amer2021high, khaki_9679989, yu2017compressing, sajedi2022subclass, Khaki_2023_CVPR}, while data-centric methods concentrate on constructing smaller datasets with enough information for training, which is the focus of this paper. A traditional data-centric approach is the coreset selection method, wherein we select a representative subset of an original dataset \cite{rebuffi2017icarl, castro2018end, belouadah2020scail, seneractive, tonevaempirical}; however, these methods have limitations as they rely on heuristics to generate a coarse approximation of the whole dataset, which may lead to a suboptimal solution for downstream tasks like image classification \cite{tonevaempirical, rebuffi2017icarl}. Dataset distillation (or condensation) \cite{wang2018dataset} is proposed as an alternative, which distills knowledge from a large training dataset into a smaller synthetic set such that a model trained on it achieves competitive testing performance with one trained on the real dataset. The condensed synthetic sets contain valuable information, making them a popular choice for various machine learning applications like continual learning \cite{wang2018dataset, zhao2021datasetDC, zhao2021datasetDSA}, neural architecture search \cite{cuidc, zhao2023dataset, zhao2021datasetDC}, federated learning \cite{xiong2022feddm, zhou2020distilled}, and privacy-preserving \cite{dong2022privacy, tsilivis2022can} tasks.

Dataset distillation was first proposed by Wang \etal \cite{wang2018dataset} where bi-level meta-learning was used to optimize model parameters on synthetic data in the inner loop and refine the data with meta-gradient updates to minimize the loss on the original data in the outer loop. Various methods have been proposed to overcome the computational expense of this method, including approximating the inner optimization with kernel methods \cite{bohdal2020flexible, nguyen2021dataset, nguyen2021dataset2, zhoudataset}, surrogate objectives like gradient matching \cite{zhao2021datasetDC, zhao2021datasetDSA, lee2022dataset}, trajectory matching \cite{cazenavette2022dataset}, and distribution matching \cite{wang2022cafe, zhao2023dataset}. The kernel-based methods and gradient matching work still require bi-level optimization and second-order derivation computation, making training a difficult task. Trajectory matching \cite{cazenavette2022dataset} demands significant GPU memory for extra disk storage and expert model training. CAFE \cite{wang2022cafe} uses dynamic bi-level optimization with layer-wise feature alignment, but it may generate biased images and incur a significant time cost (Figure \ref{fig:my_label}). Thus, these methods are not scalable for larger datasets such as ImageNet-1K \cite{deng2009imagenet}. Distribution matching (DM) \cite{zhao2023dataset} was proposed as a scalable solution for larger datasets by skipping optimization steps in the inner loop. However, DM usually underperforms compared to prior methods \cite{cazenavette2022dataset}.

In this paper, we propose a new framework called "\textbf{Data}set \textbf{D}istillation with \textbf{A}ttention \textbf{M}atching (DataDAM)" to overcome computational problems, achieve an unbiased representation of the real data distribution, and outperform the performance of the existing methods. Due to the effectiveness of randomly initialized networks in generating strong representations that establish a distance-preserving embedding of the data \cite{cao2018review, saxe2011random, giryes2016deep, zhao2023dataset}, we leverage multiple randomly initialized DNNs to extract meaningful representations from real and synthetic datasets. We align their most discriminative feature maps using the Spatial Attention Matching (SAM) module and minimize the distance between them with the MSE loss. We further reduce the last-layer feature distribution disparities between the two datasets with a complementary loss as a regularizer. Unlike existing methods \cite{zhao2021datasetDC,wang2022cafe,cazenavette2022dataset}, our approach does not rely on pre-trained network parameters or employ bi-level optimization, making it a promising tool for synthetic data generation. The generated synthetic dataset does not introduce any bias into the data distribution while outperforming concurrent methods, as shown in Figure \ref{fig:my_label}. \\
The contributions of our study are:

\textbf{[C1]}: We proposed an effective end-to-end dataset distillation method with attention matching and feature distribution alignment to closely approximate the distribution of the real dataset with low computational costs.

\textbf{[C2]}: Our method is evaluated on computer vision datasets with different resolutions, where it achieves state-of-the-art results across multiple benchmark settings. Our approach offers up to a 100x reduction in training costs while simultaneously enabling cross-architecture generalizations.

\textbf{[C3]}: Our distilled data can enhance downstream applications by improving memory efficiency for continual learning and accelerating neural architecture search through a more representative proxy dataset.
\section{Related Work}

\paragraph{Dataset Distillation.} \label{dd}
\begin{figure*}[h]
\centering
    {\includegraphics[width=0.7\textwidth]{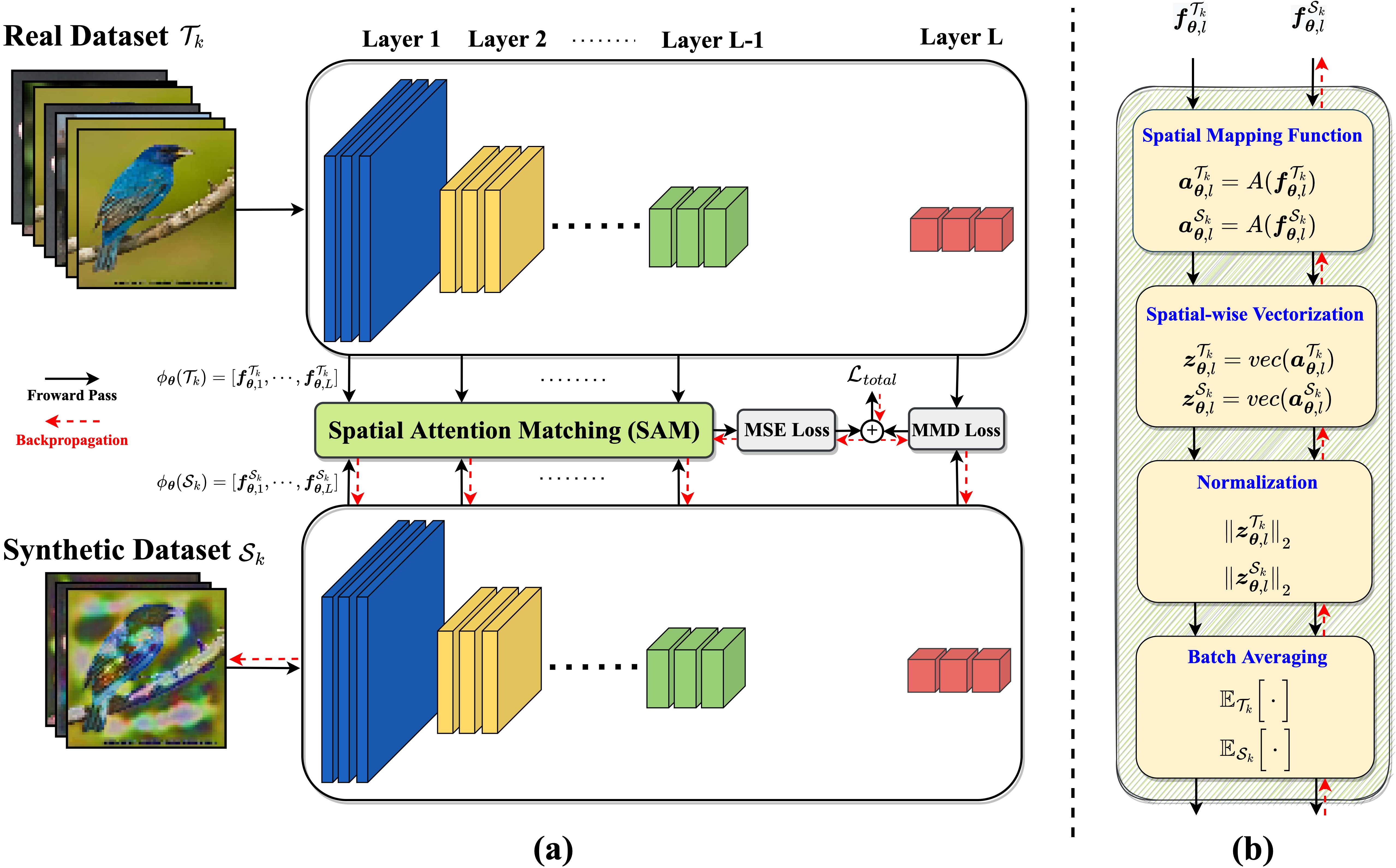}}
    \caption{(a) Illustration of the proposed DataDAM method. DataDAM includes a Spatial Attention Matching (SAM) module to capture the dataset's distribution and a complementary loss for matching the feature distributions in the last layer of the encoder network. (b) The internal architecture of the SAM module.}
    \label{fig:DataDAM}
\end{figure*}
Wang \etal \cite{wang2018dataset} first introduced dataset distillation by expressing network parameters as a function of synthetic data and optimizing the synthetic set to minimize the training loss on real training data. Later works extended this approach with soft labels \cite{bohdal2020flexible} and a generator network \cite{such2020generative}. Researchers have proposed simplifying the neural network model in bi-level optimization using kernel methods, such as ridge regression, which has a closed-form solution \cite{bohdal2020flexible, zhoudataset}, and a kernel ridge regression model with Neural Tangent Kernel \cite{lee2019wide} (NTK) that approximates the inner optimization \cite{nguyen2021dataset, nguyen2021dataset2}. Alternatively, some studies have utilized surrogate objectives to address unrolled optimization problems. Dataset condensation (DC) \cite{zhao2021datasetDC} and DCC \cite{lee2022dataset} generate synthetic images by matching the weight gradients of neural networks on real and distilled training datasets, while Zhao \etal \cite{zhao2021datasetDSA} improve gradient matching with data augmentation. MTT \cite{cazenavette2022dataset} matches model parameter trajectories trained with synthetic and real datasets, and CAFE \cite{zhao2023dataset} and DM \cite{wang2022cafe} match features generated by a model using distilled and real datasets. However, these methods have limitations, including bi-level optimization \cite{zhao2021datasetDC, zhao2021datasetDSA, wang2022cafe, lee2019wide}, second-order derivative computation \cite{zhao2021datasetDC}, generating biased examples \cite{zhao2021datasetDSA, wang2022cafe}, and massive GPU memory demands \cite{cazenavette2022dataset, zhoudataset}. In contrast, our approach matches the spatial attention map in intermediate layers, reducing memory costs while outperforming most existing methods on standard benchmarks.
\paragraph{Coreset Selection.} \label{coreset}
Coreset selection is another data-centric approach that chooses a representative subset of an original dataset using heuristic selection criteria. For example, random selection \cite{rebuffi2017icarl} selects samples randomly; Herding \cite{castro2018end, belouadah2020scail} selects the samples closest to the cluster center for each class center; K-Center \cite{seneractive} chooses multiple center points of a class to minimize the maximum distance between data points and their nearest center point; and \cite{tonevaempirical} identifies training samples that are easily forgotten during the training process. However, heuristics-based methods may not be optimal for downstream tasks like image classification, and finding an informative corset may be challenging when the dataset's information is not concentrated in a few samples. Instead, our approach learns a computationally efficient synthetic set that is not limited to a subset of the original training samples.
\paragraph {Attention Mechanism.} \label{attention}
Attention has been widely used in deep learning to improve performance on various tasks \cite{bahdanau2015neural, wang2018non, zagoruyko2016paying}, with initial applications in natural language processing by Bahdanau \etal \cite{bahdanau2015neural} for language translation. Attention has since been used in computer vision, with global attention models \cite{wang2018non} for improved classification accuracy on image datasets and convolutional block attention modules \cite{woo2018cbam} for learning to attend to informative feature maps. Attention has also been used for model compression in knowledge distillation \cite{zagoruyko2016paying}. However, this mechanism has not been explored in the context of dataset distillation. To fill this gap, we propose a spatial attention matching module to approximate the distribution of the real dataset.

\section{Methodology} \label{Meth.}

In this section, we propose a novel end-to-end framework called \textbf{Data}set \textbf{D}istillation with \textbf{A}ttention \textbf{M}atching (DataDAM), which leverages attention maps to synthesize data that closely approximates the real training data distribution. The high dimensionality of training images makes it difficult to estimate the real data distribution accurately. Therefore, we represent each training image using spatial attention maps generated by different layers within a family of randomly initialized neural networks. These maps effectively highlight the most discriminative regions of the input image that the network focuses on at different layers (early, intermediate, and last layers) while capturing low-, mid-, and high-level representation information of the image. Although each individual network provides a partial interpretation of the image, the family of these randomly initialized networks produces a more comprehensive representation.

\subsection{Dataset Distillation with Attention Matching} \label{DataDAM}

Given a large-scale dataset $\mathcal{T} = \{(\bm{x}_{i}, y_{i})\}_{i=1}^{|\mathcal{T}|}$ containing $|\mathcal{T}|$ real image-label pairs, we first initialize a learnable synthetic dataset $\mathcal{S} = \{(\bm{s}_{j}, y_{j})\}_{j=1}^{|\mathcal{S}|}$ with $|\mathcal{S}|$ synthetic image and label pairs, by using either random noise or a selection of real images obtained through random sampling or a clustering algorithm such as K-Center \cite{cuidc, seneractive}. For each class $k$, we sample a batch of real and synthetic data (\ie $B^{\mathcal{T}}_{k}$ and $B^{\mathcal{S}}_{k}$, resp.) and extract features using a neural network $\phi_{\boldsymbol{\theta}}(\cdot)$ with standard network random initialization $\boldsymbol{\theta}$ \cite{he2015delving}. Figure \ref{fig:DataDAM} shows the proposed approach, where the neural network $\phi_{\boldsymbol{\theta}}(\cdot)$, consisting of $L$ layers, is employed to embed the real and synthetic sets. The network generates feature maps for each dataset, represented as $\phi_{\boldsymbol{\theta}}({\mathcal{T}}_{k}) = [\bm{f}^{\mathcal{T}_{k}}_{\boldsymbol{\theta},1}, \cdots, \bm{f}^{\mathcal{T}_{k}}_{\boldsymbol{\theta}, L}]$ and $\phi_{\boldsymbol{\theta}}(\mathcal{S}_{k}) = [\bm{f}^{\mathcal{S}_{k}}_{\boldsymbol{\theta},1}, \cdots, \bm{f}^{\mathcal{S}_{k}}_{\boldsymbol{\theta}, L}]$, respectively. The feature $\bm{f}^{\mathcal{T}_{k}}_{\boldsymbol{\theta},l}$ is a multi-dimensional array in $\mathbb{R}^{|B^{\mathcal{T}}_{k}| \times C_{l} \times W_{l}\times H_{l}}$, coming from the real dataset in the $l^\text{th}$ layer, where $C_{l}$ represents the number of channels and $H_{l} \times W_{l}$ is the spatial dimensions. Similarly, a feature $\bm{f}^{\mathcal{S}_{k}}_{\boldsymbol{\theta},l}$ is extracted for the synthetic set.

The \textbf{S}patial \textbf{A}ttention \textbf{M}atching (SAM) module then generates attention maps for the real and synthetic images using a feature-based mapping function $A(\cdot)$. The function takes the feature maps of each layer (except the last layer) as an input and outputs two separate attention maps: $A\big(\phi_{\boldsymbol{\theta}}({\mathcal{T}}_{k})\big) = [\bm{a}^{\mathcal{T}_{k}}_{\boldsymbol{\theta},1}, \cdots, \bm{a}^{\mathcal{T}_{k}}_{\boldsymbol{\theta}, L-1}]$ and $A(\phi_{\boldsymbol{\theta}}({\mathcal{S}}_{k})) = [\bm{a}^{\mathcal{S}_{k}}_{\boldsymbol{\theta},1}, \cdots, \bm{a}^{\mathcal{S}_{k}}_{\boldsymbol{\theta}, L-1}]$ for the real and synthetic sets, respectively. Prior studies \cite{zagoruyko2016paying, zeiler2014visualizing} have shown that the absolute value of a hidden neuron activation can indicate its importance for a given input, thus we create a spatial attention map by aggregating the absolute values of the feature maps across the channel dimension. This means that the feature map $\bm{f}^{\mathcal{T}_{k}}_{\boldsymbol{\theta},l}$ of the $l^\text{th}$ layer is converted into a spatial attention map $\bm{a}^{\mathcal{T}_{k}}_{\boldsymbol{\theta},l} \in \mathbb{R}^{|B^{\mathcal{T}}_{k}| \times W_{l}\times H_{l}}$ using the following pooling operation: 
\begin{flalign} \label{pooling}
 A(\bm{f}^{\mathcal{T}_{k}}_{\boldsymbol{\theta},l}) = \sum_{i=1}^{C_{l}}\big|{(\bm{f}^{\mathcal{T}_{k}}_{\boldsymbol{\theta},l})}_{i}\big|^{p},
\end{flalign}
where, ${(\bm{f}^{\mathcal{T}_{k}}_{\boldsymbol{\theta},l})}_{i} = \bm{f}^{\mathcal{T}_{k}}_{\boldsymbol{\theta},l} (:,i,:,:) $ is the feature map of channel $i$ from the $l^\text{th}$ layer and the power and absolute value operations are applied element-wise. The resulting attention map emphasizes the spatial locations associated with neurons with the highest activations. This helps retain the most informative regions and generates a more efficient feature descriptor. In a similar manner, the attention maps for synthetic data can be obtained as $\bm{a}^{\mathcal{S}_{k}}_{\boldsymbol{\theta},l}$. The effect of parameter $p$ is studied in the supplementary materials.


To capture the distribution of the original training set at different levels of representations, we compare the normalized spatial attention maps of each layer (excluding the last layer) between the real and synthetic sets using the loss function $\mathcal{L}_\text{SAM}$, which is formulated as
\begin{flalign} \label{sam}
\displaystyle \mathop{\mathbb{E}}_{\boldsymbol{\theta}\sim P_{\boldsymbol{\theta}}}\bigg[\sum_{k=1}^{K}\sum_{l=1}^{L-1}\Big\lVert \displaystyle {\mathbb{E}}_{\mathcal{T}_{k}}\Big[\frac{\bm{z}^{\mathcal{T}_{k}}_{\boldsymbol{\theta}, l}}{{\lVert \bm{z}^{\mathcal{T}_{k}}_{\boldsymbol{\theta}, l}\rVert}_{2}}\Big] - \displaystyle \mathbb{E}_{{\mathcal{S}}_{k}}\Big[\frac{\bm{z}^{\mathcal{S}_{k}}_{\boldsymbol{\theta}, l}}{{\lVert \bm{z}^{\mathcal{S}_{k}}_{\boldsymbol{\theta}, l}\rVert}_{2}}\Big]\Big\rVert^{2}\bigg],
\end{flalign}
where, $\bm{z}^{\mathcal{T}_{k}}_{\boldsymbol{\theta}, l} = vec(\bm{a}^{\mathcal{T}_{k}}_{\boldsymbol{\theta},l}) \in \mathbb{R}^{|B^{\mathcal{T}}_{k}| \times (W_{l}\times H_{l})}$ and $\bm{z}^{\mathcal{S}_{k}}_{\boldsymbol{\theta}, l} = vec(\bm{a}^{\mathcal{S}_{k}}_{\boldsymbol{\theta},l}) \in \mathbb{R}^{|B^{\mathcal{S}}_{k}|\times (W_{l}\times H_{l})}$ are the $l^\text{th}$ pair of vectorized attention maps along the spatial dimension for the real and synthetic sets, respectively. The parameter $K$ is the number of categories in a dataset, and $P_{\boldsymbol{\theta}}$ denotes the distribution of network parameters. It should be noted that normalization of the attention maps in the SAM module improves performance on the syntactic set (see supplementary materials).

Despite the ability of $\mathcal{L}_\text{SAM}$ to approximate the real data distribution, a discrepancy still exists between the synthetic and real training sets. The features in the final layer of neural network models encapsulate the highest-level abstract information of the images in the form of an embedded representation, which has been shown to effectively capture the semantic information of the input data \cite{saito2018maximum, zhao2023dataset, ma2015hierarchical, gretton2012kernel}. Therefore, we leverage a complementary loss as a regularizer to promote similarity in the mean vectors of the embeddings between the two datasets for each class. To that end, we employ the widely known Maximum Mean Discrepancy (MMD) loss, $\mathcal{L}_{\text{MMD}}$, which is calculated within a family of kernel mean embeddings in a Reproducing Kernel Hilbert Space (RKHS) \cite{gretton2012kernel}. The $\mathcal{L}_{\text{MMD}}$ loss is formulated as
\begin{flalign} \label{mmd}
\displaystyle \mathop{\mathbb{E}}_{\boldsymbol{\theta}\sim P_{\boldsymbol{\theta}}}\bigg[\sum_{k=1}^{K}\Big\lVert \displaystyle {\mathbb{E}}_{\mathcal{T}_{k}}\Big[\tilde{\bm{f}}^{\mathcal{T}_{k}}_{\boldsymbol{\theta},L}\Big] - \displaystyle \mathbb{E}_{{\mathcal{S}}_{k}}\Big[\tilde{\bm{f}}^{\mathcal{S}_{k}}_{\boldsymbol{\theta},L}\Big]\Big\rVert_{\mathcal{H}}^{2}\bigg],
\end{flalign}
where $\mathcal{H}$ is a reproducing kernel Hilbert space. The $\tilde{\bm{f}}^{\mathcal{T}_{k}}_{\boldsymbol{\theta}, L} = vec({\bm{f}}^{\mathcal{T}_{k}}_{\boldsymbol{\theta}, L}) \in \mathbb{R}^{|B^{\mathcal{T}}_{k}| \times (C_{L} \times W_{L}\times H_{L})}$ and $\tilde{\bm{f}}^{\mathcal{S}_{k}}_{\boldsymbol{\theta}, L} = vec({\bm{f}}^{\mathcal{S}_{k}}_{\boldsymbol{\theta}, L}) \in \mathbb{R}^{|B^{\mathcal{S}}_{k}| \times (C_{L} \times W_{L}\times H_{L})}$ are the final feature maps of the real and synthetic sets in vectorized form with both the channel and spatial dimensions included. We estimate the expectation terms in Equations \ref{sam} and \ref{mmd} empirically if ground-truth data distributions are not available. Finally, we learn the synthetic dataset by solving the following optimization problem using SGD with momentum:
\begin{flalign} \label{loss}
\mathcal{S}^{*} = \argmin_{\mathcal{S}}\:\big(\mathcal{L}_{\text{SAM}} + \lambda \mathcal{L}_{\text{MMD}}\big),
\end{flalign}
where $\lambda$ is the task balance parameter. Further information on the effect of $\lambda$ is discussed in Section \ref{ablation}. Note that our approach assigns a fixed label to each synthetic sample and keeps it constant during training. A summary of the learning algorithm can be found in Algorithm \ref{alg:1}.
\begin{algorithm}[h] 
\caption{Dataset Distillation with Attention Matching}
\label{alg:1}
\textbf{Input:} \text{Real training dataset $\mathcal{T}=\{(\bm{x}_{i}, y_{i})\}_{i=1}^{|\mathcal{T}|}$}\\
\textbf{Required:} Initialized synthetic samples for $K$ classes, Deep neural network $\phi_{\bm{\theta}}$ with parameters $\bm{\theta}$, Probability distribution over randomly initialized weights $P_{\bm{\theta}}$, Learning rate $\eta_{\mathcal{S}}$, Task balance parameter $\lambda$, Number of training iterations $I$.
\begin{algorithmic}[1]
\State Initialize synthetic dataset $\mathcal{S}$ 
\For{$i = 1, 2, \cdots, I$}  \label{line:1}
	\State Sample $\bm{\theta}$ from $P_{\bm{\theta}}$
        \State Sample mini-batch pairs $B_{k}^{\mathcal{T}}$ and $B_{k}^{\mathcal{S}}$ from the real 
        \Statex \:\:\:\:\:\: and synthetic sets for each class $k$
        \State Compute $\mathcal{L}_{\text{SAM}}$ and $\mathcal{L}_{\text{MMD}}$ using Equations \ref{sam} and \ref{mmd}
        \State Calculate $\mathcal{L} = \mathcal{L}_{\text{SAM}} + \lambda \mathcal{L}_{\text{MMD}}$
        \State Update the synthetic dataset using $\mathcal{S} \leftarrow \mathcal{S} - \eta_{\mathcal{S}}\nabla_{\mathcal{S}}\mathcal{L}$
\EndFor
\end{algorithmic}

\textbf{Output:} \text{Synthetic dataset $\mathcal{S}=\{(\bm{s}_{i}, y_{i})\}_{i=1}^{|\mathcal{S}|}$}

\end{algorithm}
\section{Experiments}

\begin{table*}[!h]
\renewcommand\arraystretch{1}
\centering
\scriptsize
\setlength{\tabcolsep}{1.65pt}
\setlength{\abovecaptionskip}{0.1cm}
\resizebox{1\linewidth}{!}{
\begin{tabular}{cccc|cccc|cccccccc|c}
\toprule
\multirow{3}{*}{}           & \multirow{2}{*}{IPC} & \multirow{2}{*}{Ratio\%} & \multirow{2}{*}{Resolution} & \multicolumn{4}{c|}{Coreset Selection}   & \multicolumn{8}{c|}{Training Set Synthesis} & \multirow{2}{*}{Whole Dataset} \\ %
                            & &                & &
                            \multicolumn{1}{c}{Random}        & \multicolumn{1}{c}{Herding}       & \multicolumn{1}{c}{K-Center}      & \multicolumn{1}{c|}{Forgetting}   &                             \multicolumn{1}{c}{{DD}$^\dagger$\cite{wang2018dataset}}  & \multicolumn{1}{c}{{LD}$^\dagger$\cite{bohdal2020flexible}} & \multicolumn{1}{c}{{DC} \cite{zhao2021datasetDC}}             & \multicolumn{1}{c}{{DSA} \cite{zhao2021datasetDSA}} 	&	
                            \multicolumn{1}{c}{{DM} \cite{zhao2023dataset}}         &  \multicolumn{1}{c}{{CAFE} \cite{wang2022cafe}}         &\multicolumn{1}{c}{{KIP} \cite{nguyen2021dataset}} 
                             &\multicolumn{1}{c|}{{DataDAM}}
                              \\ \midrule




\multirow{3}{*}{CIFAR-10}  & 1   & 0.02  &32 & 14.4 $\pm$ 2.0  & 21.5 $\pm$ 1.2  & 21.5 $\pm$ 1.3 & 13.5 $\pm$ 1.2 & \multicolumn{1}{c}{-} & 25.7 $\pm$ 0.7  & 28.3 $\pm$ 0.5 & 28.8 $\pm$ 0.7 & 26.0 $\pm$ 0.8 & 31.6 $\pm$ 0.8 & 29.8 $\pm$ 1.0 & \multicolumn{1}{c|}{\bf{32.0 $\pm$ 1.2}} & \multirow{3}{*}{84.8 $\pm$ 0.1} \\

& 10  & 0.2 &32& 26.0 $\pm$ 1.2  & 31.6 $\pm$ 0.7 & 14.7 $\pm$ 0.9 & 23.3 $\pm$ 1.0 & 36.8 $\pm$ 1.2  & 38.3 $\pm$ 0.4  & 44.9 $\pm$ 0.5  & 52.1 $\pm$ 0.5 & 48.9 $\pm$ 0.6 & 50.9 $\pm$ 0.5 & 46.1 $\pm$ 0.7 & \multicolumn{1}{c|}{\bf{54.2 $\pm$ 0.8}}&\\ 

& 50  & 1 &32& 43.4 $\pm$ 1.0  & 40.4 $\pm$ 0.6 & 27.0 $\pm$ 1.4 & 23.3 $\pm$ 1.1 & \multicolumn{1}{c}{-} & 42.5 $\pm$ 0.4  & 53.9 $\pm$ 0.5  & 60.6 $\pm$ 0.5 & 63.0 $\pm$ 0.4 & 62.3 $\pm$ 0.4 & 53.2 $\pm$ 0.7 &\multicolumn{1}{c|}{\bf{67.0 $\pm$ 0.4}}& \\ 
\midrule
                                                            
\multirow{3}{*}{CIFAR-100} & 1   & 0.2  &32&  4.2 $\pm$ 0.3  & 8.3 $\pm$ 0.3 & 8.4 $\pm$ 0.3  &  4.5 $\pm$ 0.2 & \multicolumn{1}{c}{-} & 11.5 $\pm$ 0.4  & 12.8 $\pm$ 0.3  & 13.9 $\pm$ 0.3 & 11.4 $\pm$ 0.3 & 14.0 $\pm$ 0.3 & 12.0 $\pm$ 0.2 & \multicolumn{1}{c|}{\bf{14.5 $\pm$ 0.5}} & \multirow{3}{*}{56.2 $\pm$ 0.3}\\ 
                              
& 10  & 2  &32& 14.6 $\pm$ 0.5  & 17.3 $\pm$ 0.3  & 17.3 $\pm$ 0.3 & 15.1 $\pm$ 0.3   & \multicolumn{1}{c}{-} & \multicolumn{1}{c}{-} & 25.2 $\pm$ 0.3  & 32.3 $\pm$ 0.3 & 29.7 $\pm$ 0.3  & 31.5 $\pm$ 0.2 & 29.0 $\pm$ 0.3 & \multicolumn{1}{c|}{\bf{34.8 $\pm$ 0.5}}& \\

& 50  & 10 &32& 30.0 $\pm$ 0.4  & 33.7 $\pm$ 0.5  & 30.5 $\pm$ 0.3  & \multicolumn{1}{c|}{-} & \multicolumn{1}{c}{-} & \multicolumn{1}{c}{-}  & 30.6 $\pm$ 0.6  &  42.8 $\pm$ 0.4 & 43.6 $\pm$ 0.4  & 42.9 $\pm$ 0.2 & \multicolumn{1}{c}{-} & \multicolumn{1}{c|}{\bf{49.4 $\pm$ 0.3}}& \\  \midrule

\multirow{3}{*}{Tiny ImageNet} & 1   & 0.2 &64&  1.4 $\pm$ 0.1  &  2.8 $\pm$ 0.2  &  \multicolumn{1}{c}{-} & 1.6 $\pm$ 0.1  & \multicolumn{1}{c}{-}  & \multicolumn{1}{c}{-} & 5.3 $\pm$ 0.1 & 5.7 $\pm$ 0.1 &  3.9 $\pm$ 0.2  & \multicolumn{1}{c}{-} & \multicolumn{1}{c}{-} & \multicolumn{1}{c|}{\bf{8.3 $\pm$ 0.4}}  & \multirow{3}{*}{37.6 $\pm$ 0.4}\\ 

& 10  & 2  &64&  5.0 $\pm$ 0.2  &  6.3 $\pm$ 0.2  &  \multicolumn{1}{c}{-} & 5.1 $\pm$ 0.2   &\multicolumn{1}{c}{-}  & \multicolumn{1}{c}{-} & 12.9 $\pm$ 0.1  & 16.3 $\pm$ 0.2  & 12.9 $\pm$ 0.4  & \multicolumn{1}{c}{-} & \multicolumn{1}{c}{-}
 & \multicolumn{1}{c|}{\bf{18.7 $\pm$ 0.3}}\\

& 50  & 10 &64& 15.0 $\pm$ 0.4  & 16.7 $\pm$ 0.3  &  \multicolumn{1}{c}{-} & 15.0 $\pm$ 0.3   &\multicolumn{1}{c}{-} & \multicolumn{1}{c}{-} & 12.7 $\pm$ 0.4  & 5.1 $\pm$ 0.2  & 25.3 $\pm$ 0.2  & \multicolumn{1}{c}{-} & \multicolumn{1}{c}{-} & \multicolumn{1}{c|}{\bf{28.7 $\pm$ 0.3}}\\
\bottomrule
\end{tabular}}
\caption{The performance (testing accuracy \%) comparison to state-of-the-art methods. We distill the given number of images per class using the training set, train a neural network on the synthetic set from scratch, and evaluate the network on the testing data. IPC: image(s) per class. Ratio~(\%): the ratio of distilled images to the whole training set. The works {DD}$^\dagger$ and {LD}$^\dagger$ use AlexNet \cite{krizhevsky2017imagenet} for CIFAR-10 dataset. All other methods use a 128-width ConvNet for training and evaluation. \textbf{Bold entries} are the best results. Note: some entries are marked as absent due to scalability issues or unreported values. For more information, refer to the supplementary materials.}
\label{main1}
\end{table*}
In this section, we demonstrate the effectiveness of DataDAM in improving the performance of dataset distillation. We introduce the datasets and implementation details for reproducibility (Section \ref{exp.set}), compare our method with state-of-the-art benchmarks (Section \ref{sota}), conduct ablation studies to evaluate each component's efficacy and transferability across various architectures (Section \ref{ablation}), and show some visualizations (Section \ref{vis}). Finally, we demonstrate the applicability of our method to the common tasks of continual learning and neural architecture search (Section \ref{app}).
\subsection{Experimental Setup} \label{exp.set}

 \textbf{Datasets.}
Our method was evaluated on CIFAR10/100 datasets \cite{krizhevsky2009learning}, which have a resolution of 32 $\times$ 32, in line with state-of-the-art benchmarks. For medium-resolution data, we resized the Tiny ImageNet \cite{le2015tiny} and ImageNet-1K \cite{deng2009imagenet} datasets to 64 $\times$ 64. Previous work on dataset distillation \cite{cazenavette2022dataset} introduced subsets of ImageNet-1K that focused on categories and aesthetics, including assorted objects, dog breeds, and birds. We utilized these subsets, namely ImageNette, ImageWoof, and ImageSquawk, which consist of 10 classes, as high-resolution (128 $\times$ 128) datasets in our experimental studies. For more detailed information on the datasets, please refer to the supplementary materials.

\textbf{Network Architectures.}
We use a ConvNet architecture \cite{gidaris2018dynamic} for the distillation task, similar to prior research. The default ConvNet has three identical convolutional blocks and a linear classifier. Each block includes a 128-kernel 3 $\times$ 3 convolutional layer, instance normalization, ReLU activation, and 3 $\times$ 3 average pooling with a stride of 2. We adjust the network for medium- and high-resolution data by adding a fourth and fifth convolutional block to account for the higher resolutions, respectively. In all experiments, we initialize the network parameters using normal initialization \cite{he2015delving}.
 
\textbf{Evaluation.} 
We evaluate the methods using standard measures from prior studies \cite{zhao2023dataset, zhao2021datasetDC, wang2022cafe, zhao2021datasetDSA}. We generate five sets of small synthetic images using 1, 10, and 50 images per class (IPC) from a real training dataset. Next, we train 20 neural network models on each synthetic set using an SGD optimizer with a learning rate of 0.01. We report the mean and standard deviation over 100 models for each experiment to assess the effectiveness of the performance of distilled datasets. Additionally, we evaluate computational costs using run-time expressed per step, averaged over 100 iterations, and peak GPU memory usage during 100 iterations of training. Finally, we visualize the unbiasedness of state-of-the-art methods using t-SNE visualization \cite{van2008visualizing}. 


\begin{table}[!h]
\renewcommand\arraystretch{1}
\centering
\scriptsize
\setlength{\tabcolsep}{1.6pt}
\setlength{\abovecaptionskip}{0.1cm}
\resizebox{1\linewidth}{!}{
\begin{tabular}{cccc|ccc|c}
\toprule
        & {IPC} & {Ratio\%} & {Resolution} & Random & DM \cite{zhao2023dataset} & DataDAM & {Whole Dataset} \\ 
 \midrule
 
\multirow{3}{*}{ImageNet-1K} & 1   & 0.078 & 64 &  0.5 $\pm$ 0.1  &   1.3 $\pm$ 0.1 & {\bf{2.0 $\pm$ 0.1}}  & \multirow{3}{*}{33.8 $\pm$ 0.3}\\ 
& 2  & 0.156  &64&  0.9 $\pm$ 0.1   & 1.6 $\pm$ 0.1 & {\bf{2.2 $\pm$ 0.1}}\\
& 10  & 0.780  &64& 3.1 $\pm$ 0.2  & 5.7 $\pm$ 0.1   & \multicolumn{1}{c|}{\bf{6.3 $\pm$ 0.0}}\\
& 50  & 3.902  &64& 7.6 $\pm$ 1.2 & 11.4 $\pm$ 0.9  & \multicolumn{1}{c|}{\bf{15.5 $\pm$ 0.2}}\\
\midrule

\multirow{2}{*}{ImageNette} 
& 1 & 0.105 & 128&  23.5 $\pm$ 4.8 & 32.8 $\pm$ 0.5  & \textbf{34.7 $\pm$ 0.9}& \multirow{2}{*}{87.4 $\pm$ 1.0}\\
& 10 & 1.050& 128&  47.7 $\pm$ 2.4  & 58.1 $\pm$ 0.3 & \textbf{59.4 $\pm$ 0.4}\\
\midrule
    
\multirow{2}{*}{ImageWoof}  
& 1 & 0.110 & 128& 14.2 $\pm$ 0.9 & 21.1 $\pm$ 1.2 & \textbf{24.2 $\pm$ 0.5}& \multirow{2}{*}{67.0 $\pm$ 1.3}\\
& 10 & 1.100 & 128& 27.0 $\pm$ 1.9 &  31.4 $\pm$ 0.5 & \textbf{34.4 $\pm$ 0.4}\\
\midrule

\multirow{2}{*}{ImageSquawk}  
& 1 & 0.077 & 128&  21.8 $\pm$ 0.5 & 31.2 $\pm$ 0.7  & \textbf{36.4 $\pm$ 0.8}& \multirow{2}{*}{87.5 $\pm$ 0.3}\\
& 10 & 0.770 & 128& 40.2 $\pm$ 0.4 & 50.4 $\pm$ 1.2  &\textbf{55.4 $\pm$ 0.9}\\     
\bottomrule
\end{tabular}}
\caption{The performance (testing accuracy \%) comparison to state-of-the-art methods on ImageNet-1K \cite{deng2009imagenet} and ImageNet subsets \cite{cazenavette2022dataset}.}
\label{main2}
\end{table}

\textbf{Implementation Details.} 
We employ the SGD optimizer with a fixed learning rate of 1 to learn synthetic datasets with 1, 10, and 50 IPCs. We learn low- and medium/high-resolution synthetic images in 8000 iterations with a task balance ($\lambda$) of 0.01 and 0.02, respectively. Following from \cite{zhao2021datasetDSA}, we apply the differentiable augmentation strategy for learning and evaluating the synthetic set. For dataset reprocessing, we utilized the Kornia implementation of Zero Component Analysis (ZCA) with default parameters, following previous works \cite{nguyen2021dataset, cazenavette2022dataset}. All experiments are conducted on two Nvidia A100 GPUs. Further details on hyperparameters are available in the supplementary materials.

\subsection{Comparison to State-of-the-art Methods} \label{sota}
\textbf{Competitive Methods.}
We evaluate DataDAM against four corset selection approaches and eight advanced methods for training set synthesis. The corset selection methods include Random selection \cite{rebuffi2017icarl}, Herding \cite{castro2018end, belouadah2020scail}, K-Center \cite{seneractive}, and Forgetting \cite{tonevaempirical}. We also compare our approach with state-of-the-art distillation methods, including Dataset Distillation \cite{wang2018dataset} (DD), Flexible Dataset Distillation \cite{bohdal2020flexible} (LD), Dataset Condensation \cite{zhao2021datasetDC} (DC), Dataset Condensation with Differentiable Siamese Augmentation \cite{zhao2021datasetDSA} (DSA), Distribution Matching \cite{zhao2023dataset} (DM), Aligning Features \cite{wang2022cafe} (CAFE), Kernel Inducing Points \cite{nguyen2021dataset, nguyen2021dataset2} (KIP). As a distribution-matching centered work, we don't directly compare with Matching Training Trajectories \cite{cazenavette2022dataset} (MTT) as they incur the additional cost of training expert models and significant overhead in the distillation process (see Table~\ref{tab:runtime}), hence not an equitable comparison with our strategy. To ensure reproducibility, we downloaded publicly available distilled data for each baseline method and trained models using our experimental setup. We make minor adjustments to some methods to ensure a fair comparison, and for those that did not conduct experiments on certain data, we implemented them using the released author codes. For details on the implementation of baselines and comparisons to other methods such as generative models \cite{parmar2021dual, brocklarge, li2015generative}, please refer to the supplementary materials.

\textbf{Performance Comparison.}
We compare our method with selection- and synthesis-based approaches in Tables \ref{main1} and \ref{main2}. The results demonstrate that training set synthesis methods outperform coreset methods, especially when the number of images per class is limited to 1 or 10. This is due to the fact that synthetic training data is not limited to a specific set of real images. Moreover, our method consistently outperforms all baselines in most settings for low-resolution datasets, with improvements on the top competitor, DM, of 4.0\% and 5.8\% for the CIFAR10/100 datasets when using IPC50. This indicates that our DataDAM can achieve up to 88\% of the upper-bound performance with just 10\% of the training dataset on CIFAR100 and up to 79\% of the performance with only 1\% of the training dataset on CIFAR10. For medium- and high-resolution datasets, including Tiny ImageNet, ImageNet-1K, and ImageNet subsets, DataDAM also surpasses all baseline models across all settings. While existing methods fail to scale up to the ImageNet-1K due to memory or time constraints, DataDAM achieved accuracies of 2.0\%, 2.2\%, 6.3\%, and 15.5\% for 1, 2, 10, and 50 IPC, respectively, surpassing DM and Random by a significant margin. This improvement can be attributed to our methodology, which captures essential layer-wise information through spatial attention maps and the feature map of the last layer. Our ablation studies provide further evidence that the performance gain is directly related to the discriminative ability of the method in the synthetic image learning scheme.

\begin{table}[h]
\centering
\scriptsize
 \setlength{\abovecaptionskip}{0.1cm}
  \setlength{\tabcolsep}{5.8pt}
  \renewcommand\arraystretch{0.9}
\begin{tabular}{cccccc}
\toprule
&   \texttt{T}\textbackslash \texttt{E} & ConvNet      & AlexNet      & VGG-11          & ResNet-18   \\ 	\midrule


DC \cite{zhao2021datasetDC} & ConvNet  & 53.9$\pm$0.5 & 28.8$\pm$0.7 & 38.8$\pm$1.1 & 20.9$\pm$1.0 \\
 
CAFE \cite{wang2022cafe} & ConvNet  & 62.3$\pm$0.4 & 43.2$\pm$0.4 & 48.8$\pm$0.5 & 43.3$\pm$0.7 \\

DSA \cite{zhao2021datasetDSA}  & ConvNet & 60.6$\pm$0.5 & 53.7$\pm$0.6 & 51.4$\pm$1.0 & 47.8$\pm$0.9\\

DM \cite{zhao2023dataset}  & ConvNet & 63.0$\pm$0.4 & 60.1$\pm$0.5 & 57.4$\pm$0.8 & 52.9$\pm$0.4 \\

KIP \cite{nguyen2021dataset}  & ConvNet & 56.9$\pm$0.4 & 53.2$\pm$1.6 & 53.2$\pm$0.5 & 47.6$\pm$0.8 \\

\midrule
\multirow{3}{*}{DataDAM} & ConvNet & \bf{67.0$\pm$0.4} &\bf{63.9$\pm$0.9} & \bf{64.8$\pm$0.5} & \bf{60.2$\pm$0.7}\\

& AlexNet  & 61.8$\pm$0.6 & 60.6$\pm$0.9 & 61.8$\pm$0.6 & 56.4$\pm$0.7\\

& VGG-11  & 56.5$\pm$0.4 & 53.7$\pm$1.5 & 56.2$\pm$0.6 & 52.0$\pm$0.7\\
\bottomrule
\end{tabular}
\caption{{Cross-architecture testing performance (\%) on CIFAR10 with 50 images per class. The synthetic set is trained on one architecture (T) and then evaluated on another architecture (E).}}
\label{tab:crsarc}
\end{table}

\textbf{Cross-architecture Generalization.} \label{cross} 
In this section, we test our learned synthetic data across different unseen neural architectures, consistent with state-of-the-art benchmarks \cite{zhao2021datasetDC, zhao2023dataset}. To that end, synthetic data was generated from CIFAR10 using one architecture (T) with IPC50 and then transferred to a new architecture (E), where it was trained from scratch and tested on real-world data. Popular CNN architectures like ConvNet \cite{gidaris2018dynamic}, AlexNet \cite{krizhevsky2017imagenet}, VGG-11 \cite{simonyan2014very}, and ResNet-18 \cite{he2016deep} are used to examine the generalization performance.


Table \ref{tab:crsarc} shows that DataDAM outperforms state-of-the-art across unseen architectures when the synthetic data is learned with ConvNet. We achieve a margin of 3.8\% and 7.4\% when transferring to AlexNet and VGG-11, respectively, surpassing the best method, DM. Additionally, the remaining architectures demonstrate improvement due to the robustness of our synthetic images and their reduced architectural bias, as seen in the natural appearance of the distilled images (Figure \ref{fig:synimages}).

\textbf{Training Cost Analysis.}
In dataset distillation, it is crucial to consider the resource-time costs of various methods, particularly in terms of scalability. This study compares our method to state-of-the-art benchmarks presented in Table \ref{tab:runtime}. We demonstrate a significantly lower run-time by almost 2 orders of magnitude compared to most state-of-the-art results. Our method, like DM, has an advantage over methods such as DC, DSA, and MTT that require costly inner-loop bi-level optimization. It should be noted that DataDAM can leverage information from randomly initialized neural networks without training and consistently achieve superior performance.

\begin{table}[h]
  \centering
  \setlength{\abovecaptionskip}{0.1cm}
  \resizebox{0.48\textwidth}{!}{
  \begin{tabular}{c|ccc|ccc}
    \toprule
    \multirow{2}{*}{Method}&\multicolumn{3}{|c}{run time(sec)}&\multicolumn{3}{|c}{GPU memory(MB)}\\
    & IPC1 & IPC10 & IPC50 & IPC1 & IPC10 & IPC50 \\
    \midrule
    DC\cite{zhao2021datasetDC}&0.16~$\pm$~0.01&3.31~$\pm$~0.02&15.74~$\pm$~0.10&3515&3621&4527\\
    DSA\cite{zhao2021datasetDSA}&0.22~$\pm$~0.02&4.47~$\pm$~0.12&20.13~$\pm$~0.58&3513&3639&4539\\
    DM\cite{zhao2023dataset}&0.08~$\pm$~0.02&0.08~$\pm$~0.02&0.08~$\pm$~0.02&3323&3455&3605\\
    MTT\cite{cazenavette2022dataset}&0.36~$\pm$~0.23&0.40~$\pm$~0.20&OOM&2711&8049&OOM\\
    DataDAM& 0.09~$\pm$~0.01&0.08~$\pm$~0.01&0.16~$\pm$~0.04& 3452 & 3561 & 3724\\
    \bottomrule
  \end{tabular}
  }
  \caption{Training time and GPU memory comparisons for state-of-the-art synthesis methods. Run time is expressed per step, averaged over 100 iterations. GPU memory is expressed as the peak memory usage during 100 iterations of training. All methods were run on an A100 GPU for CIFAR-10. OOM (out-of-memory) is reported for methods that are unable to run within the GPU memory limit.}
    \label{tab:runtime}
\end{table}

\subsection{Ablation Studies} \label{ablation} 
In this section, we evaluate the robustness of our method under different experimental configurations. All experiments averaged performance over 100 randomly initialized ConvNets across five synthetic sets. The CIFAR10 dataset is used for all studies. The most relevant ablation studies to our method are included here; further ablative experiments are included in the supplementary materials.

\begin{figure}
    \centering
    \setlength{\abovecaptionskip}{0.1cm}
    \includegraphics[width=0.48\textwidth]{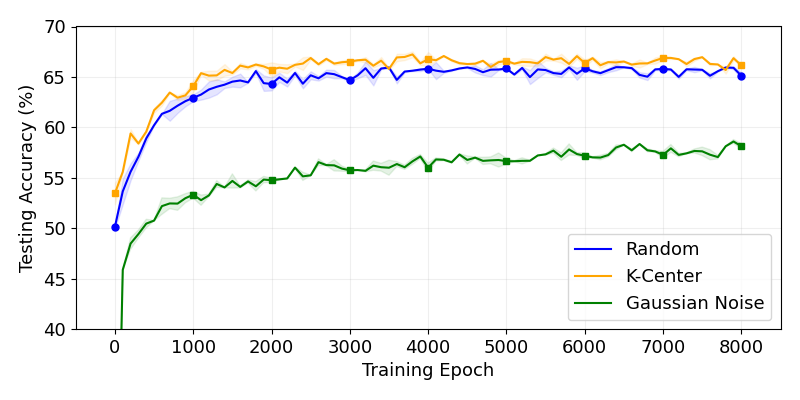}
    \caption{Test accuracy evolution of synthetic image learning on CIFAR10 with IPC50 under three different initializations: Random, K-Center, and Gaussian noise.}
    \label{fig:Init}
\end{figure}

\textbf{Exploring the importance of different initialization methods for synthetic images.}
In dataset distillation, synthetic images are usually initialized through Gaussian noise or sampled from the real data; however, the choice of initialization method has proved to be crucial to the overall performance \cite{cuidc}. To assess the robustness of DataDAM, we conducted an empirical evaluation with an IPC50 under three initialization conditions: Random selection, K-Center \cite{cuidc, seneractive}, and Gaussian noise (Figure \ref{fig:Init}). As reported in \cite{cuidc}, other works including \cite{zhao2023dataset, zhao2021datasetDSA, zhao2021datasetDC} have seen benefits to testing performance and convergence speed by leveraging K-Center as a smart selection. Empirically, we show that our method is robust across both random and K-Center with only a minute performance gap, and thus the initialization of synthetic data is not as crucial to our final performance. Finally, when comparing with noise, we notice a performance reduction; however, based on the progression over the training epochs, it appears our method is successful in transferring the information from the real data onto the synthetic images. For further detailed experimental results, please refer to the supplementary materials.

\begin{figure}
    \centering
    \includegraphics[width=0.47\textwidth]{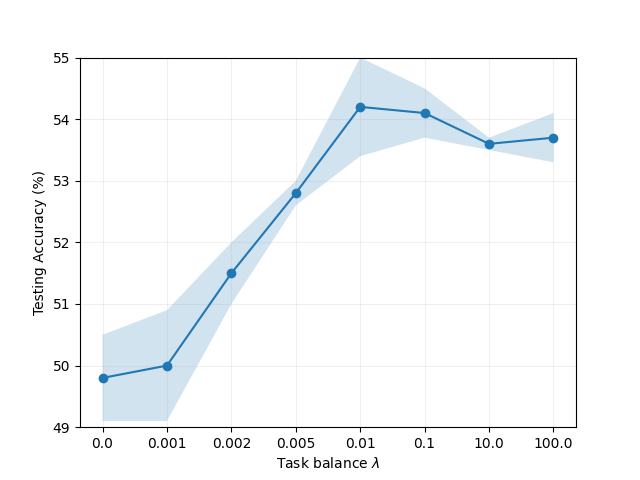}
    \caption{The effect of task balance $\lambda$ on the testing accuracy (\%) for CIFAR10 dataset with IPC10 configuration.}
    \label{fig:lambda}
    \vspace{-1.7em}
\end{figure}

\textbf{Evaluation of task balance $\boldsymbol{\lambda}$ in DataDAM.}
It is common in machine learning to use regularization to prevent overfitting and improve generalization. In the case of DataDAM, the regularizing coefficient $\lambda$ controls the trade-off between the attention matching loss $\mathcal{L}_{\text{SAM}}$ and the maximum mean discrepancy loss $\mathcal{L}_{\text{MMD}}$, which aims to reduce the discrepancy between the synthetic and real training distributions. The experiments conducted on the CIFAR10 dataset with IPC 10 showed that increasing the value of $\lambda$ improved the performance of DataDAM up to a certain point (Figure \ref{fig:lambda}). This is because, at lower values of $\lambda$, the attention matching loss dominates the training process, while at higher values of $\lambda$, the regularizer contributes more effectively to the overall performance. The results in Figure \ref{fig:lambda} also indicate that the method is robust to larger regularization terms, as shown by the plateau to the right of 0.01. Therefore, a task balance of 0.01 is chosen for all experiments on low-resolution data and 0.02 on medium- and high-resolution data.

\textbf{Evaluation of loss components in DataDAM.}
We conducted an ablation study to evaluate the contribution of each loss component, namely spatial attention matching loss ($\mathcal{L}_{\text{SAM}}$) and the complementary loss ($\mathcal{L}_{\text{MMD}}$), to the final performance of DataDAM. As seen in table \ref{tab:component}, the joint use of $\mathcal{L}_{\text{MMD}}$ and $\mathcal{L}_{\text{SAM}}$ led to state-of-the-art results, while using $\mathcal{L}_{\text{MMD}}$ alone resulted in significant underperformance, as it emphasizes the extraction of high-level abstract data but fails to capture different level representations of the real training distribution. On the other hand, $\mathcal{L}_{\text{SAM}}$ alone outperformed the base complementary loss, indicating the extracted discriminative features contain significant information about the training but still have room for improvement. To highlight the importance of intermediate representations, we compared our attention-based transfer approach with the transfer of layer-wise feature maps, similar to CAFE \cite{wang2022cafe}, and demonstrated a significant performance gap (see "Feature Map Transfer" in Table \ref{tab:component}). Overall, our findings support the use of attention to match layer-wise representations and a complementary loss to regulate the process.

\begin{table}[h]
\centering
\setlength{\abovecaptionskip}{0.1cm}
\renewcommand\arraystretch{0.9}
\scriptsize
        \setlength{\tabcolsep}{8pt}
\begin{tabular}{ccc|c}
	\toprule
	$\mathcal{L}_{\text{MMD}}$ & $\mathcal{L}_{\text{SAM}}$ & Feature Map Transfer &
 Testing Performance (\%)    \\
	\midrule
  
	\checkmark & - & - &48.9 $\pm$ 0.6 \\
  - & \checkmark & - &49.8 $\pm$ 0.7 \\
  - & - &\checkmark &47.2 $\pm$ 0.3 \\
  
	 \checkmark & \checkmark & - & \bf{54.2 $\pm$ 0.8} \\
\bottomrule
\end{tabular}
\caption{Evaluation of loss components in DataDAM.}
\label{tab:component}
\end{table}

\textbf{Exploring the effect of each layer in DataDAM.}
Following the previous ablation, it is equally important to examine how each layer affects the final performance. As shown in Table \ref{tab:nonsubset}, different layers perform differently since each provides different levels of information about the data distributions. This finding supports the claim that matching spatial attention maps in individual layers alone cannot obtain promising results. As a result, to improve the overall performance of the synthetic data learning process, it is crucial to transfer different levels of information about the real data distribution using the SAM module across all intermediate layers.
\begin{table}[H]
\renewcommand\arraystretch{0.9}
\centering
\scriptsize
\setlength{\tabcolsep}{11pt}
\setlength{\abovecaptionskip}{0.1cm}
\begin{tabular}{ccc|c}
	\toprule
	Layer 1 &Layer 2  & Last Layer &  Testing Performance (\%)     \\
	\midrule
	          - & - &   \checkmark& 48.9 $\pm$ 0.6 \\
	 \checkmark & - &  \checkmark &  50.2 $\pm$ 0.4\\
      - &  \checkmark & \checkmark &   51.5 $\pm$ 1.0\\
        \checkmark&  \checkmark & - &   49.8 $\pm$ 0.7\\
      \checkmark & \checkmark & \checkmark  &  \bf{54.2 $\pm$ 0.8}\\
\bottomrule
\end{tabular}
\caption{Evaluation of each layer's impact in ConvNet (3-layer). The output is transferred under $\mathcal{L}_{\text{MMD}}$ while the effects of the specified layers are measured through $\mathcal{L}_{\text{SAM}}$. We evaluate the performance of the CIFAR10 dataset with IPC10.}
\label{tab:nonsubset}
\end{table}

\textbf{Network Distributions.}
We investigate the impact of network initialization on DataDAM's performance by training 1000 ConvNet architectures with random initializations on the original training data and categorizing their learned states into five buckets based on testing performance. We sampled networks from each bucket and trained our synthetic data using IPCs 1, 10, and 50. As illustrated in Table \ref{tab:distribution}, our findings indicate that DataDAM is robust across various network initializations. This is attributed to the transfer of attention maps that contain relevant and discriminative information rather than the entire feature map statistics, as shown in \cite{wang2022cafe}. These results reinforce the idea that achieving state-of-the-art performance does not require inner-loop model training.
\begin{table}[H]
\renewcommand\arraystretch{0.9}
\setlength{\abovecaptionskip}{0.1cm}
\centering
\scriptsize
\setlength{\tabcolsep}{1.2pt}
\begin{tabular}{cc|c|c|c|c|c}
\toprule
  IPC & Random & 0-20 & 20-40 & 40-60 & 60-80 & $\geq$80 \\ \midrule
1  & $\mathbf{32.0\pm{2.0}}$   & $30.8\pm{1.1}$  & $30.7\pm{1.7}$  & $31.5\pm{1.9}$  & $26.2\pm{1.8}$  & $26.9\pm{1.3}$                            \\
10 & $\mathbf{54.2\pm{0.8}}$   & $54.0\pm{0.7}$  & $53.1\pm{0.5}$  & $52.1\pm{0.8}$  & $51.2\pm{0.7}$  & $51.7\pm{0.7}$                               \\
50 & $\mathbf{67.0\pm{0.4}}$   & $66.2\pm{0.4}$  & $66.4\pm{0.4}$  & $\mathbf{67.0\pm{0.5}}$  & $65.8\pm{0.5}$  & $65.3\pm{0.6}$                           \\\bottomrule
\end{tabular}
\caption{Performance of synthetic data learned with IPCs 1, 10, and 50 for different network initialization. Models are trained on the training set and grouped by their respective accuracy levels.}
\label{tab:distribution}
\end{table}

\subsection{Visualization} \label{vis}

\begin{figure}[H]
    \centering
    \includegraphics[width=0.47\textwidth]{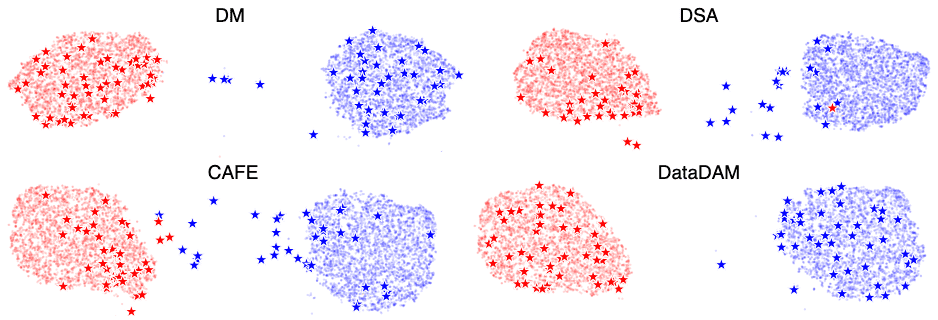}
    \caption{Distributions of synthetic images learned by four methods on CIFAR10 with IPC50. The stars represent the synthetic data dispersed amongst the original training dataset.}
    \label{fig:smalltsne}
\end{figure}

\textbf{Data Distribution.}
To evaluate whether our method can capture a more accurate distribution from the original dataset, we use t-SNE \cite{van2008visualizing} to visualize the features of real and synthetic sets generated by DM, DSA, CAFE, and DataDAM in the embedding space of the ResNet-18 architecture. Figure \ref{fig:smalltsne} shows that methods such as DSA and CAFE are biased towards the edges of their clusters and not representative of the training data. Much like DM, our results indicate a more equalized distribution, allowing us to better capture the data distribution. Preserving dataset distributions is of utmost importance in fields like ethical machine learning since methods that cannot be impartial in capturing data distribution can lead to bias and discrimination. Our method's capacity to capture the distribution of data makes it more appropriate than other approaches in these conditions, particularly in fields such as facial detection for privacy \cite{ciftci2023my}.

\begin{figure*}[h]
  \centering
  \begin{subfigure}[b]{0.24\textwidth}
    \includegraphics[width=\linewidth]{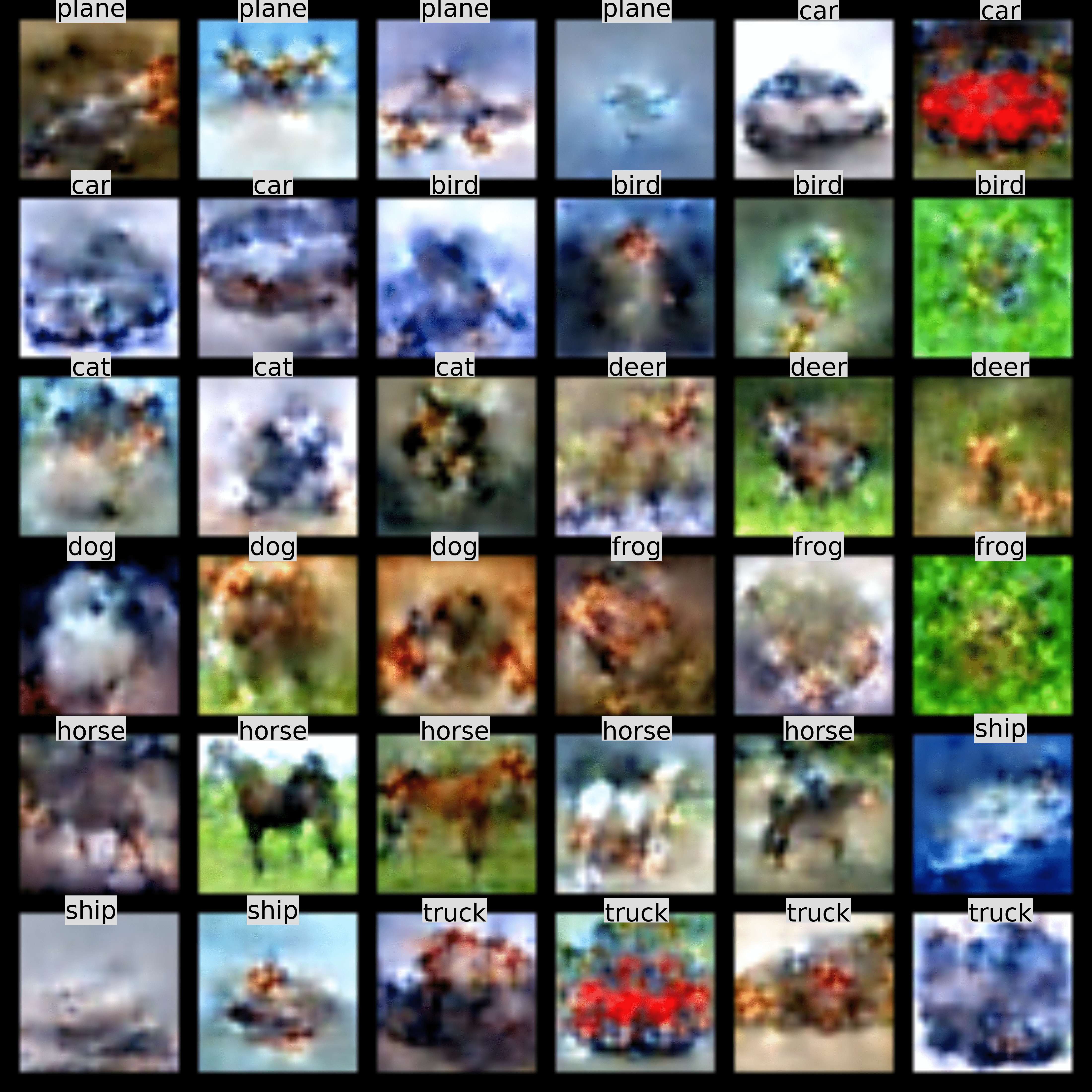}
    \caption{CIFAR10}
    \label{fig:sub1}
  \end{subfigure}\hfill
  \begin{subfigure}[b]{0.24\textwidth}
    \includegraphics[width=\linewidth]{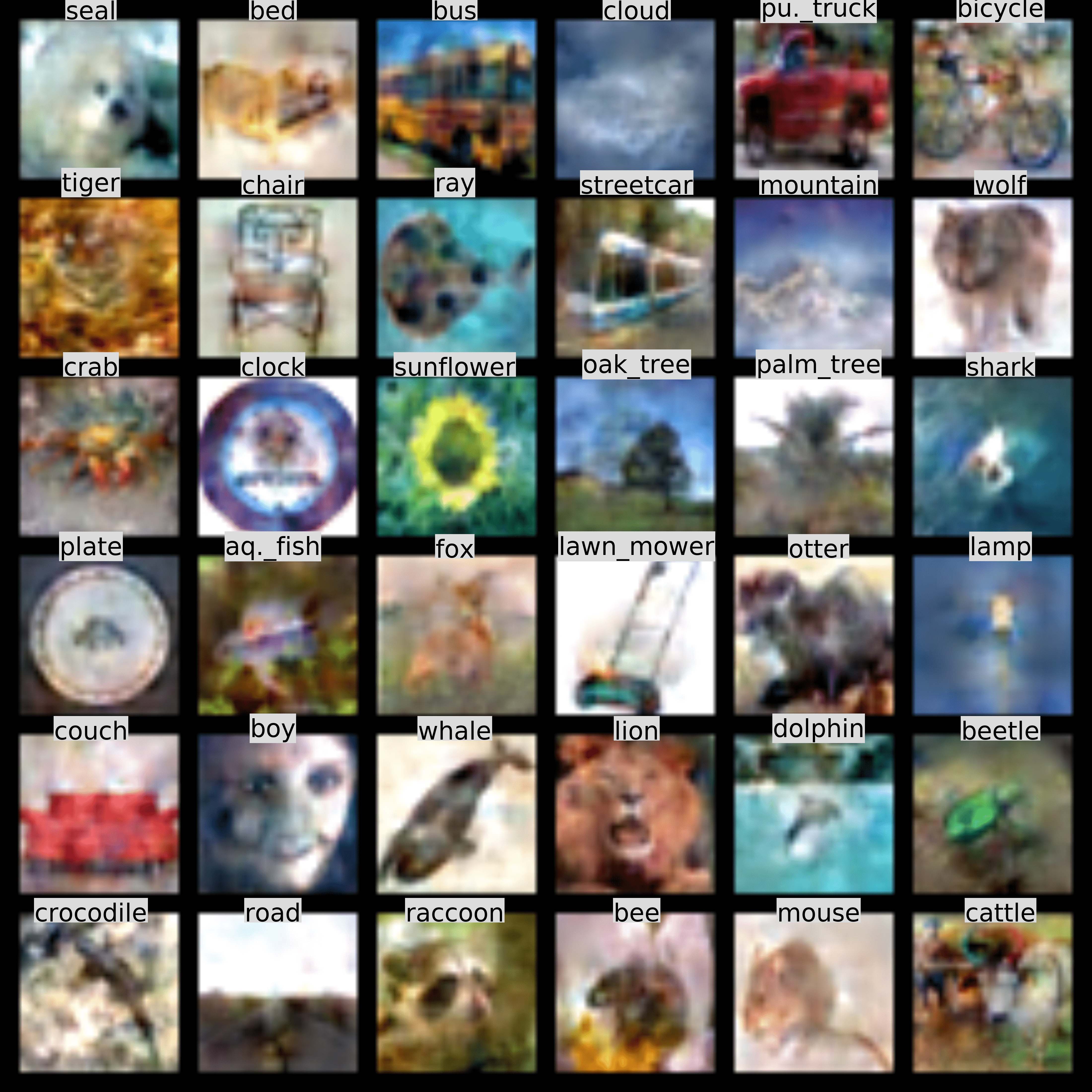}
    \caption{CIFAR100}
    \label{fig:sub2}
  \end{subfigure}\hfill
  \begin{subfigure}[b]{0.24\textwidth}
    \includegraphics[width=\linewidth]{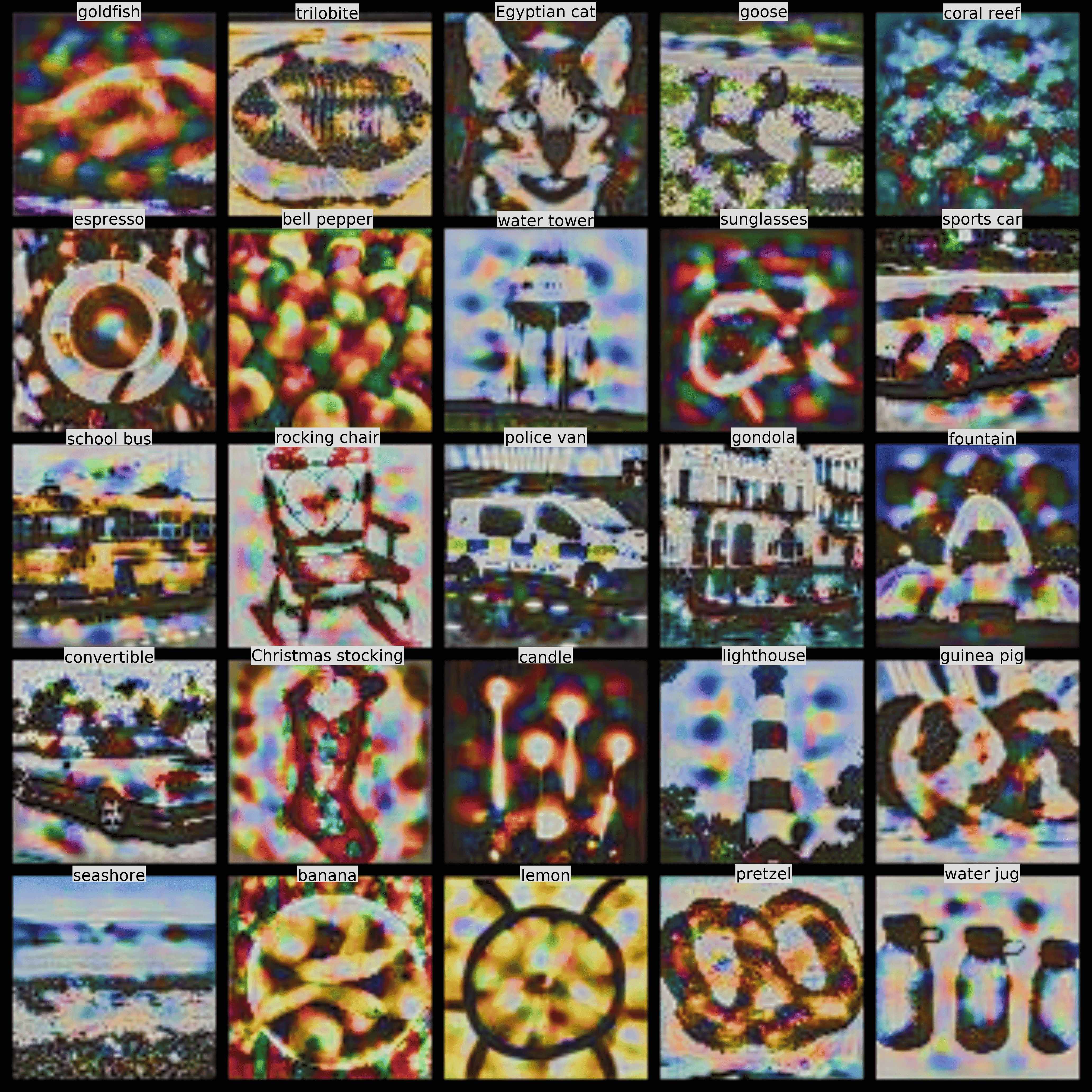}
    \caption{Tiny ImageNet}
    \label{fig:sub3}
  \end{subfigure}\hfill
  \begin{subfigure}[b]{0.24\textwidth}
    \includegraphics[width=\linewidth]{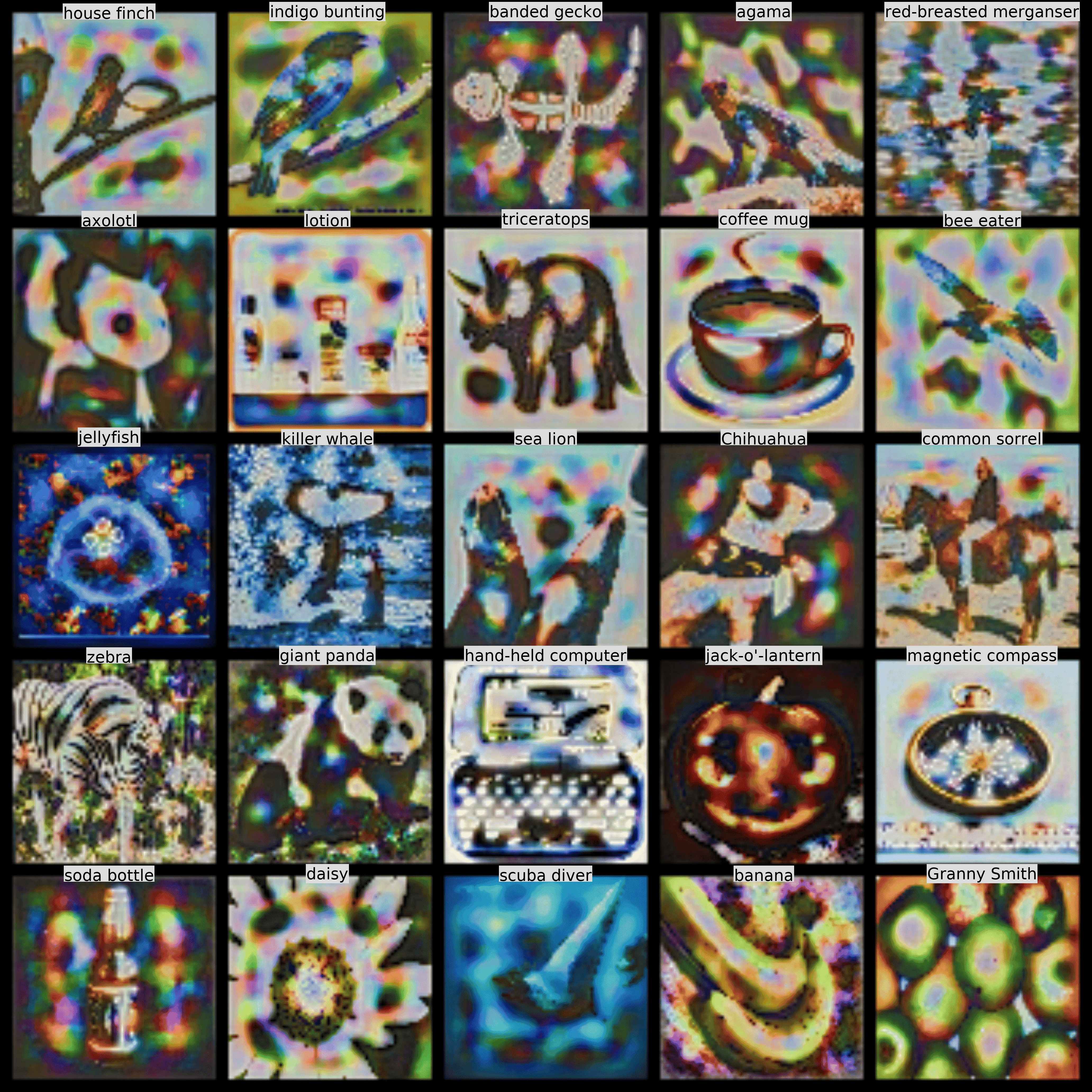}
    \caption{ImageNet-1K}
    \label{fig:sub4}
  \end{subfigure}
  
  \caption{Example distilled images from 32x32 CIFAR10/100 (IPC10), 64x64 Tiny ImageNet (IPC1), and 64x64 ImageNet-1K (IPC1).} 
  \label{fig:synimages}
  \vspace{-1em}
\end{figure*}
\textbf{Synthetic Images.}
We have included samples from our learned synthetic images for different resolutions in Figure \ref{fig:synimages}. In low-resolution images, the objects are easily distinguishable, and their class labels can be recognized intuitively. As we move to higher-resolution images, the objects become more outlined and distinct from their backgrounds. These synthetic images have a natural look and can be transferred well to different architectures. Moreover, the high-resolution images accurately represent the relevant colors of the objects and provide more meaningful data for downstream tasks. For more visualizations, refer to the supplementary materials.
\subsection{Applications} \label{app}
We assess the effectiveness of DataDAM's performance through the use of two prevalent applications involving dataset distillation algorithms: \textit{continual learning} and \textit{neural architecture search}.

\textbf{Continual Learning.}
Continual learning trains a model incrementally with new task labels to prevent catastrophic forgetting \cite{rebuffi2017icarl}. One approach is to maintain a replay buffer that stores balanced training examples in memory and train the model exclusively on the latest memory, starting from scratch \cite{rebuffi2017icarl, bang2021rainbow, prabhu2020gdumb}. Efficient storage of exemplars is crucial for optimal continual learning performance, and condensed data can play a significant role. We use the class-incremental setting from \cite{zhao2023dataset} with an augmented buffer size of 20 IPC to conduct class-incremental learning on the CIFAR100 dataset. We compare our proposed memory construction approach with random \cite{prabhu2020gdumb}, herding \cite{castro2018end, belouadah2020scail, rebuffi2017icarl}, DSA \cite{zhao2021datasetDSA}, and DM \cite{zhao2023dataset} methods at 5 and 10 learning steps. In each step, including the initial one, we added 400 and 200 distilled images to the replay buffer, respectively, following the class split of \cite{zhao2023dataset}. The test accuracy is the performance metric, and default data preprocessing and ConvNet are used for each approach.

\begin{figure}[H]
    \centering
    \includegraphics[width=0.485\textwidth]{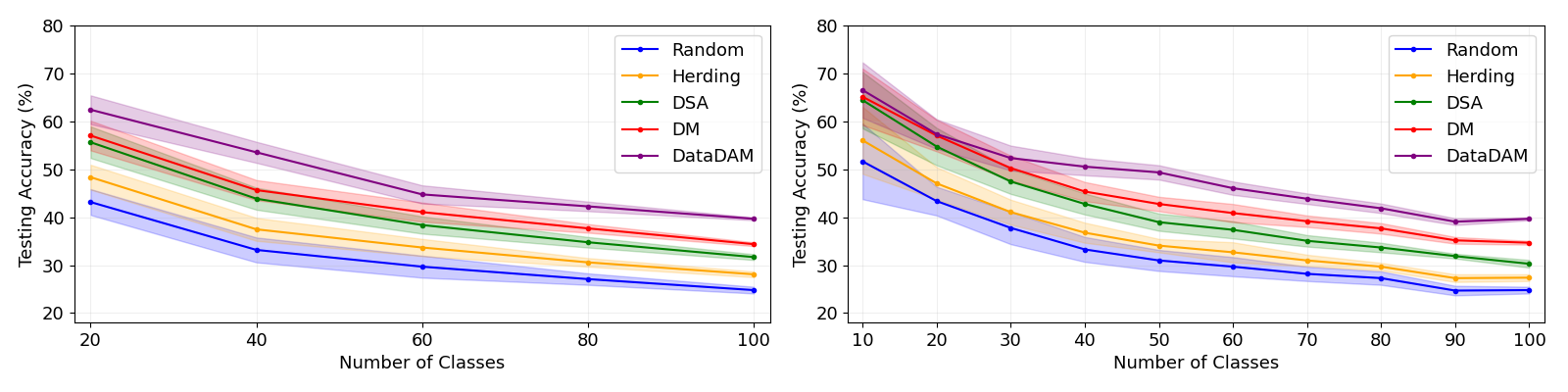}
    \caption{(Left): Showcases 5-step and (Right): Showcases 10-step continual learning with tolerance region.}
    \label{fig:cont}
\end{figure}

Figure \ref{fig:cont} shows that our memory construction approach consistently outperforms others in both settings. Specifically, DataDAM achieves final test accuracies of 39.7\% and 39.7\% in 5-step and 10-step learning, respectively, outperforming DM (34.4\% and 34.7\%), DSA (31.7\% and 30.3\%), herding (28.1\% and 27.4\%), and random (24.8\% and 24.8\%). Notably, the final performance of DataDAM, DM, and random selection methods remains unchanged upon increasing the number of learning steps, as these methods independently learn the synthetic datasets for each class. Our findings reveal that DataDAM provides more informative training to the models than other baselines, resulting in more effective prevention of memory loss associated with past tasks.
\textbf{Neural Architecture Search.}
Our synthetic sets can be used as a proxy set to accelerate model evaluation in Neural Architecture Search (NAS). Following \cite{zhao2021datasetDC}, we establish a 720 ConvNet search space on CIFAR10 with a grid varying in network depth, width, activation, normalization, and pooling layers. We compared our method with Random, DSA, CAFE, early stopping, and DM. Each architecture was trained on the proxy set (synthetic 50 IPC) for 200 epochs and the whole dataset for 100 epochs to establish a baseline performance metric. Early stopping still uses the entire dataset, but we limit the iterations to those of the proxy set, as in \cite{zhao2023dataset}. For each method, we rank all the architectures based on the validation performance and report the testing accuracy of the best-selected model when trained on the whole dataset in Table \ref{tab:nas}. DataDAM achieved the best accuracy among the competitors, with an accuracy of 89.0\%, which is very similar to the original training data at 89.2\%, indicating the potential of our proxy set to accurately represent the training data. Furthermore, we calculated Spearman's correlation over the entire search space to evaluate the robustness of our learned data in architecture searching. The correlation is calculated between the testing performances of each method when trained on the proxy versus the original training data. Our method achieves the highest correlation (0.72), indicating that it generates a suitable proxy set that is generalizable across the entire search space and encodes the most important and relevant information from the training data into a condensed form. For more experimentation with NAS, refer to the supplementary materials.
\begin{table}[H]
\renewcommand\arraystretch{0.7}
\centering
\scriptsize
\setlength{\tabcolsep}{1.2pt}
\begin{tabular}{ccccccc|c}
\toprule
                    & Random            & DSA               & DM             & CAFE & Ours & Early-stopping & Whole Dataset \\ \midrule
Performance (\%)    &        88.9         &             87.2   &         87.2   &  83.6 &  \bf{89.0}  &   88.9 & 89.2  \\
Correlation    &        0.70         &        0.66        &       0.71     &  0.59 &  \bf{0.72}   &  0.69  &  1.00\\
Time cost (min)     & 206.4  &   206.4    &  206.6 &     206.4  &   206.4  &   206.2    & 5168.9 \\ 
Storage (imgs)      & \bf{500}          & \bf{500}          & \bf{500}        &   \bf{500}  &   \bf{500} & $5\times 10^4$    & $5\times 10^4$   \\ \bottomrule
\end{tabular}
\vspace{-8pt}
\caption{{Neural architecture search on CIFAR10.}}
\label{tab:nas}
\end{table}

\section{Conclusion and Limitations}

Our proposed method, Dataset Distillation with Attention Matching (DataDAM), efficiently captures real datasets' most informative and discriminative information. It consists of two modules, spatial attention matching (SAM) and last-layer feature alignment, that match attention maps and embedded representations generated by different layers in randomly initialized neural networks, respectively. We conduct extensive experiments on datasets with different resolutions to show that DataDAM could lower CNN training costs while maintaining superior generalization performance. We also offer two applications that take advantage of our distilled set: continual learning and neural architecture search. In the future, we plan to apply DataDAM to more fine-grained datasets and explore the analytical concepts behind them.
\paragraph{Limitations.} DataDAM exhibits robust generalization across various CNN architectures, but it is limited to convolutional networks due to its formulation. For example, it, along with other data distillation algorithms, faces challenges in achieving successful cross-architecture generalization on ViT (Vision Transformer) models. Additionally, all data distillation methods, including DataDAM, need to be re-optimized when the distillation ratio changes, which can limit efficiency in some applications.


\newpage

{\small
\bibliographystyle{ieee_fullname}
\bibliography{main}
}

\section{Supplementary Materials}


\subsection{Implementation Details} \label{impl}
\subsubsection{Datasets} \label{datasets}
We carried out experiments on the following datasets: CIFAR10/100 \cite{krizhevsky2009learning}, TinyImageNet \cite{le2015tiny}, ImageNet-1K \cite{deng2009imagenet}, and subsets of ImageNet-1K including ImageNette \cite{howard2019imagenette}, ImageWoof \cite{howard2019imagenette}, and ImageSquawk \cite{cazenavette2022dataset}. CIFAR10/100 is a standard computer vision dataset consisting of natural images with colored 32x32 pixels. It has 10 coarse-grained labels (CIFAR10) and 100 fine-grained labels (CIFAR100), each with 50,000 training samples and 10,000 tests. The classes of the CIFAR10 are "Airplane", "Car", "Bird", "Cat", "Deer", "Dog", "Frog", "Horse", "Ship", and "Truck," which are mutually exclusive. TinyImageNet is a subset of the ImageNet-1K dataset with 200 classes. The dataset contains 100,000 high-resolution training images and 10,000 test examples that are downsized to 64x64. ImageNet-1K is a standard large-scale dataset with 1,000 classes, including 1,281,167 training examples and 50,000 testing images. Following \cite{zhoudataset, zhao2023dataset}, we resize ImageNet-1K images to 64x64 resolution to match TinyImageNet. Compared to CIFAR10/100, TinyImageNet and ImageNet-1K are more challenging because of their diverse classes and higher image resolution. To further extend dataset distillation, we take a step forward by applying our method to even higher-resolution images, specifically 128x128 subsets of ImageNet. In previous dataset distillation research \cite{cazenavette2022dataset}, subsets were introduced based on categories and aesthetics, encompassing birds, fruits, and cats. In this study, we utilize ImageNette (assorted objects), ImageWoof (dog breeds), and ImageSquawk (birds) to provide additional examples of our algorithm's effectiveness. For a detailed enumeration of ImageNet classes in each of our datasets, please refer to Table \ref{tab:classes}.
\begin{table*}[h]
\scriptsize
\centering
\setlength{\tabcolsep}{3pt}
\begin{tabular}{c|cccccccccc}
Dataset    & 0                & 1                & 2               & 3                  & 4              & 5           & 6               & 7           & 8                  & 9                \\ \hline

ImageNette \cite{howard2019imagenette}   & Tench            & \makecell{English\\Springer} & \makecell{Cassette\\Player} & Chainsaw           & Church         & French Horn & \makecell{Garbage\\Truck}   & Gas Pump    & Golf Ball          & Parachute          \\\hline
ImageWoof  \cite{howard2019imagenette}     & \makecell{Australian \\ Terrier}        & Border Terrier       & Samoyed     & Beagle        & Shih-Tzu   & \makecell{English\\Foxhound}        & \makecell{Rhodesian\\Ridgeback}           & Dingo     & Golden Retriever       & \makecell{English\\Sheepdog}             \\\hline
ImageSqauwk \cite{cazenavette2022dataset}     & Peacock          & Flamingo              & Macaw        & Pelican             & \makecell{King\\Penguin}     & Bald Eagle     & Toucan      & Ostrich    & Black Swan              & Cockatoo           \\\hline
\end{tabular}
\caption{Class listings for our ImageNet subsets.}
\label{tab:classes}
\end{table*}
\subsubsection{Data Preprocessing} \label{preprocess}
We implemented a standardized preprocessing approach for all datasets, following the methodology outlined in \cite{zhao2021datasetDSA}. To ensure optimal model performance during both training and evaluation, we utilized several popular transformations, including color jittering, cropping, cutout, scaling, and rotation, as differentiable augmentation strategies across all datasets. For the CIFAR10/100 datasets, we additionally applied Kornia zero-phase component analysis (ZCA) whitening, using the same setting as \cite{cazenavette2022dataset}. However, we refrained from using ZCA preprocessing for the medium- and high-resolution datasets due to the computational expense of the full-size ZCA transformation. As a result, the distilled images for these datasets display checkboard artifacts (see Figures \ref{fig:tinyimagenetipc1}, \ref{fig:imagenetipc1}, \ref{fig:imagenetteipc10}, \ref{fig:imagenetwoofipc10}, and \ref{fig:imagenetsquackipc10}). It is worth noting that we visualized the distilled images by directly applying the reverse transformation based on the corresponding data preprocessing without any further modifications.
\subsubsection{Implementations of Prior Works} \label{prior}
To ensure fair comparisons with prior works, we obtained publicly available distilled data for each baseline method and trained models using our experimental setup. We utilized the same ConvNet architecture with three, four, or five layers, depending on the image resolutions, and applied the same preprocessing technique across all methods. In cases where our results were comparable or inferior to those reported in the original papers, we presented their default numbers directly. Regarding the Kernel Inducing Points (KIP) method \cite{nguyen2021dataset2, nguyen2021dataset}, we made a slight modification by employing a 128-kernel ConvNet instead of the original 1024-kernel version. We did our best to reproduce prior methods that did not conduct experiments on some datasets by following the released author codes. However, for methods that encountered scalability issues on high-resolution datasets, we were unable to obtain the relevant performance scores.
\subsubsection{Hyperparameters} \label{hyperparam}
\begin{table*}
    \centering
    \resizebox{\textwidth}{!}{
    \normalsize{
    \begin{tabular}{c|c|c||c||c}
    \hline  
    \multicolumn{3}{c||}{\textbf{Hyperparameters}} &
    \multirow{2}{*}{} \textbf{Options/} &
    \multirow{2}{*}{\textbf{Value}} \\
    \cline{1-3}
    \textbf{Category} & \textbf{Parameter Name} & \textbf{Description} & \textbf{Range} & \\
    \hline 
    \hline
    
    \multirow{12}{*}{\textbf{Optimization}}
    & \textbf{Learning Rate $\boldsymbol{\eta_{\mathcal{S}}}$ (images) } & Step size towards global/local minima & $(0, 10.0]$ & IPC $\leq$ 50: $ 1.0$ \\ 
    & & &  & IPC $>$ 50: $10.0$\\
    \cline{2-5}

    & \textbf{Learning Rate $\boldsymbol{\eta_{\mathcal{\boldsymbol{\theta}}}}$ (network)} & Step size towards global/local minima & $(0, 1.0]$ & $0.01$ \\ 
    \cline{2-5}
    
    \multirow{4}{*}{} & 
    
    \multirow{2}{*}{\textbf{Optimizer (images)}} & \multirow{2}{*}{Updates synthetic set to approach global/local minima} & SGD with & Momentum: $0.5$ \\
    & & & Momentum & Weight Decay: $0.0$\\
    \cline{2-5}
    & \multirow{2}{*}{\textbf{Optimizer (network)}} & \multirow{2}{*}{Updates model to approach global/local minima} & SGD with & Momentum: $0.9$ \\
    & & & Momentum & Weight Decay: $5e-4$\\
    
    \cline{2-5}
    
    \multirow{3}{*}{} & 

    \multirow{1}{*}{\textbf{Scheduler (images)}} & - & - & - \\
    \cline{2-5}
    &\multirow{2}{*}{\textbf{Scheduler (network)}} & \multirow{2}{*}{Decays the learning rate over epochs} & \multirow{2}{*}{StepLR} & Decay rate: $0.5$ \\
    &&&& Step size: $15.0$\\
    \cline{2-5}
     & \textbf{Iteration Count} & Number of iterations for learning synthetic data & $[1, \infty)$ & 8000\\
    \hline
    \multirow{5}{*}{\textbf{Loss Function}}
    & \multirow{2}{*}{\textbf{Task Balance $\lambda$}} & \multirow{2}{*}{Regularization Multiplier} & \multirow{2}{*}{$[0, \infty)$} & Low Resolution: $0.01$ \\
    &&&& High Resolution: $0.02$\\
    \cline{2-5}
    &  \textbf{Power Value} \bf{$p$} & Exponential power for amplification in the SAM module & $[1, \infty)$ & 4 \\
    \cline{2-5}
    & \textbf{Loss Configuration} & Type of error function used to measure distribution discrepancy & - & Mean Squared Error \\
    \cline{2-5}
    & \textbf{Normalization Type} & Type of normalization used in the SAM module on attention maps & - & L2 \\
    \cline{2-5}
    
    
    \hline
    
    
    \multirow{8}{*}{\textbf{DSA Augmentations}}
    & \multirow{3}{*}{\textbf{Color}} &  \multirow{3}{*}{Randomly adjust (jitter) the color components of an image} & brightness & 1.0\\
    & & & saturation & 2.0\\
    & & & contrast & 0.5\\
    \cline{2-5}
    
     & \textbf{Crop} & Crops an image with padding & ratio crop pad & 0.125 \\
     \cline{2-5}
     & \textbf{Cutout} & Randomly covers input with a square & cutout ratio & 0.5 \\
    \cline{2-5}
    & \textbf{Flip} & Flips an image with probability p in range: & $(0, 1.0]$ & $0.5$ \\
    \cline{2-5}
    & \textbf{Scale} & Shifts pixels either column-wise or row-wise & scaling ratio & $1.2$ \\
    \cline{2-5}
    & \textbf{Rotate} & Rotates image by certain angle & $0^{\circ} - 360^{\circ}$ & $[-15^{\circ}, +15^{\circ}]$ \\
    \cline{2-5}

    \hline
    
    
    
    
    \multirow{3}{*}{\textbf{Encoder Parameters}} & \textbf{Conv Layer Weights} & The weights of convolutional layers & $\mathbb{R}$ bounded by kernel size & Uniform Distribution  \\
    \cline{2-5}
    
    & \textbf{Activation Function} & The non-linear function at the end of each layer & - & ReLU \\
    \cline{2-5}
    & \textbf{Normalization Layer} & Type of normalization layer used after convolutional blocks & - & InstanceNorm\\
    \hline
    \end{tabular}
    }       }
    \caption{Hyperparameters Details.}
    \label{tab: hyperparameters}
\end{table*}
In order to ensure that our methodology can be reproduced, we have included a Table \ref{tab: hyperparameters} listing all the hyperparameters used in this work. For the baseline methods, we utilized the default parameters that the authors specified in their original papers. We used the same hyperparameter settings across all experiments unless otherwise stated. Specifically, we employed an SGD optimizer with a learning rate of 1 for learning synthetic sets and a learning rate of 0.01 for training neural network models. For low-resolution datasets, we used a 3-layer ConvNet, while for medium- and high-resolution datasets, we followed the recommendation of \cite{zhao2023dataset} and used a 4-layer and 5-layer ConvNet, respectively. In all experiments, we used a mini-batch of 256 real images from each class to learn the synthetic set. Additionally, we conducted ablation studies on certain hyperparameters, such as task balance $\lambda$ and the power parameter $p$ in the Spatial Attention Matching (SAM) modules, which are discussed in Section \ref{ablation}.


\subsection{Additional Results and Further Analysis} \label{add-results}

\subsubsection{Comparison to More Baselines} \label{baselines}
We conducted a comparison between images created by the DataDAM and popular generative models such as variational auto-encoders (VAEs) \cite{kingma2013auto, parmar2021dual} and generative adversarial networks (GANs) \cite{goodfellow2020generative, mirza2014conditional, brocklarge, li2015generative} to evaluate their data efficiency. For this purpose, we selected state-of-the-art models, including the DC-VAE \cite{parmar2021dual}, cGAN \cite{mirza2014conditional}, BigGAN \cite{brocklarge}, and GMMN \cite{li2015generative}. The DC-VAE generates a model with dual contradistinctive losses, which improves the generative autoencoder's inference and synthesis abilities simultaneously. The cGAN model is conditioned on both the generator and discriminator, while BigGAN uses differentiable augmentation techniques \cite{zhao2021datasetDSA}. On the other hand, GMMN aims to learn an image generator that can map a uniform distribution to a real image distribution. We trained these models on the CIFAR10 dataset with varying numbers of images per class (1, 10, and 50 IPCs) using ConvNet's (3-layer) architecture \cite{gidaris2018dynamic} and evaluated their performance on real testing images. Our results, presented in Table \ref{tab:vae_gan}, indicate that our proposed method significantly outperforms these generative models. The DataDAM generates superior training images that offer more informative data for training DNNs, while the primary goal of the generative models is to create realistic-looking images that can deceive humans. Therefore, the efficiency of images produced by generative models is similar to that of randomly selected coresets.

We also employed another baseline approach, which is learning synthetic images through distribution matching using vanilla maximum mean discrepancy \cite{gretton2012kernel} (MMD) in the pixel space. By utilizing MMD loss with a linear kernel, we achieved improved performance compared to randomly selected real images and generative models (see Table \ref{tab:vae_gan}). However, DataDAM surpasses the results of vanilla MMD since it generates more informative synthetic images by utilizing the information of the feature extractor at various levels of representation.
\begin{table}[H]
\scriptsize
\setlength{\tabcolsep}{2pt}
\renewcommand\arraystretch{0.9}
\begin{tabular}{cccccccc}
\toprule
IPC & Random        & DC-VAE    &     cGAN      & BigGAN        & GMMN   & MMD   & {DataDAM} \\ \midrule
1       & 14.4$\pm$2.0  & 15.7$\pm$2.1   & 16.3$\pm$1.4 & 15.8$\pm$1.2  & 16.1$\pm$2.0 & 22.7$\pm$0.6  & \bf{32.0$\pm$1.2}     \\
10      & 26.0$\pm$1.2  & 29.8$\pm$1.0   & 27.9$\pm$1.1 & 31.0$\pm$1.4  & 32.2$\pm$1.3  & 34.9$\pm$0.3  & \bf{54.2$\pm$0.8}     \\
50      & 43.4$\pm$1.0  & 44.0$\pm$0.8  & 43.8$\pm$0.9 & 46.2$\pm$0.9  &  45.3$\pm$1.0 & 50.9$\pm$0.3  & \bf{67.0$\pm$0.4}     \\ \bottomrule
\end{tabular}
\caption{Comparison of the DataDAM's performance to popular generative models and the MMD baseline on the CIFAR10 dataset using ConvNets. The "Random" category denotes randomly selected real images.}
\label{tab:vae_gan}
\end{table}

\subsubsection{More Ablation Studies} \label{ablation}

\begin{figure}
    \centering
    \includegraphics[width=0.48\textwidth]{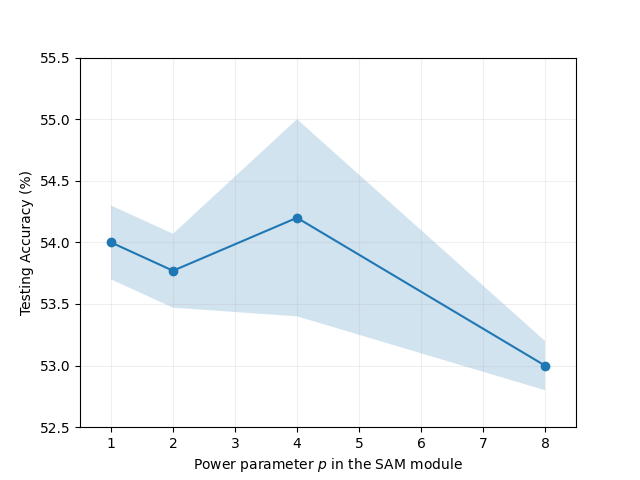}
    \caption{The effect of the power parameter $p$ on the final testing accuracy (\%) for the CIFAR10 dataset with IPC 10 configuration.}
    \label{pvalue}
\end{figure}

\textbf{Evaluation of power parameter $\boldsymbol{p}$ in the SAM module.}
This section examines how the $p$-norm impacts the efficiency of spatial-wise attention maps in the SAM module. In Figure \ref{pvalue}, we evaluate the testing accuracy of the DataDAM on CIFAR10 with IPC 10 for various values of $p$. Our method proves to be robust across a broad range of $p$ values, indicating that it is not significantly affected by changes in the degree of discrepancy measured by $\mathcal{L}_{\text{SAM}}$. However, when the power is raised to $8$, the DataDAM gives more weight to spatial locations that correspond to the neurons with the highest activations. In other words, it prioritizes the most discriminative parts, potentially ignoring other important components that may be crucial in approximating the data distribution. This could negatively impact the testing performance to some extent.

\begin{figure}
    \centering
    \includegraphics[width=0.48\textwidth]{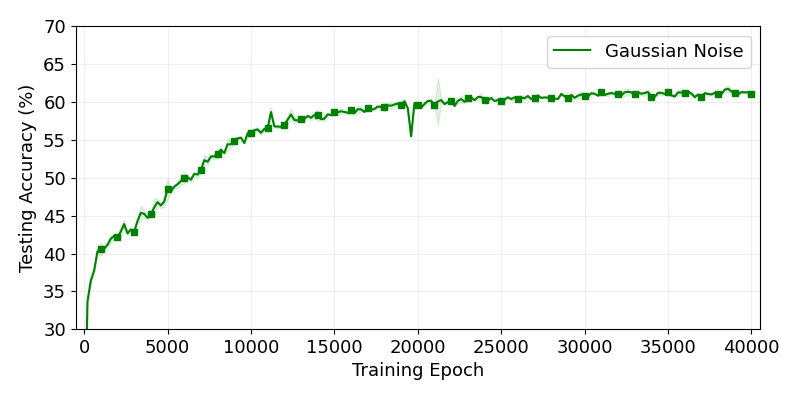}
    \caption{Test accuracy evolution of synthetic image learning on CIFAR10 with IPC 50 under Gaussian noise initialization.}
    \label{fig:noise}
\end{figure}

\begin{figure*}[htbp]
  \centering
  \begin{subfigure}[b]{0.12\textwidth}
    \includegraphics[width=\linewidth]{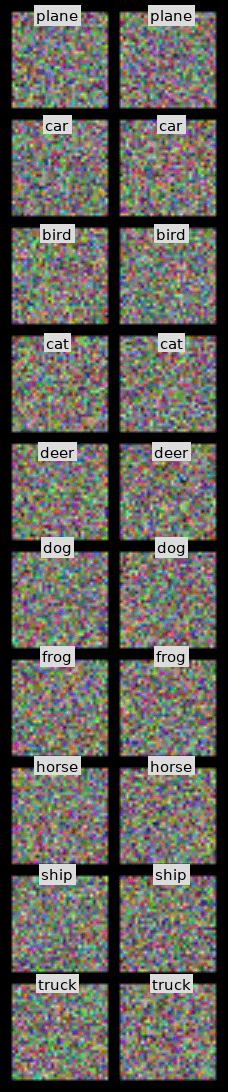}
    \caption{Iteration \\0}
    \label{fig:sub1}
  \end{subfigure}\hfill
  \begin{subfigure}[b]{0.12\textwidth}
    \includegraphics[width=\linewidth]{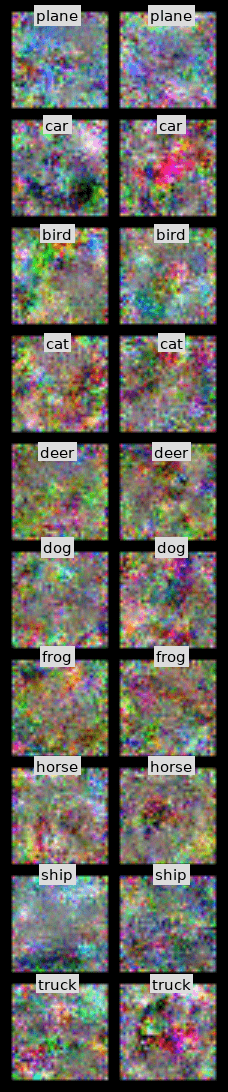}
    \caption{Iteration \\600}
    \label{fig:sub2}
  \end{subfigure}\hfill
  \begin{subfigure}[b]{0.12\textwidth}
    \includegraphics[width=\linewidth]{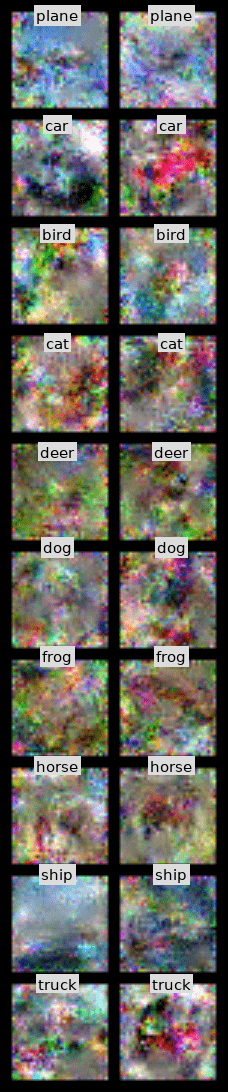}
    \caption{Iteration \\2000}
    \label{fig:sub1}
  \end{subfigure}\hfill
  \begin{subfigure}[b]{0.12\textwidth}
    \includegraphics[width=\linewidth]{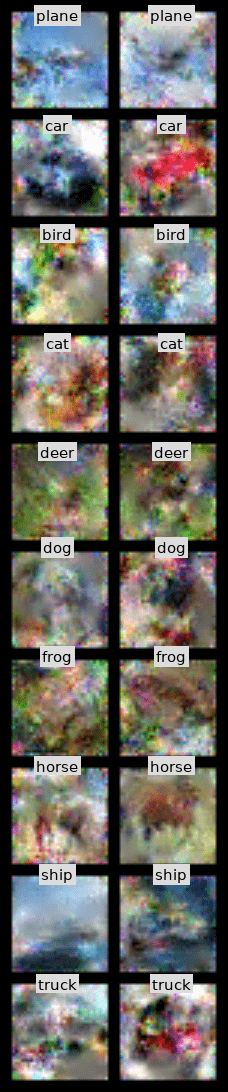}
    \caption{Iteration \\5000}
    \label{fig:sub1}
  \end{subfigure}\hfill
  \begin{subfigure}[b]{0.12\textwidth}
    \includegraphics[width=\linewidth]{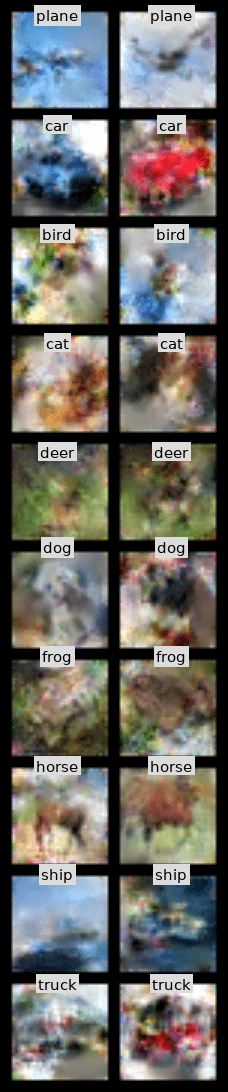}
    \caption{Iteration \\15000}
    \label{fig:sub1}
  \end{subfigure}\hfill
    \begin{subfigure}[b]{0.12\textwidth}
    \includegraphics[width=\linewidth]{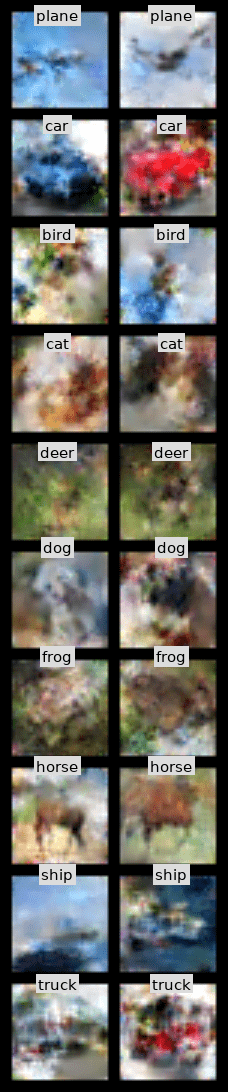}
    \caption{Iteration \\25000}
    \label{fig:sub1}
  \end{subfigure}\hfill
  \begin{subfigure}[b]{0.12\textwidth}
    \includegraphics[width=\linewidth]{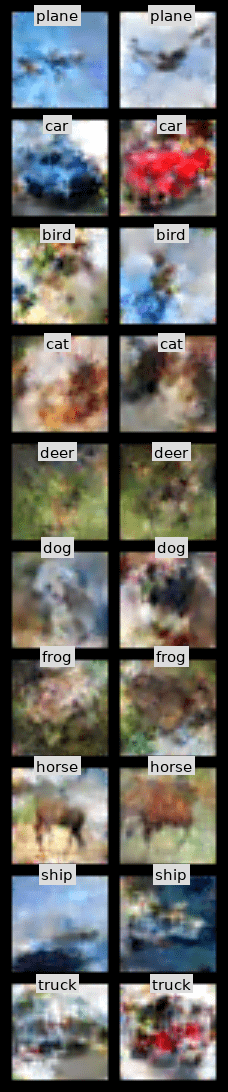}
    \caption{Iteration \\35000}
    \label{fig:sub1}
  \end{subfigure}\hfill
  \begin{subfigure}[b]{0.12\textwidth}
    \includegraphics[width=\linewidth]{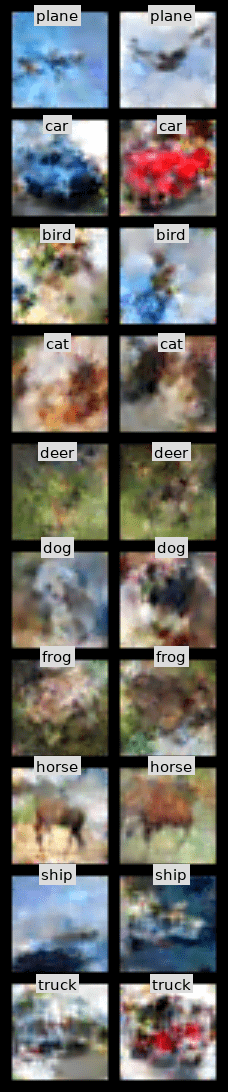}
    \caption{Iteration \\40000}
    \label{fig:sub1}
  \end{subfigure}\hfill
  \caption{The learning process of all classes in the CIFAR10 dataset (IPC 50) initialized from Gaussian noise. We take two random images for each class and visualize their progression over the 40,000 training epochs.}
  \label{fig:noiseiter}
\end{figure*}

\textbf{Exploring the effect of Gaussian noise initialization for synthetic images on DataDAM.}
To augment our results in the main paper, we present an extended training configuration for initialization from Gaussian noise. We conducted this experiment on CIFAR10 with IPC 50. As seen in Figure \ref{fig:noise}, the Gaussian noise initialization scheme takes longer to converge to a competitive accuracy level. Despite underperforming in comparison to Random and K-Center initialization, it still demonstrates the ability of our proposed method to distill information from the real dataset onto pure random noise. Moreover, it is capable of outperforming competitive methods, particularly KIP \cite{nguyen2021dataset2} and DSA \cite{zhao2021datasetDSA}. In Figure \ref{fig:noiseiter}, we provide visualizations of the synthetic data generated from random noise during different iterations. These visualizations highlight how our method successfully transfers information from the real dataset to the random noise, especially when comparing the initial noise image with the final iteration.


\begin{figure}
    \centering
    \includegraphics[width=0.45\textwidth]{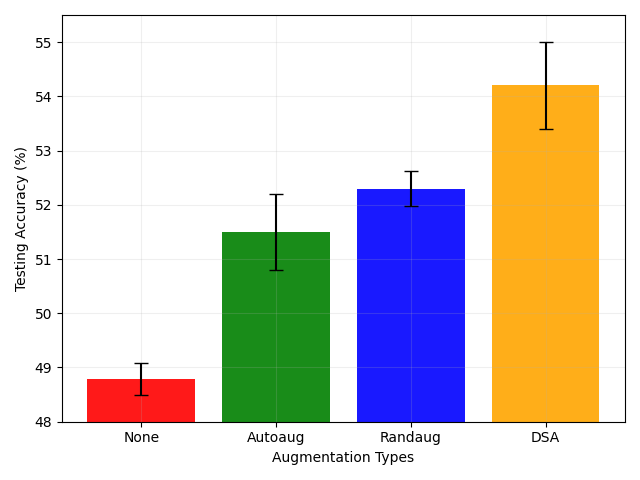}
    \caption{The effect of different augmentation strategies during the evaluation phase on the final testing accuracy (\%) for the CIFAR10 dataset with IPC 10 configuration.}
    \label{fig:augment}
\end{figure}

\textbf{Exploring the effect of different augmentation strategies in DataDAM.}
In this section, we explore the impact of augmentation methods on the effectiveness of our approach when evaluated on the CIFAR10 dataset with an IPC 10 configuration. We treat our method as a black box, as in the work of \cite{cuidc}, and assess the effects of various augmentation techniques such as AutoAugment \cite{cubuk2018autoaugment}, RandAugment \cite{cubuk2020randaugment}, DSA \cite{zhao2021datasetDSA}, and no augmentation on the distilled datasets during the evaluation phase. The results are presented in Figure \ref{fig:augment}. Our observations reveal that DSA delivers significantly better performance as it is integrated into the training process of the synthetic dataset and is more compatible with the learning phase of the distilled images. Additionally, our findings indicate that augmentation is vital for training on synthetic data, as evidenced by the substantial differences between different augmentation methods and no augmentation. Therefore, applying augmentation techniques to our distilled images during evaluation can substantially enhance model performance.

\begin{figure}
    \centering
    \includegraphics[width=0.45\textwidth]{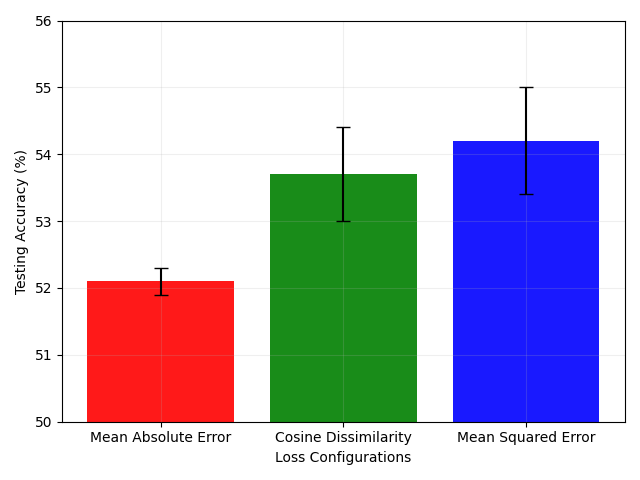}
    \caption{The effect of loss configurations of $\mathcal{L}_{\text{SAM}}$ on the final testing accuracy (\%) for the CIFAR10 dataset with IPC 10 configuration.}
    \label{fig:lossconf}
\end{figure}

\textbf{Exploring the effect of different loss configurations in $\mathcal{L}_{\text{SAM}}$.}
In this section, we explore the impact of different loss configurations on attention loss ($\mathcal{L}_{\text{SAM}}$). To conduct this evaluation, we employed mean absolute error (MAE), cosine dissimilarity, and mean square error (MSE) as objective functions for $\mathcal{L}_{\text{SAM}}$ to train a synthetic dataset on CIFAR10 with IPC 10. The results presented in Figure \ref{fig:lossconf} demonstrate that MSE yields the best results. Nonetheless, it is crucial to note that even with any of these configurations, our method still outperforms most of the competitive methods. Therefore, we can conclude that our approach performs well with any loss configuration, but a well-designed configuration can result in a substantial performance improvement of up to $2.0 \%$ in our ablation study.

\begin{figure}
    \centering
    \includegraphics[width=0.45\textwidth]{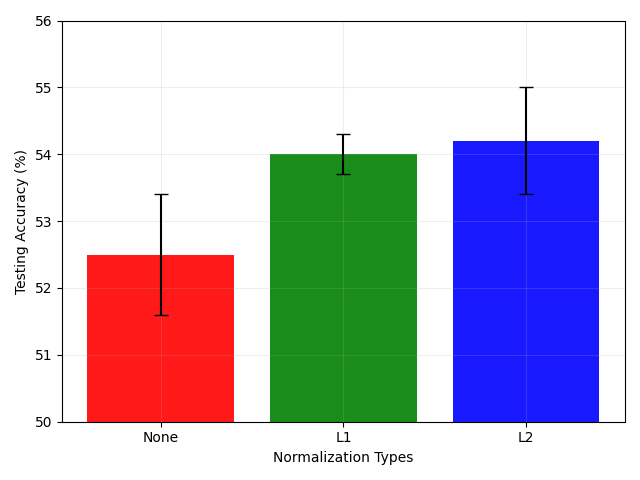}
    \caption{The effect of different normalization blocks of the SAM module on the final testing accuracy (\%) for the CIFAR10 dataset with IPC 10 configuration.}
    \label{fig:norm}
\end{figure}

\textbf{Exploring the effect of normalization in the SAM module.}
In this section, we aim to evaluate the impact of the normalization block in the internal structure of the SAM module on testing accuracy. We conducted experiments by training distilled images for CIFAR10 with IPC 10 and testing three normalization techniques: $L_{1}$ normalization, $L_{2}$ normalization, and no normalization. The results, as shown in Figure \ref{fig:norm}, indicate that $L_{2}$ normalization is the most effective in terms of testing accuracy. By adding normalization, we reduce the magnitude of the attention loss $\mathcal{L}_{\text{SAM}}$ in backpropagation, thus decreasing the chance of overshooting the global minima in the optimization space when modifying the input image's pixels. We can observe that both normalization schemes work well, but the absence of normalization leads to significant performance degradation. Therefore, we conclude that while the appropriate use of normalization is critical for the performance of the DataDAM, the type of normalization is not as significant.



\subsubsection{More Experiments and Analysis on Neural Architecture Search} \label{nas}
Taking inspiration from \cite{zhao2021datasetDC, zhao2021datasetDSA, zhao2023dataset}, we define a search space consisting of 720 ConvNets on the CIFAR10 dataset. We evaluate the models using our distilled data with IPC 50 as a proxy set under the neural architecture search (NAS) framework. We start with a base ConvNet and construct a uniform grid that varies in depth $D \in$ \{1, 2, 3, 4\}, width $W \in$ \{32, 64, 128, 256\}, activation function $A \in$ \{\text{Sigmoid, ReLu, LeakyReLu}\}, normalization technique $N \in$ \{None, BatchNorm, LayerNorm, InstanceNorm, GroupNorm\}, and pooling operation $P \in$ \{None, MaxPooling, AvgPooling\}. These candidates are then evaluated based on their validation performance and ranked accordingly.

\begin{figure*}
  \centering
  \begin{subfigure}[b]{0.32\textwidth}
    \includegraphics[width=\linewidth]{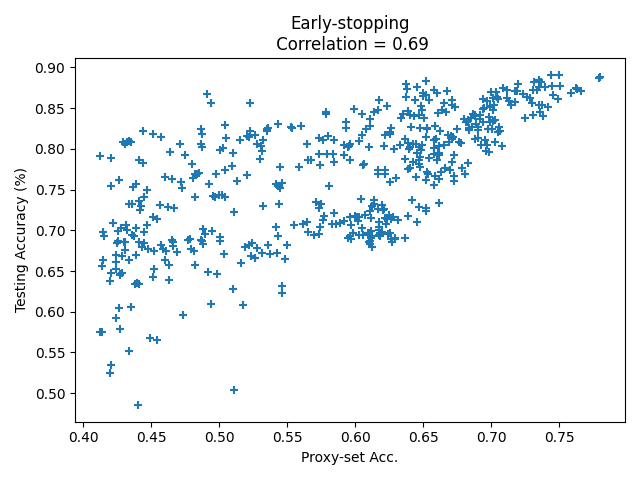}
    \caption{EarlyStop}
    \label{fig:sub1}
  \end{subfigure}\hfill
  \begin{subfigure}[b]{0.32\textwidth}
    \includegraphics[width=\linewidth]{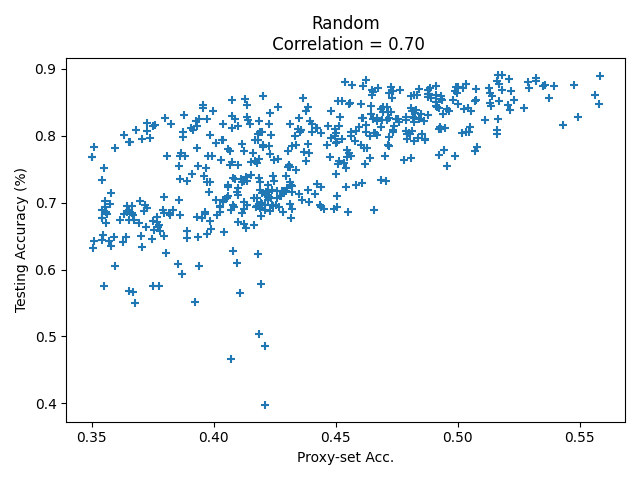}
    \caption{Random}
    \label{fig:sub2}
  \end{subfigure}\hfill
  \begin{subfigure}[b]{0.32\textwidth}
    \includegraphics[width=\linewidth]{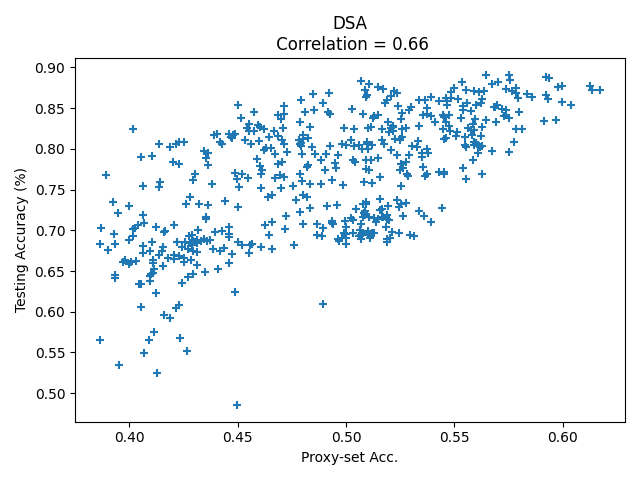}
    \caption{DSA}
    \label{fig:sub2}
  \end{subfigure}\hfill
    \begin{subfigure}[b]{0.32\textwidth}
    \includegraphics[width=\linewidth]{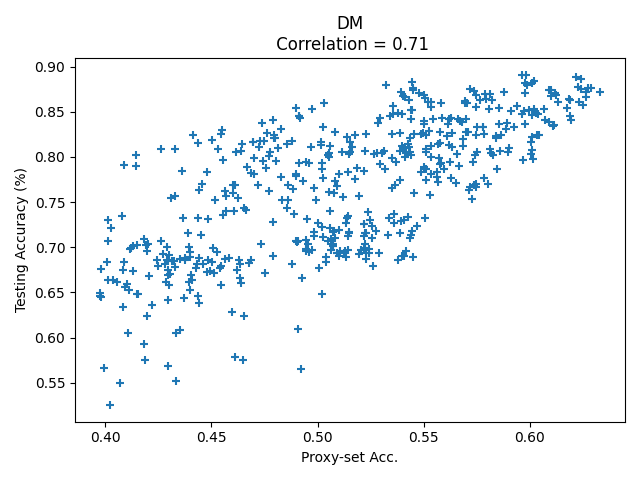}
    \caption{DM}
    \label{fig:sub2}
  \end{subfigure}\hfill
    \begin{subfigure}[b]{0.32\textwidth}
    \includegraphics[width=\linewidth]{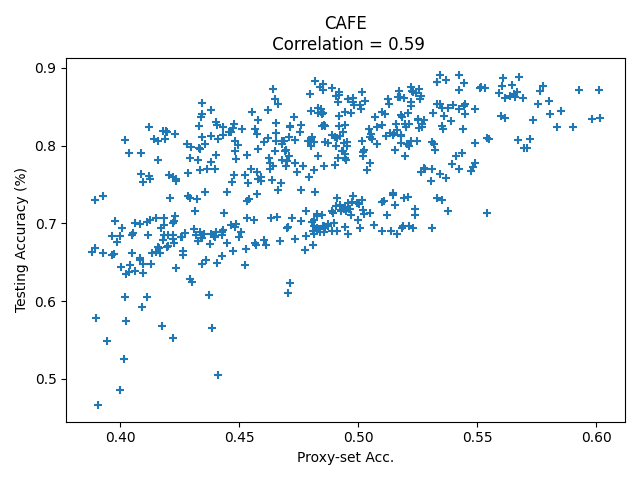}
    \caption{CAFE}
    \label{fig:sub2}
  \end{subfigure}\hfill
    \begin{subfigure}[b]{0.32\textwidth}
    \includegraphics[width=\linewidth]{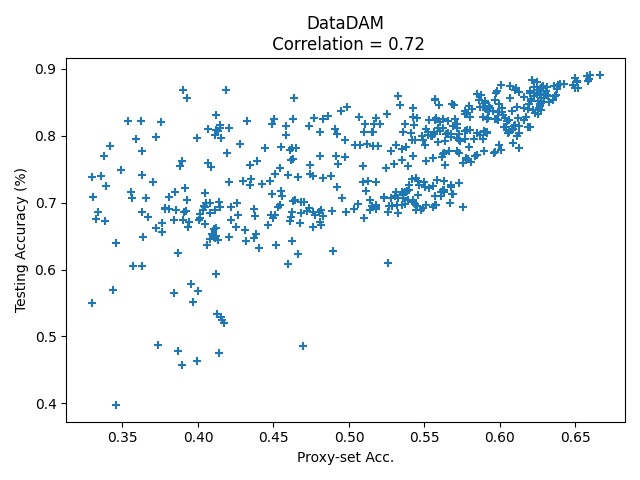}
    \caption{DataDAM}
    \label{fig:sub2}
  \end{subfigure}\hfill
  \caption{Performance rank correlation between proxy set and whole dataset training across all 720 architectures.}
  \label{fig:nasfull}
\end{figure*}

\begin{figure*}[t]
  \centering
  \begin{subfigure}[b]{0.32\textwidth}
    \includegraphics[width=\linewidth]{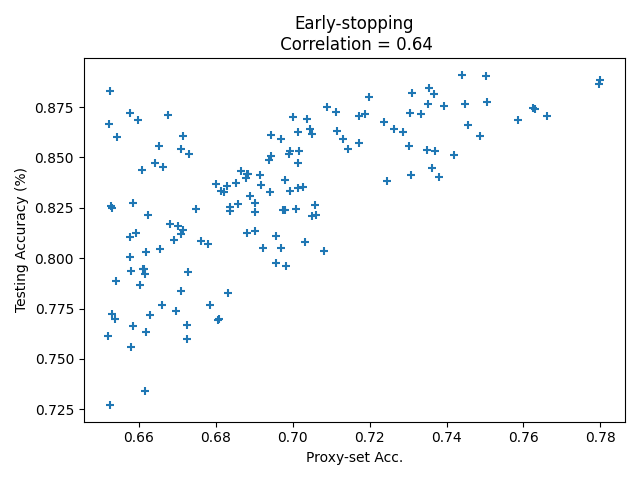}
    \caption{EarlyStop}
    \label{fig:sub1}
  \end{subfigure}\hfill
  \begin{subfigure}[b]{0.32\textwidth}
    \includegraphics[width=\linewidth]{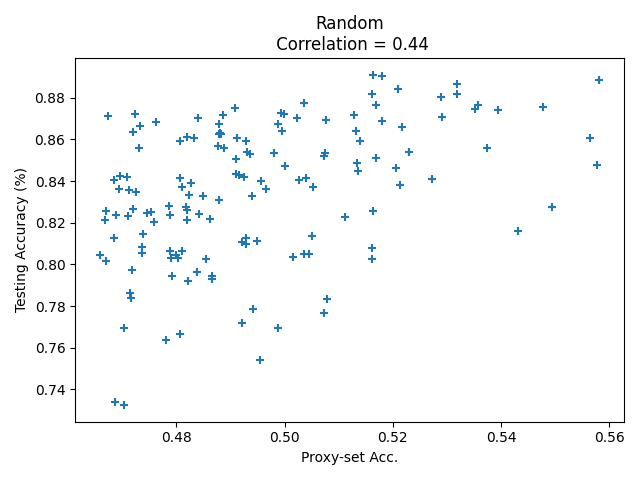}
    \caption{Random}
    \label{fig:sub2}
  \end{subfigure}\hfill
  \begin{subfigure}[b]{0.32\textwidth}
    \includegraphics[width=\linewidth]{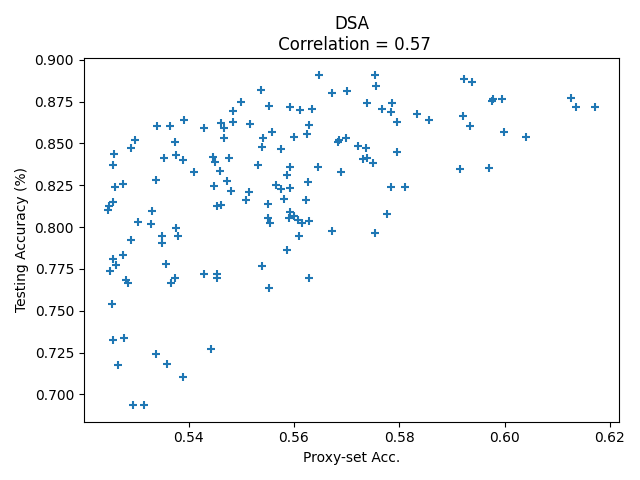}
    \caption{DSA}
    \label{fig:sub2}
  \end{subfigure}\hfill
    \begin{subfigure}[b]{0.32\textwidth}
    \includegraphics[width=\linewidth]{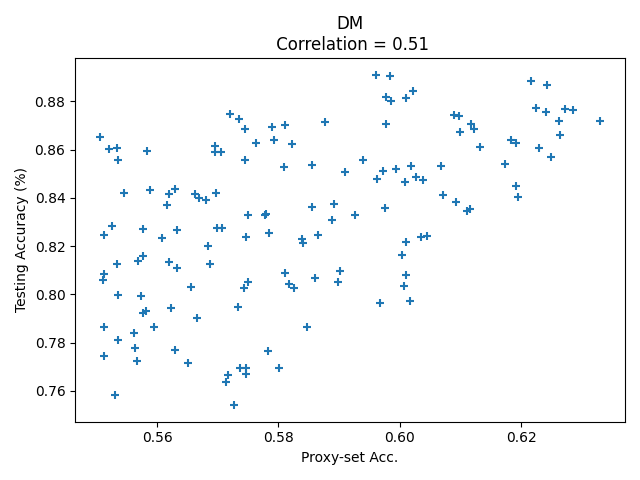}
    \caption{DM}
    \label{fig:sub2}
  \end{subfigure}\hfill
    \begin{subfigure}[b]{0.32\textwidth}
    \includegraphics[width=\linewidth]{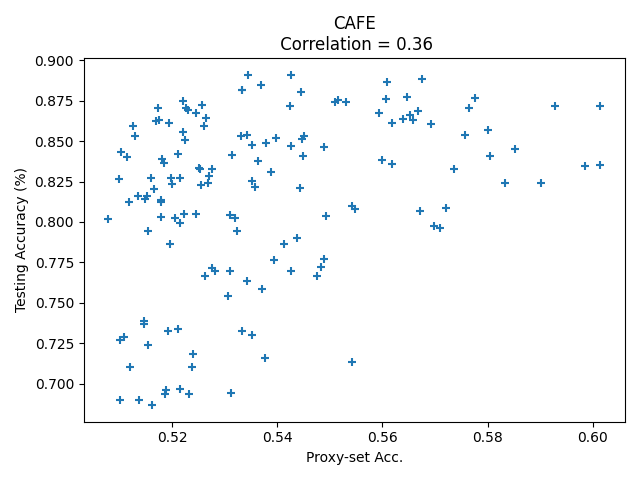}
    \caption{CAFE}
    \label{fig:sub2}
  \end{subfigure}\hfill
    \begin{subfigure}[b]{0.32\textwidth}
    \includegraphics[width=\linewidth]{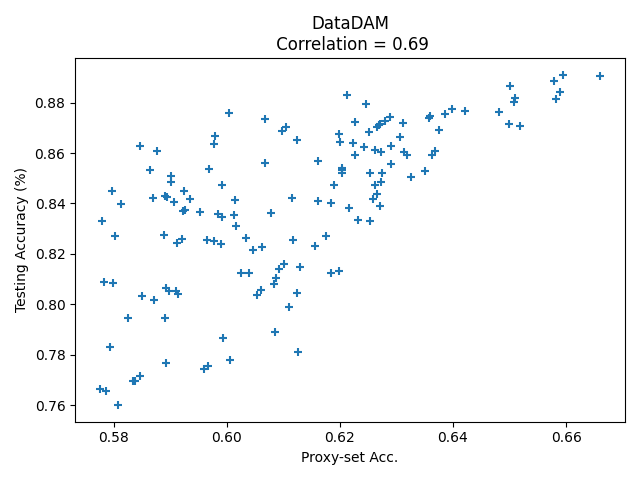}
    \caption{DataDAM}
    \label{fig:sub2}
  \end{subfigure}\hfill
  \caption{Performance rank correlation between proxy-set and whole-dataset training across the top 20\% of the search space (selecting 144 architectures with the highest validation accuracy).}
  \label{fig:nas20}
\end{figure*}

In Figure \ref{fig:nasfull}, we displayed the performance rank correlation between the proxy set, generated using various methods, and the whole training dataset using Spearman’s correlation across all 720 architectures. Each point in the graph represents a selected architecture. The x-axis represents the test accuracy of the model trained on the proxy set, while the y-axis represents the accuracy of the model trained on the whole dataset. Our analysis shows that all methods perform well. However, DataDAM has a higher concentration of dots close to the straight line, indicating a better proxy set for obtaining more reliable performance rankings of candidate architectures. These results are on par with the DataDAM’s performance correlation (0.72), which is higher than other prior works. To further assess the effectiveness of our approach, we conducted an analysis of the top 20\% of the search space, selecting 144 architectures with the highest validation accuracy. As depicted in Figure \ref{fig:nas20}, our method outperforms most of the state-of-the-art methods, except for early stopping, where we only beat it by a small margin. Our evaluation of the correlation graphs indicates that DataDAM is capable of accurately correlating the performance of models trained on the proxy dataset with their performance on the whole training dataset. We substantiate these findings by presenting quantitative results of performance and Spearman's correlation in Table \ref{tab:nas20}.

\begin{table}[H]
\renewcommand\arraystretch{0.6}
\centering
\scriptsize
\setlength{\tabcolsep}{1.2pt}
\begin{tabular}{ccccccc|c}
\toprule
                    & Random            & DSA               & DM             & CAFE & Ours & Early-stopping & Whole Dataset \\ \midrule
Performance (\%)    &        88.9         &             87.2   &         87.2   &  83.6 &  \bf{89.0}  &   88.9 & 89.2  \\
Correlation Top 20\%    &        0.44         &        0.57        &       0.51     &  0.36 &  \bf{0.69}   &  0.64  &  1.00\\
Time cost (min)     & 33.0  &   31.2    &  32.2 &     \bf{30.7}  &   34.8  &   37.1    & 5168.9 \\ 
Storage (imgs)      & \bf{500}          & \bf{500}          & \bf{500}        &   \bf{500}  &   \bf{500} & $5\times 10^4$    & $5\times 10^4$   \\ \bottomrule
\end{tabular}
\caption{{Neural architecture search on CIFAR10 with a search space of the top 20\% of the sample space with the highest validation accuracy.}}
\label{tab:nas20}
\end{table}

\begin{figure*}
    \centering
    \includegraphics[width=\textwidth]{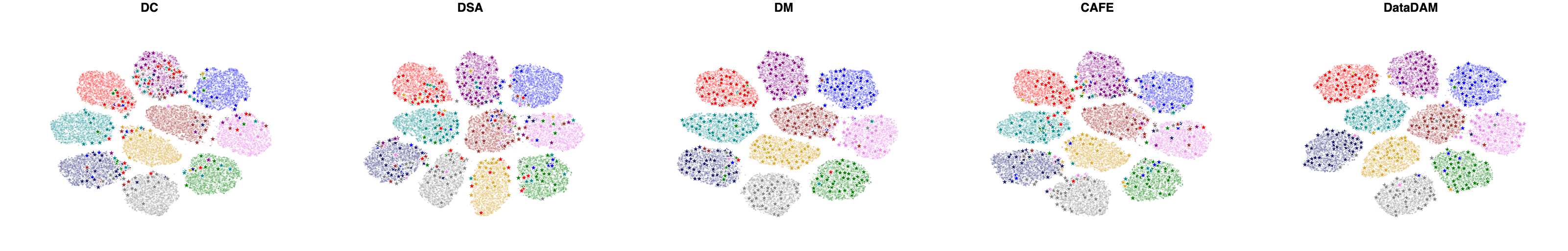}
    \caption{Distributions of the synthetic images learned by five methods on the CIFAR10 dataset with IPC 50. The stars represent the synthetic data dispersed amongst the original dataset. The classes are as follows: \color{green} plane, 
    \color{Goldenrod} car, 
    \color{blue} bird, 
    \color{violet} cat, 
    \color{MidnightBlue} deer, 
    \color{brown} dog, 
    \color{Mulberry} frog,
    \color{gray} horse,
    \color{Emerald} ship,
    \color{red} truck.}
    \label{fig:tsne}
\end{figure*}

\textbf{Experiments on NAS-Bench-201.}
To conduct a more comprehensive analysis of the neural architecture search, we expanded the search space by including NAS-Bench-201 \cite{dongbench} as recommended in \cite{cuidc}. Our aim is to compare the performance of DataDAM against other methods using the CIFAR10 dataset with IPC 50 as the proxy set. To create a search space, we randomly selected 100 networks from the 15,635 available models in NAS-Bench-201. We followed the configuration and settings presented in \cite{cuidc}, which involve training all models using five random seeds and ranking them based on their average accuracy on a validation set comprising 10,000 images. We used two metrics to evaluate the effectiveness of NAS: the performance correlation ranking between models trained on synthetic and real datasets and the top-1 performance in the search space. In contrast to the previous search space that concentrated on 720 ConvNet architectures, we observed a distinct trend in this larger NAS benchmark with modern architectures. According to Table \ref{tab:nas-bench201}, while most methods achieved negative correlations between performance on the proxy set and the entire dataset, our method had a small positive correlation and obtained competitive outcomes on the original dataset. This implies that DataDAM preserves the true strength of the underlying model more effectively than previous works. Nevertheless, despite the encouraging performance gains achieved by the best single model, utilizing the distilled data to guide model design remains a significant challenge. It is important to mention that the rank correlation presented in Table \ref{tab:nas-bench201} for the original real dataset is not 1.0. This is because a smaller architecture was used and the ranking was based on a validation set, as pointed out in \cite{cuidc}.
\begin{table}[H]
\vspace{-2.9mm}
    \centering
    \resizebox{0.48\textwidth}{!}{
    \begin{tabular}{cccccccc|c}
    \toprule
     & Random & 
     DC & DSA & DM & KIP & MTT & DataDAM & Whole Dataset\\ 
    \midrule
    Correlation & -0.06 & 
    -0.19 & -0.37 & -0.37 & -0.50& -0.09 & \bf{0.07} & 0.7487\\
    Top 1 (\%) & 91.9 & 
    86.44 & 73.54 & 92.16 & 92.91 & 73.54 & \bf{93.96} & 93.5\\
    \bottomrule
    \end{tabular}}
        \caption{Spearman’s rank correlation results were obtained using NAS-Bench-201. The best performance achieved on the test set is 94.36\% \cite{cuidc}.}
    \label{tab:nas-bench201}
    \vspace{-3mm}
\end{table}
\subsection{Additional Visualizations and Analysis} \label{vis}
\subsubsection{More Analysis on Data Distribution} \label{datadist}
\vspace{-0.98mm}
To complement the data distribution visualization results presented in the main paper, we have included t-SNE \cite{van2008visualizing} illustrations for all categories in Figure \ref{fig:tsne}. We utilized t-SNE to show the features of real and synthetic sets generated by DC \cite{zhao2021datasetDC}, DSA \cite{zhao2021datasetDSA}, DM \cite{zhao2023dataset}, CAFE \cite{wang2022cafe}, and DataDAM in the embedding space of the ResNet-18 \cite{he2016deep} architecture. The visualizations were applied to the CIFAR10 dataset with IPC 50 for all methodologies. As depicted in Figure \ref{fig:tsne}, our approach, similar to DM, preserves the distribution of data with a well-balanced spread over the entire dataset. Conversely, other methods, such as DC, DSA, and CAFE, exhibit a significant bias toward the boundaries of certain clusters and have high false-positive rates for the majority of the classes. To put it simply, the t-SNE visualization validates that our method maintains a considerable degree of impartiality in accurately capturing the dataset distribution uniformly across all categories.

\subsubsection{Extended Visualizations of Synthetic Images} \label{vis-syn}
\textbf{Visualization of the synthetic images trained with different model architectures in DataDAM.}
In this section, we present a qualitative comparison of the generated distilled images using different architectures to demonstrate how the choice of architecture influences the quality of the synthetic set. We assess the efficacy of the distilled data trained using ConvNet \cite{gidaris2018dynamic}, AlexNet \cite{krizhevsky2017imagenet}, and VGG-11 \cite{simonyan2014very} architectures on the CIFAR10 dataset with IPC 50. Our results, as depicted in Figure \ref{fig:synimages}, reveal that the distilled data can encode the inductive bias of the chosen architecture. Specifically, the distilled images produced by the simplest architecture, i.e., ConvNet \cite{gidaris2018dynamic}, exhibit a natural appearance and can transfer well to other architectures (see Table 3 of the main paper). In contrast, the distilled images generated by modern architectures like VGG-11 \cite{simonyan2014very} exhibit different brightness and contrast than natural images. We found that increasing the complexity and number of convolutional layers in the feature extraction process led to brighter and more contrasting distilled images. This is likely because the attention loss ($\mathcal{L}_{\text{SAM}}$) becomes more potent, resulting in a more substantial modulation effect on the input image pixels during backpropagation. This trend is noticeable in the distilled images generated by AlexNet \cite{krizhevsky2017imagenet} and VGG-11 \cite{simonyan2014very}. We note that the synthetic images may reflect the similarity between architectures, as evidenced by the similarity between the images produced by AlexNet and ConvNet. This finding suggests that the inductive biases of these two architectures are comparable.

\begin{figure*}[htbp]
  \centering
   \begin{subfigure}[b]{0.48\textwidth}
    \includegraphics[width=\linewidth]{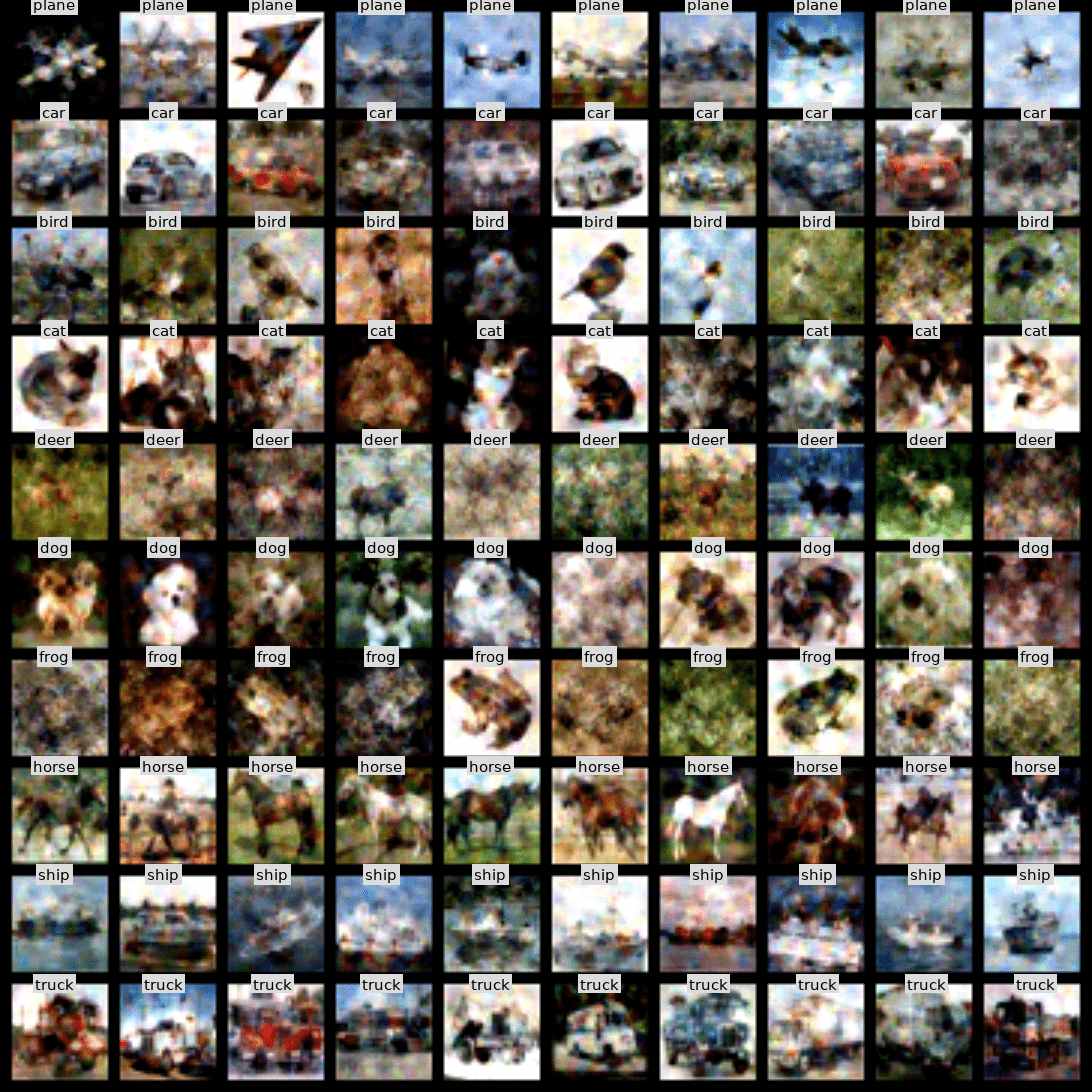}
    \caption{$\mathcal{L}_{\text{MMD}}$}
    \vspace{+5pt}
    \label{fig:sub4}
  \end{subfigure}\hfill
  \begin{subfigure}[b]{0.48\textwidth}
    \includegraphics[width=\linewidth]{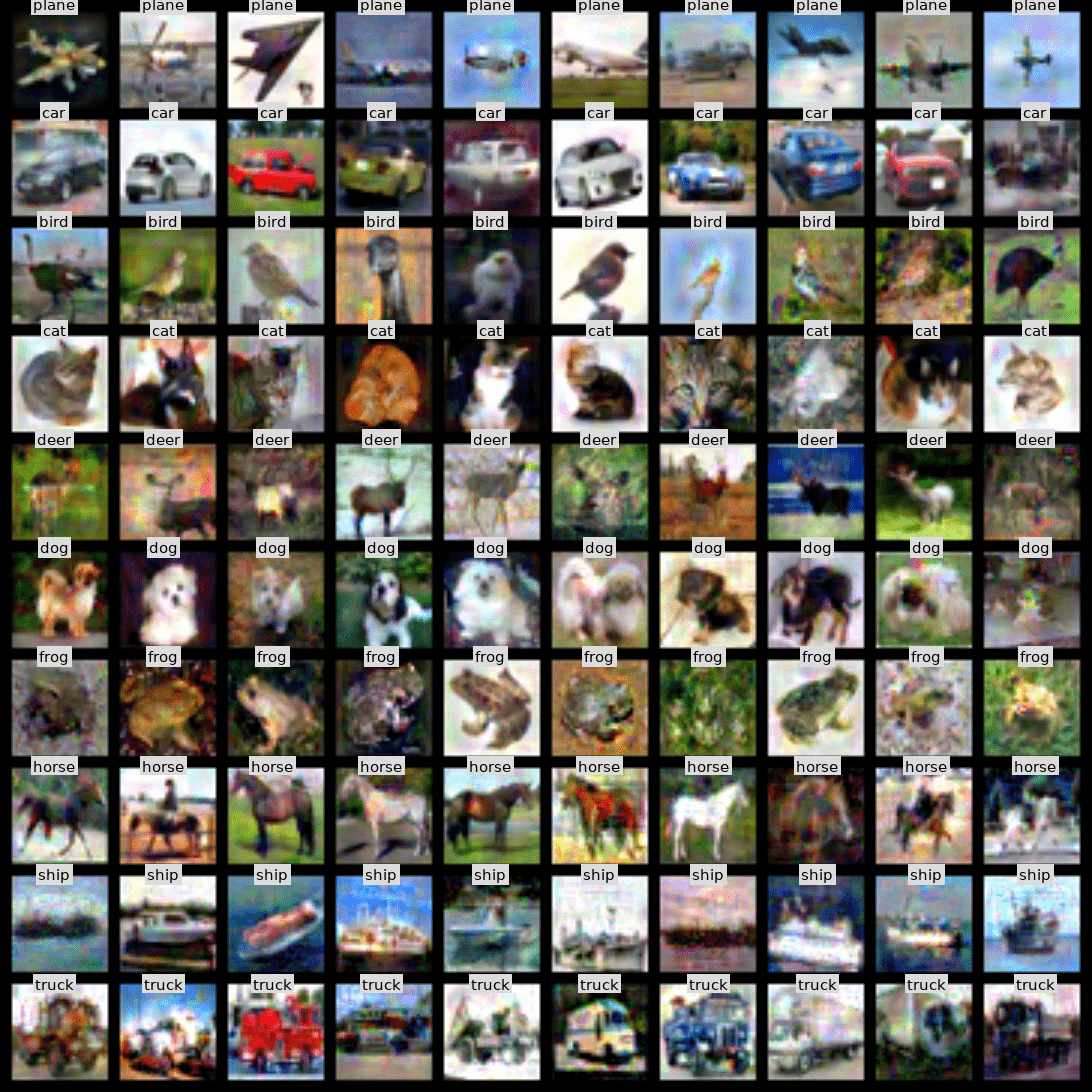}
    \caption{$\mathcal{L}_{\text{SAM}}$}
    \vspace{+5pt}
    \label{fig:sub1}
  \end{subfigure}\hfill
  \begin{subfigure}[b]{0.48\textwidth}
    \includegraphics[width=\linewidth]{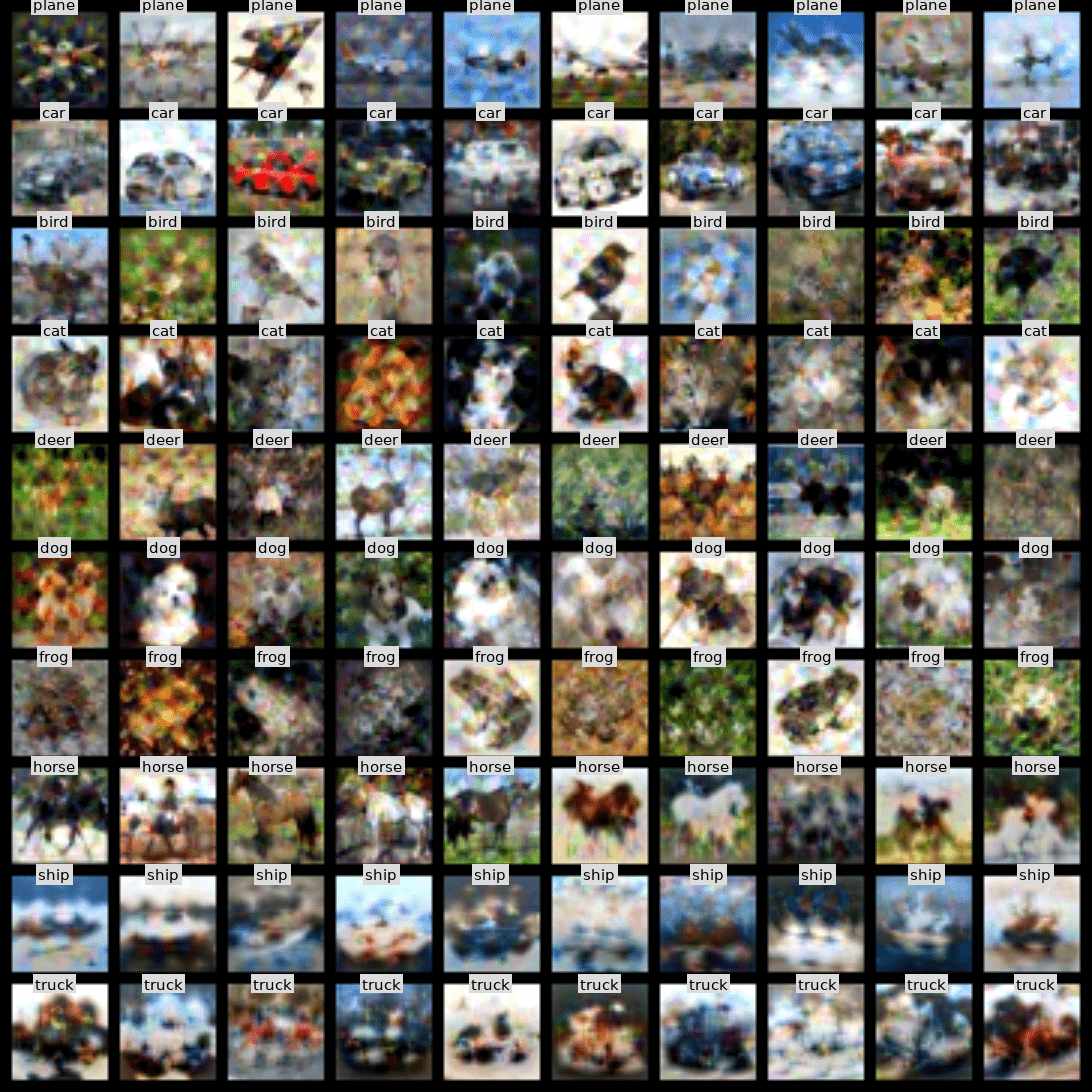}
    \caption{Feature Map Transfer Loss}
    \label{fig:sub2}
  \end{subfigure}\hfill
  \begin{subfigure}[b]{0.48\textwidth}
    \includegraphics[width=\linewidth]{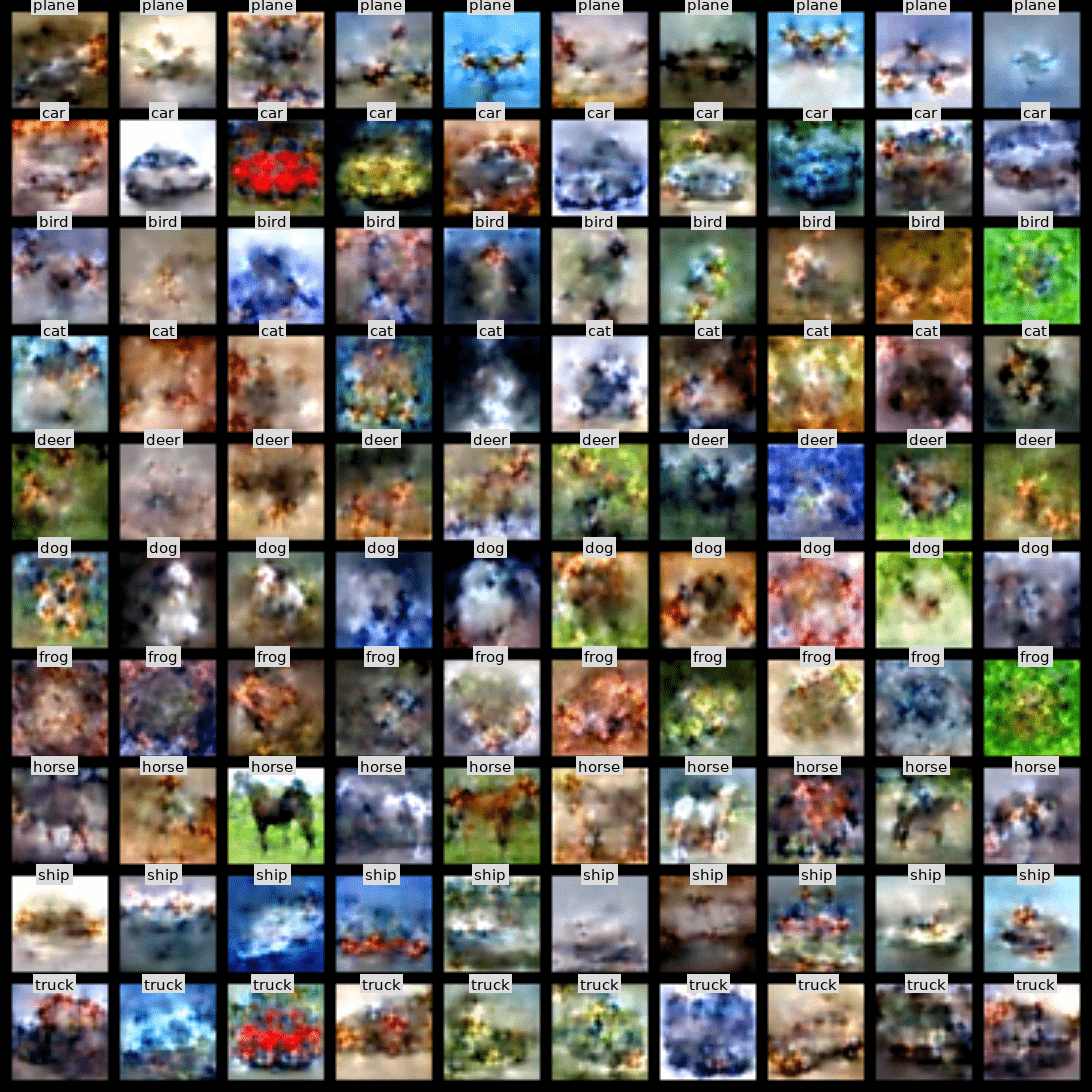}
    \caption{DataDAM}
    \label{fig:sub3}
  \end{subfigure}\hfill
  \caption{Learned synthetic images with different loss functions on the CIFAR10 dataset with IPC 10.}
  \label{fig:lossvis}
\end{figure*}

\textbf{Visualization of the synthetic images trained with different loss components in DataDAM.}
This section involves a comparison of the synthetic images generated by utilizing different loss objectives, namely only $\mathcal{L}_{\text{MMD}}$, only $\mathcal{L}_{\text{SAM}}$, layer-wise feature map transfer loss, and the DataDAM loss. The CIFAR10 dataset with IPC 10 was used for this evaluation to qualitatively assess the contribution of each loss component. As shown in Figure \ref{fig:lossvis}, the visualization of DataDAM is a linear combination of the $\mathcal{L}_{\text{SAM}}$ and $\mathcal{L}_{\text{MMD}}$ visualizations, resulting in a brighter and more contrasted image compared to each loss component individually. The generated synthetic sets by $\mathcal{L}_{\text{SAM}}$ and layer-wise feature transfer loss are somewhat similar since both losses match the information of feature maps generated by the real and synthetic datasets. However, the images distilled by $\mathcal{L}_{\text{SAM}}$ are brighter and more contrasted due to the matching of the most discriminative parts of the images.
\begin{figure*}
  \centering
   \begin{subfigure}{0.509\textwidth}
    \includegraphics[width=\linewidth]{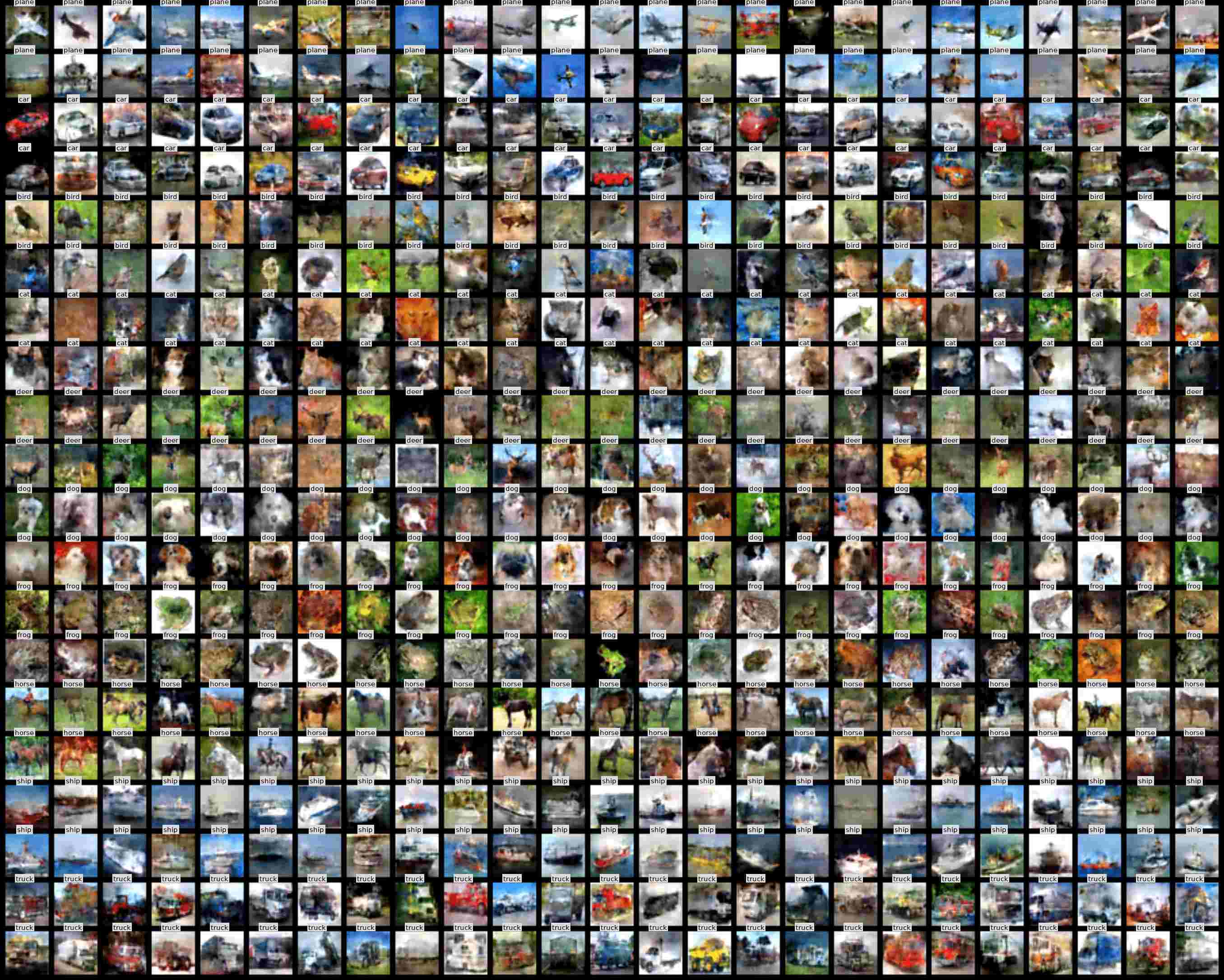}
    \vspace{-17pt}
    \caption{ConvNet}
    \label{fig:sub4}
  \end{subfigure}
  \begin{subfigure}{0.509\textwidth}
    \includegraphics[width=\linewidth]{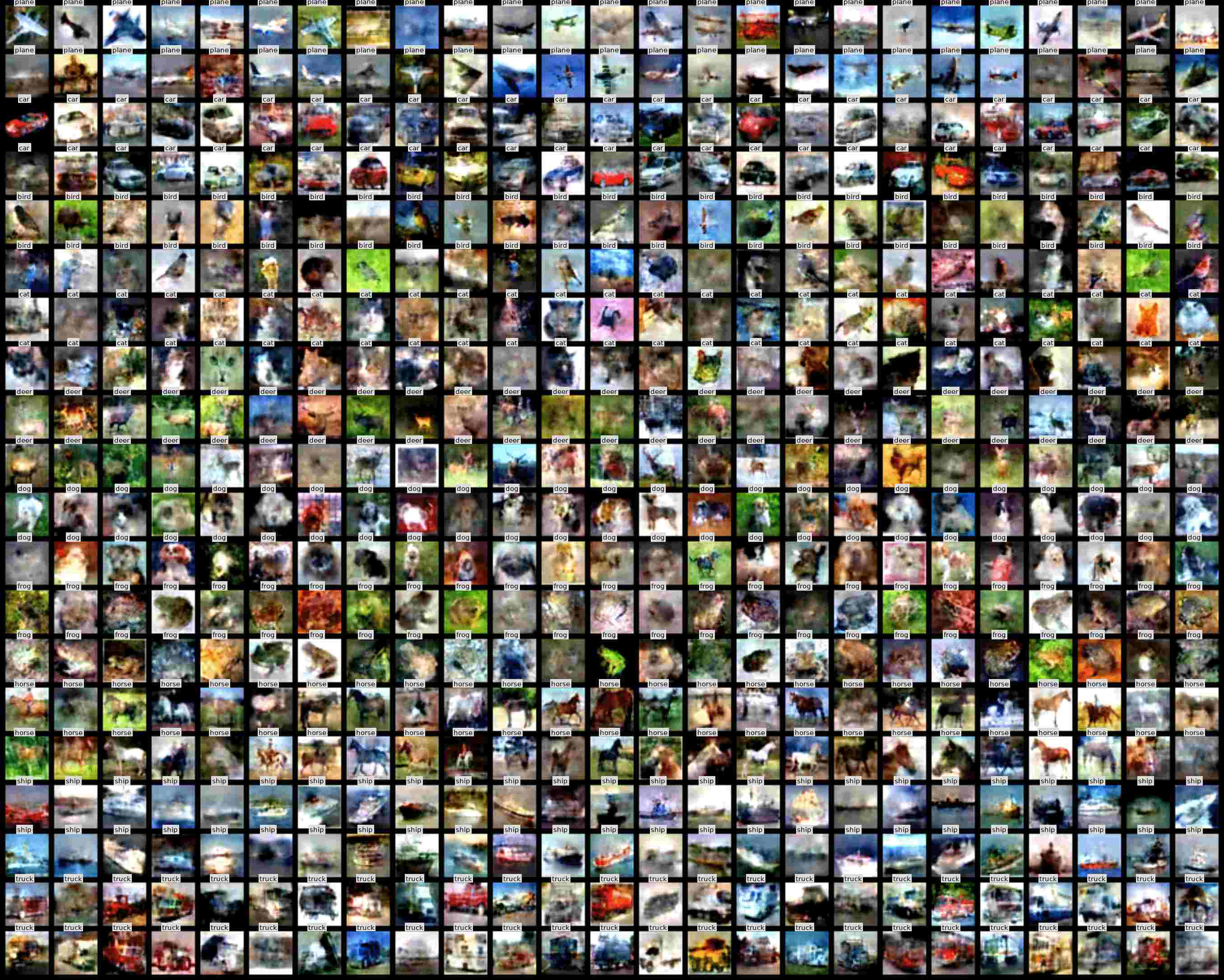}
    \vspace{-17pt}
    \caption{AlexNet}
    \label{fig:sub1}
  \end{subfigure}\hfill
  \begin{subfigure}{0.509\textwidth}
    \includegraphics[width=\linewidth]{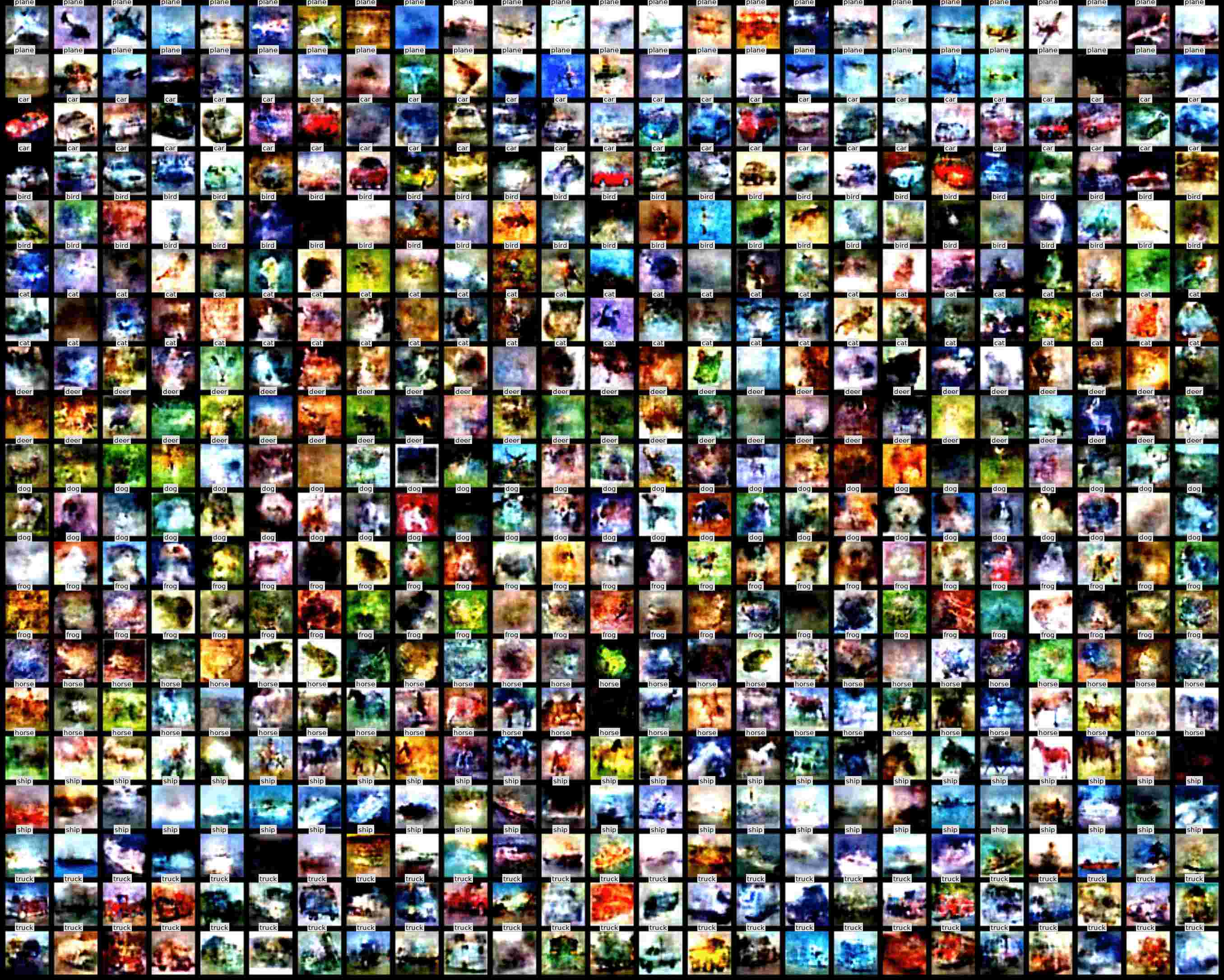}
    \vspace{-17pt}
    \caption{VGG-11}
    \label{fig:sub2}
  \end{subfigure}\vspace{-8pt} 
  \caption{Learned synthetic images with different model architectures on the CIFAR10 dataset with IPC 50.}
  \label{fig:synimages}
\end{figure*}

\begin{figure*}[htbp]
  \centering
   \begin{subfigure}[b]{0.33\textwidth}
    \includegraphics[width=\linewidth]{SupFigures/MMD_Only.jpg}
    \caption{Last layer ($\mathcal{L}_{\text{MMD}}$)}
    \label{fig:sub4}
  \end{subfigure}\hfill
  \begin{subfigure}[b]{0.33\textwidth}
    \includegraphics[width=\linewidth]{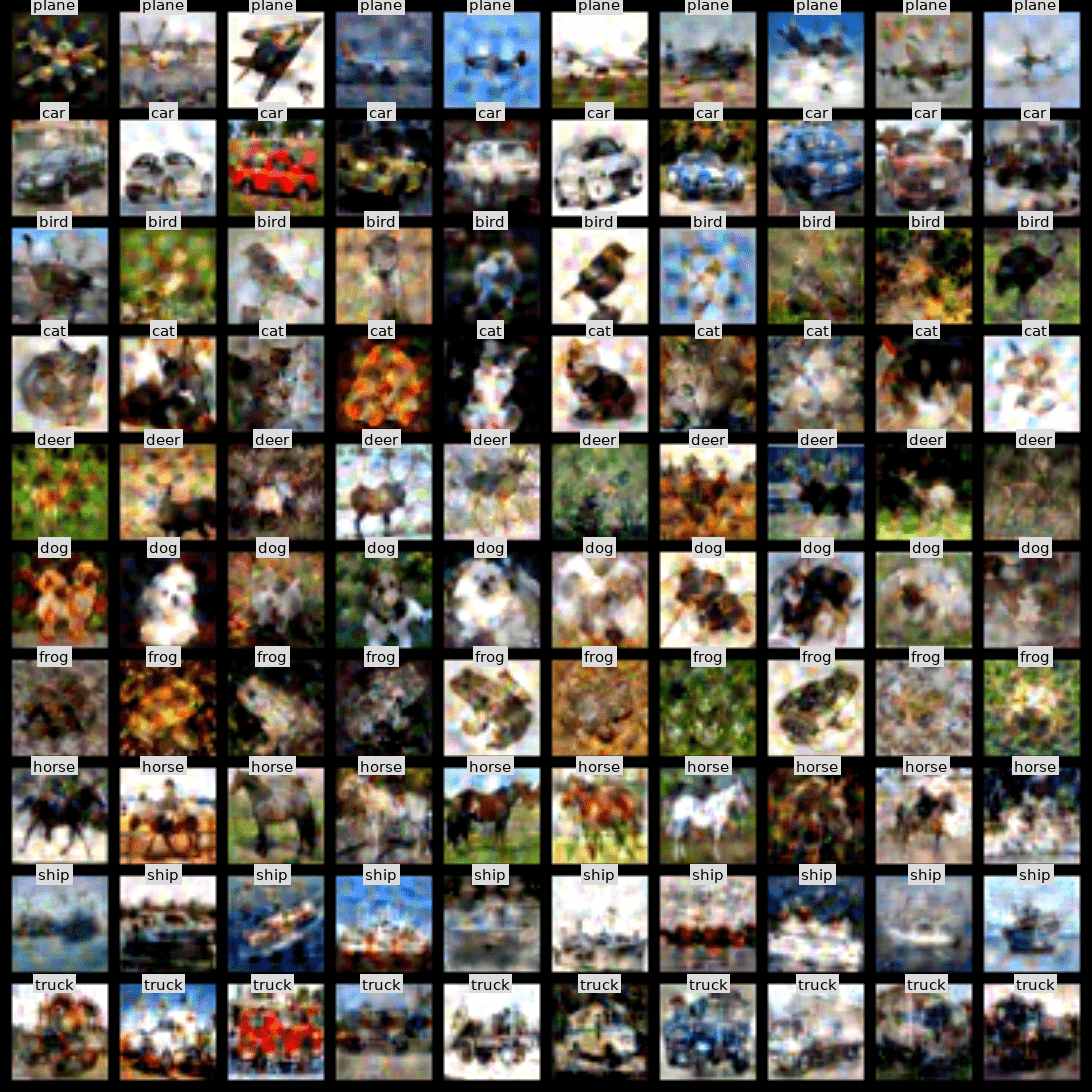}
    \caption{Layer 1 and Last Layer}
    \label{fig:sub1}
  \end{subfigure}\hfill
  \begin{subfigure}[b]{0.33\textwidth}
    \includegraphics[width=\linewidth]{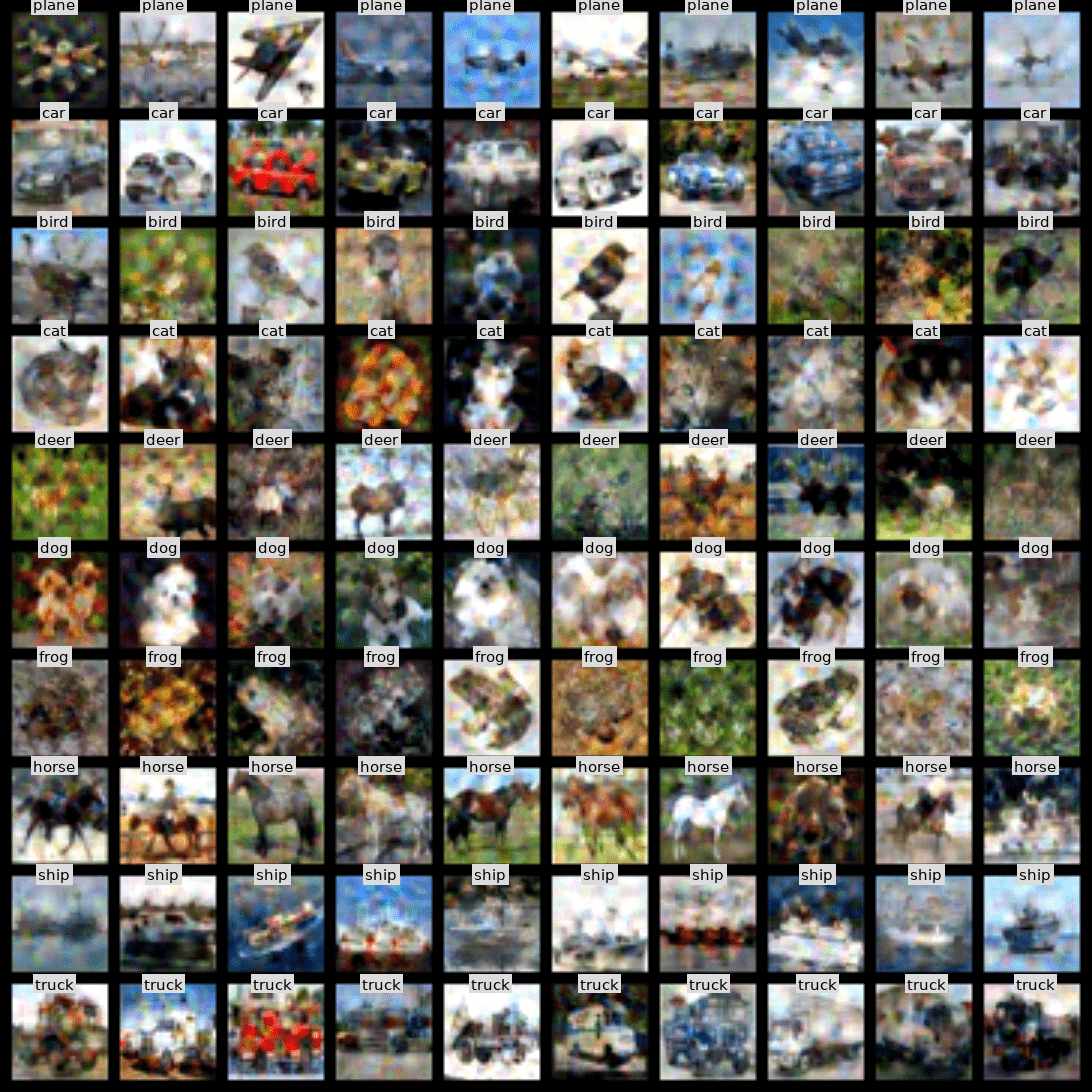}
    \caption{Layer 2 and Last Layer}
    \label{fig:sub2}
  \end{subfigure}\hfill
  \centering
  \begin{subfigure}[b]{0.33\textwidth}
    \includegraphics[width=\linewidth]{SupFigures/LSAM.jpg}
    \caption{Layer 1 and Layer 2 ($\mathcal{L}_{\text{SAM}}$)}
    \label{fig:sub3}
  \end{subfigure}
    \begin{subfigure}[b]{0.33\textwidth}
    \includegraphics[width=\linewidth]{SupFigures/FinalCIFAR10Grid.jpg}
    \caption{All layers (DataDAM)}
    \label{fig:sub3}
  \end{subfigure}\hfill
  \caption{Learned synthetic images with different matching layers on the CIFAR10 dataset with IPC 10.} 
  \label{fig:layervis}
\end{figure*}

\textbf{Visualization of the synthetic images trained with different layers in DataDAM.}
We conducted an experiment to analyze the distilled images produced by matching different layers of the ConvNet on real and synthetic datasets. Our study focused specifically on the CIFAR10 dataset with IPC 10. Figure \ref{fig:layervis} demonstrates that the layers performed differently as each layer conveyed distinct information regarding the data distributions. Our approach, DataDAM, utilizes all intermediate and final layers, resulting in distilled images that possess greater brightness and contrast. This is primarily due to the matching of attention maps in each layer as well as the embedding representation of the final layer.

\begin{figure*}
  \centering
   \begin{subfigure}{0.509\textwidth}
    \includegraphics[width=\linewidth]{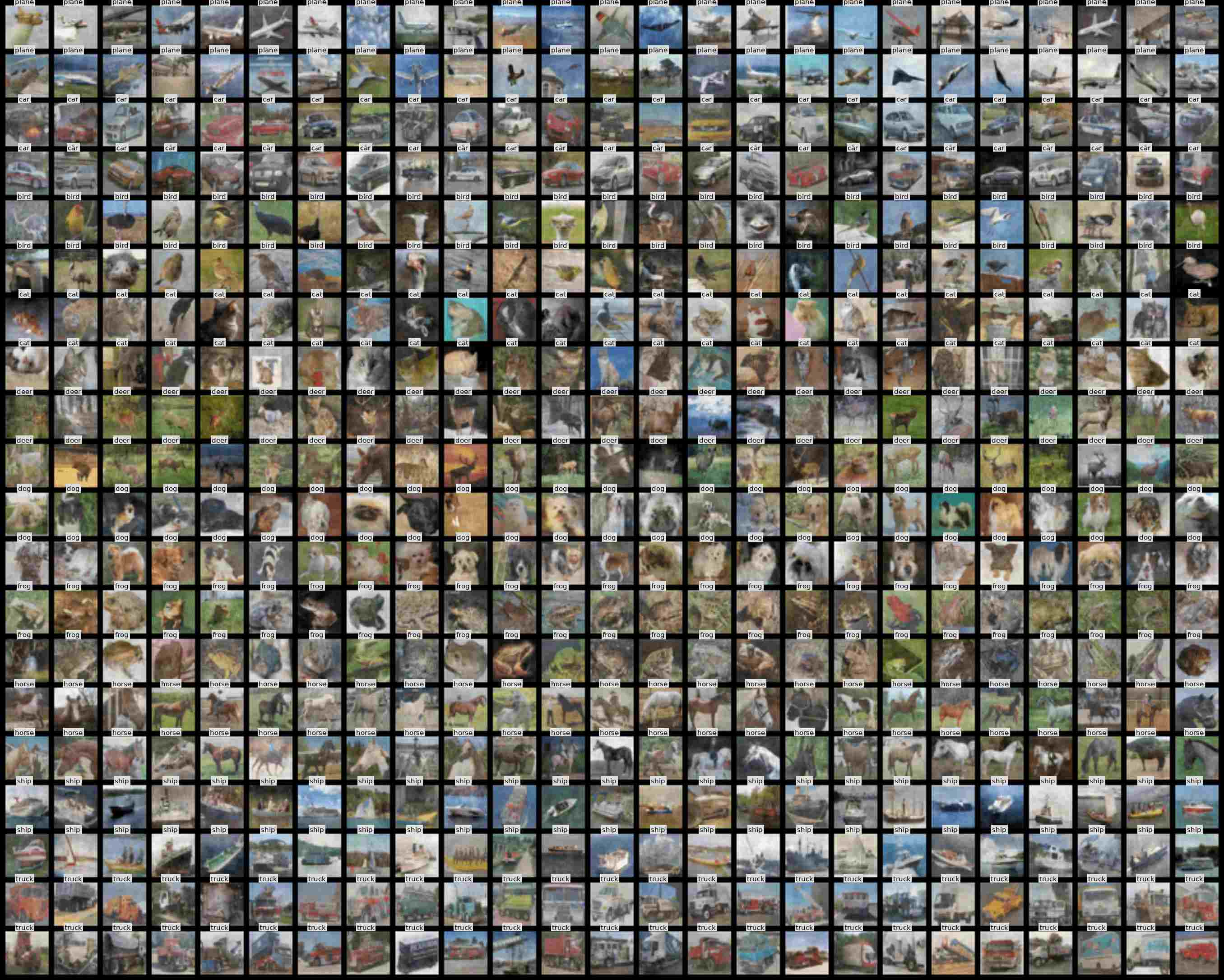}
    \vspace{-17pt}
    \caption{Random Initialization}
    \label{fig:sub4}
  \end{subfigure}
  \begin{subfigure}{0.509\textwidth}
    \includegraphics[width=\linewidth]{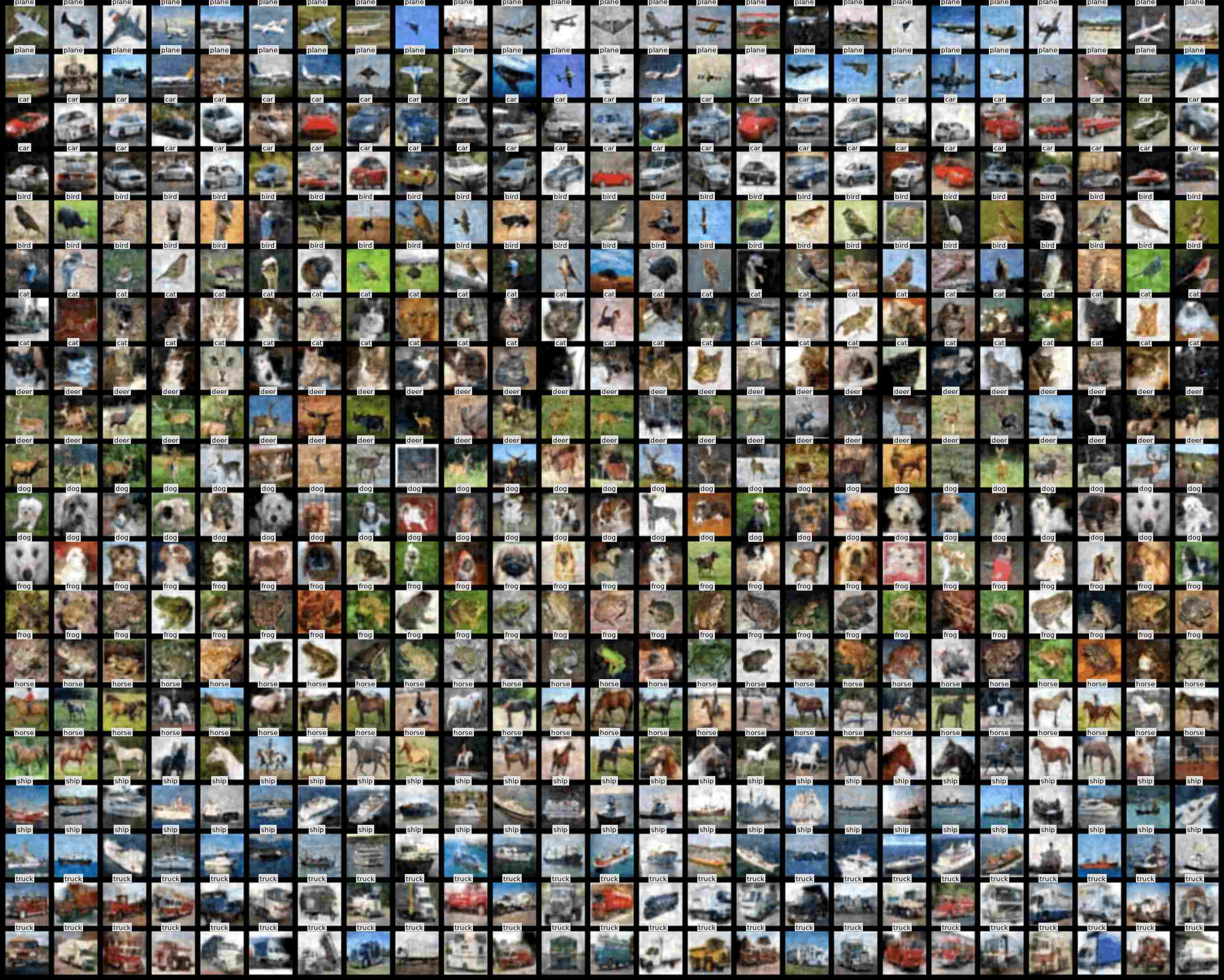}
    \vspace{-17pt}
    \caption{K-Center Initialization}
    \label{fig:sub1}
  \end{subfigure}\hfill
  \begin{subfigure}{0.509\textwidth}
    \includegraphics[width=\linewidth]{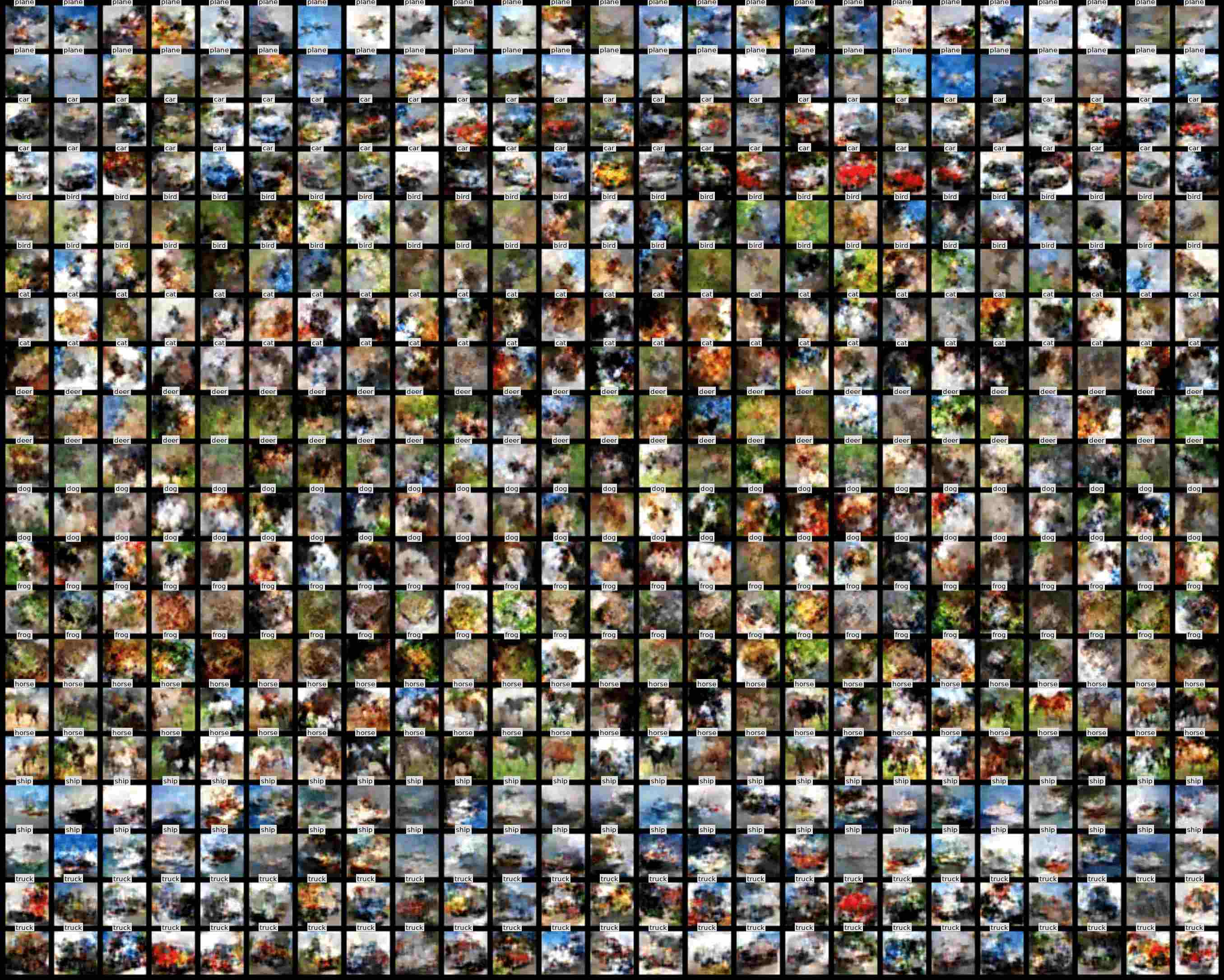}
    \vspace{-17pt}
    \caption{Gaussian noise Initialization}
    \label{fig:sub2}
  \end{subfigure}\vspace{-8pt} 
  \caption{Learned synthetic images with different initialization strategies on the CIFAR10 dataset with IPC 50.}
  \label{fig:visinit}
\end{figure*}

\textbf{Visualization of the synthetic images trained with different initialization strategies in DataDAM.} 
In this section, we presented the distilled images for the CIFAR10 dataset generated by IPC 50 using three distinct initialization methods: Random, K-Center \cite{seneractive, cuidc}, and Gaussian noise. Figure \ref{fig:visinit} illustrates the learned representations of the synthesis images produced using each initialization strategy. We observed a striking resemblance between the distilled images obtained through Random and K-Center initialization, which further confirms the results presented in the main paper. In contrast, the images generated using Gaussian noise initialization have noticeable differences in comparison to others, but they have still been learned effectively, and they contain crucial information for each class. In summary, these qualitative observations provide additional evidence that our model is robust enough to handle variations in initialization conditions.

\begin{figure*}
    \centering
    \includegraphics[width=\textwidth]{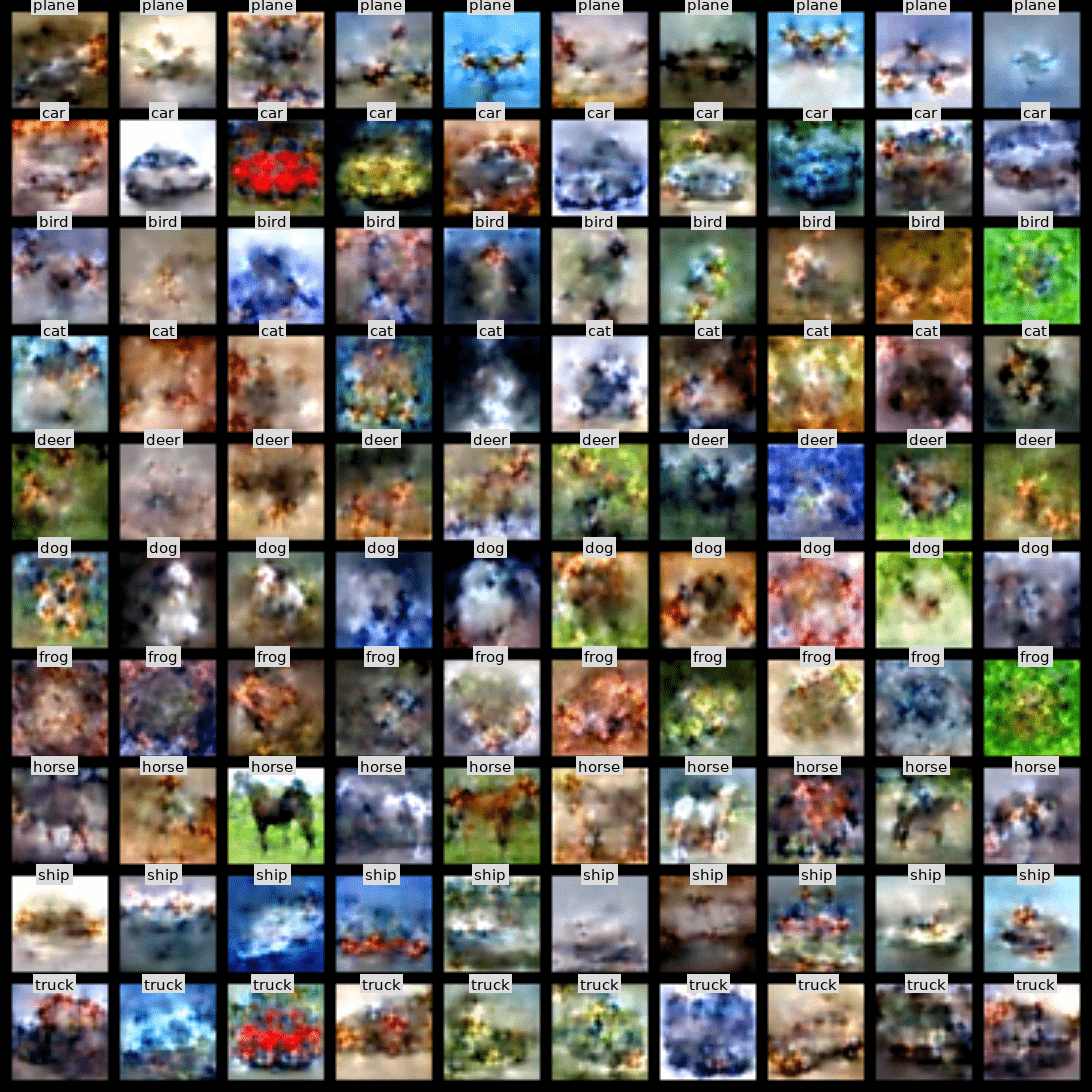}
    \caption{Distilled Image Visualization: CIFAR10 dataset with IPC 10.}
    \label{fig:cifar10ipc10}
\end{figure*}

\begin{figure*}
    \centering
    \includegraphics[width=\textwidth]{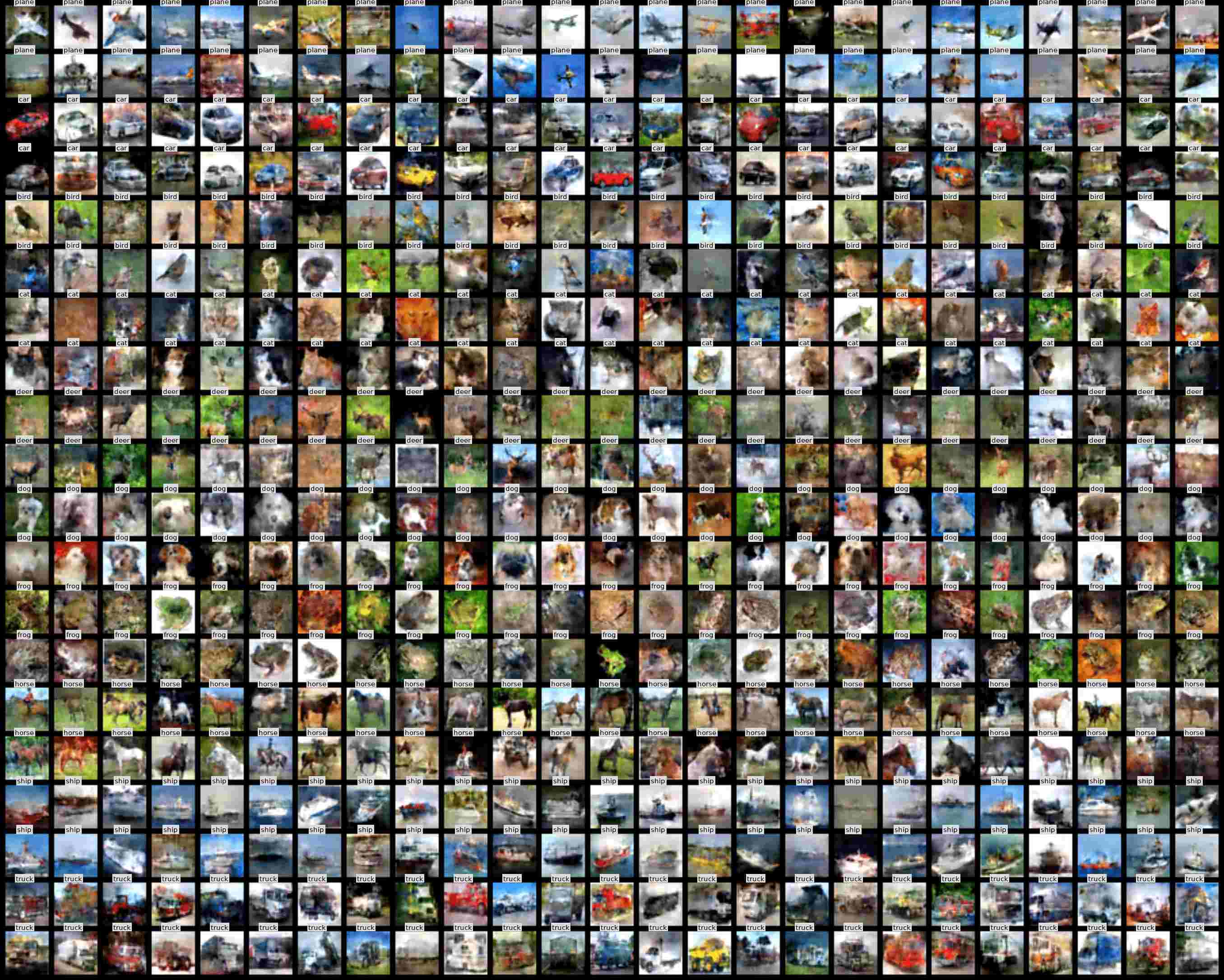}
    \caption{Distilled Image Visualization: CIFAR10 dataset with IPC 50.}
    \label{fig:cifar10ipc50}
\end{figure*}

\begin{figure*}
    \centering
    \includegraphics[width=\textwidth]{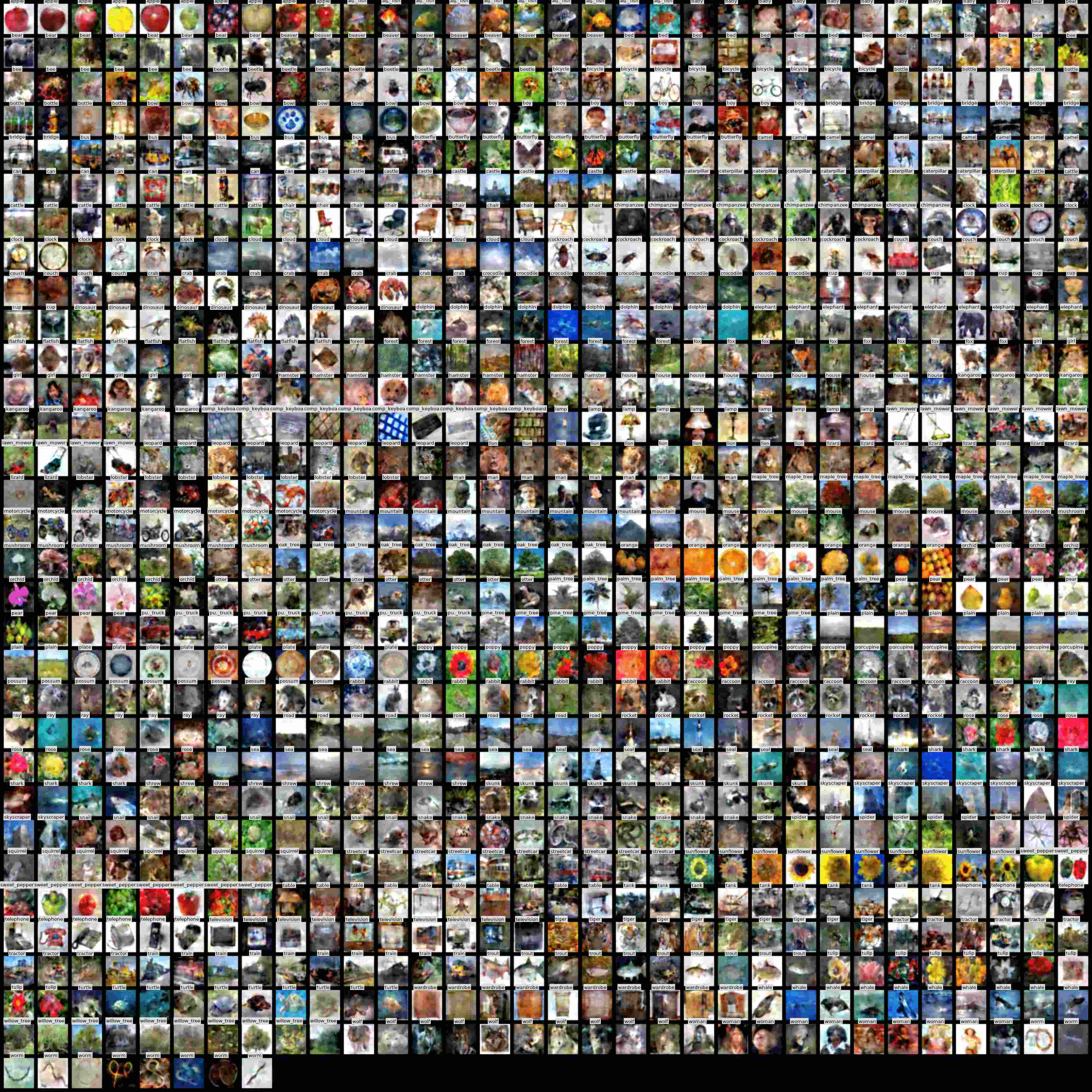}
    \caption{Distilled Image Visualization: CIFAR100 dataset with IPC 10.}
    \label{fig:cifar100ipc10}
\end{figure*}

\begin{figure*}
    \centering
    \includegraphics[width=\textwidth]{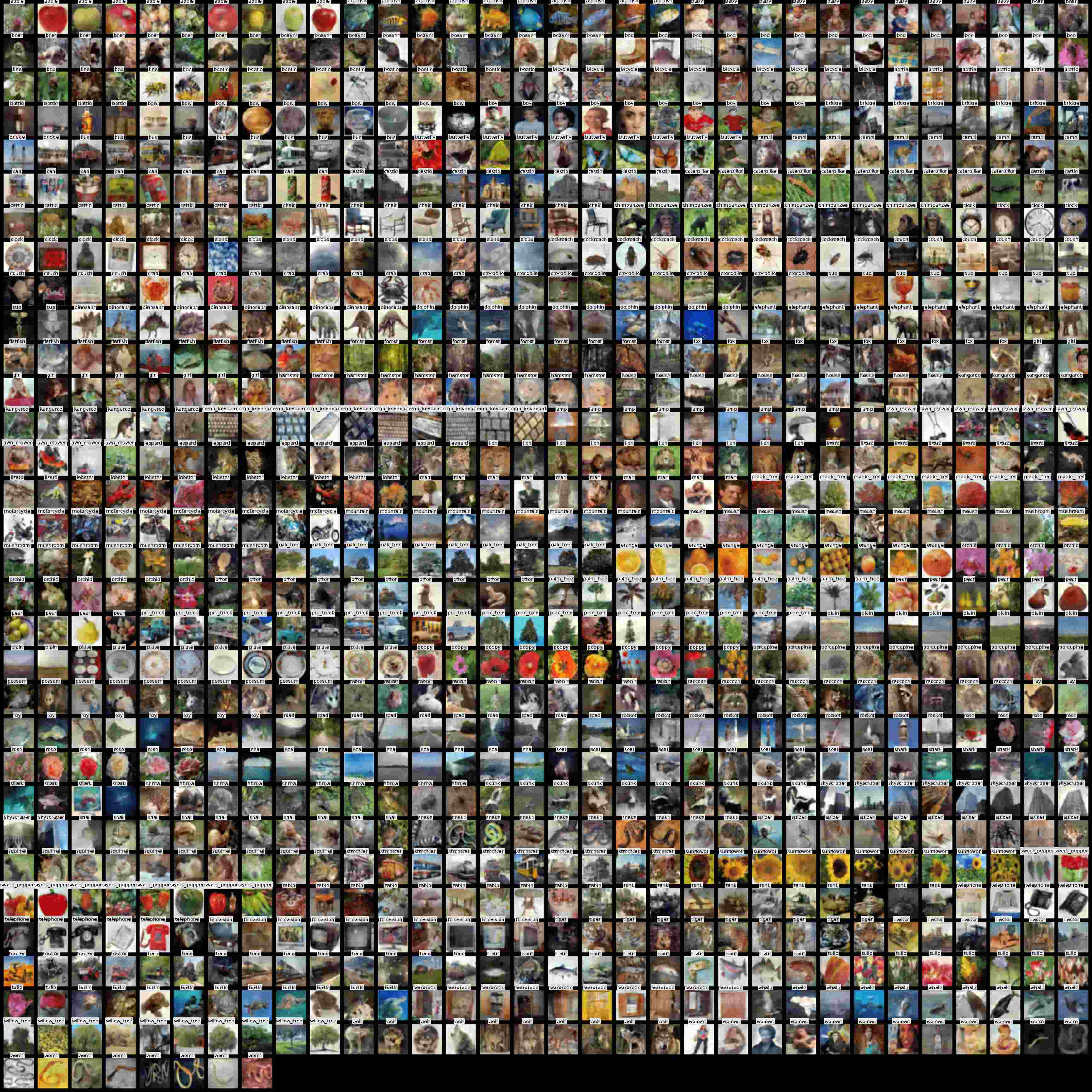}
    \caption{Distilled Image Visualization: CIFAR100 dataset with IPC 50 (10 randomly selected images for each class).}
    \label{fig:cifar100ipc50}
\end{figure*}

\begin{figure*}
    \centering
    \includegraphics[width=\textwidth]{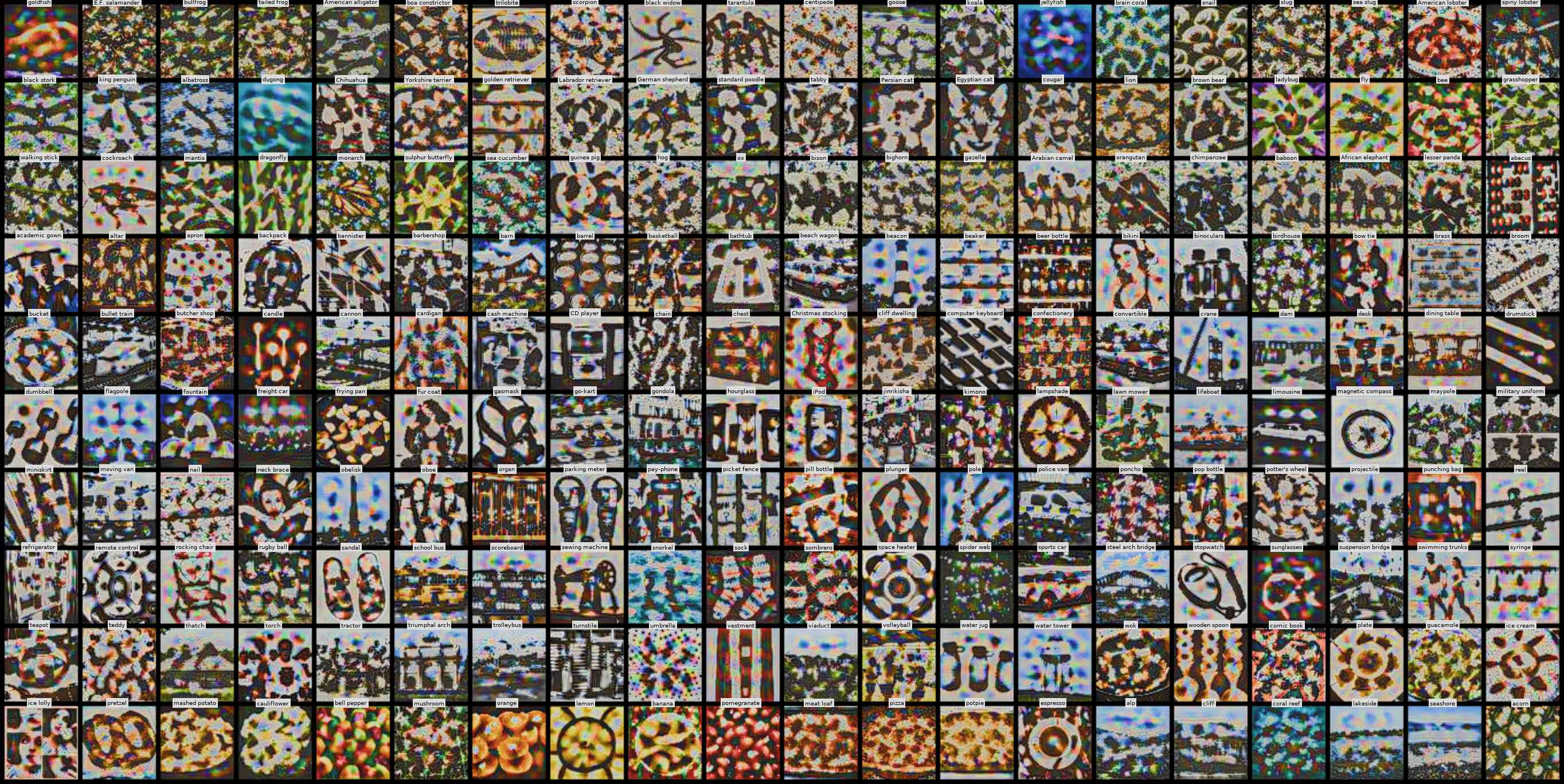}
    \caption{Distilled Image Visualization: TinyImageNet dataset with IPC 1.}
    \label{fig:tinyimagenetipc1}
\end{figure*}

\begin{figure*}
    \centering
    \includegraphics[width=\textwidth]{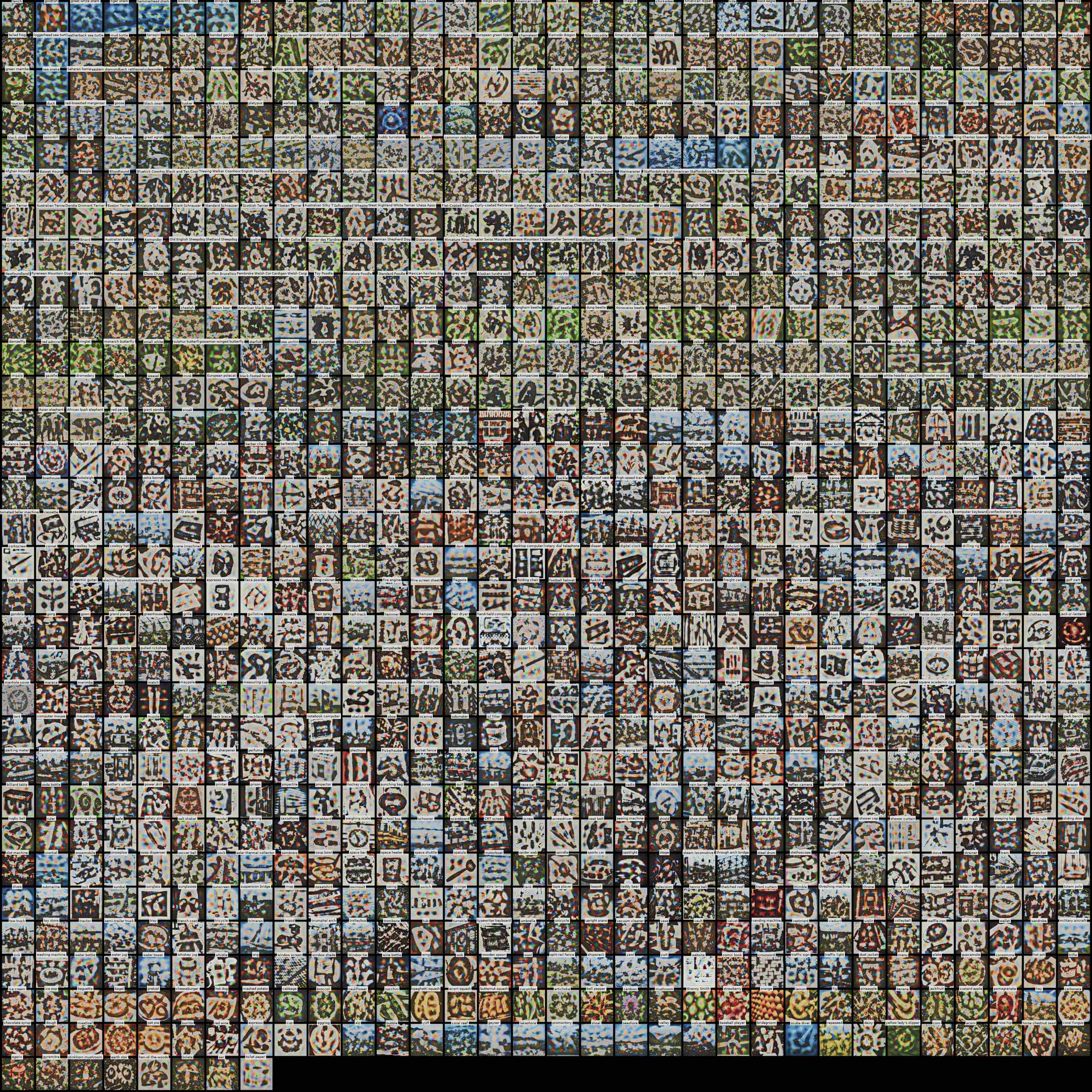}
    \caption{Distilled Image Visualization: ImageNet-1K dataset with IPC 1.}
    \label{fig:imagenetipc1}
\end{figure*}

\begin{figure*}
    \centering
    \includegraphics[width=\textwidth]{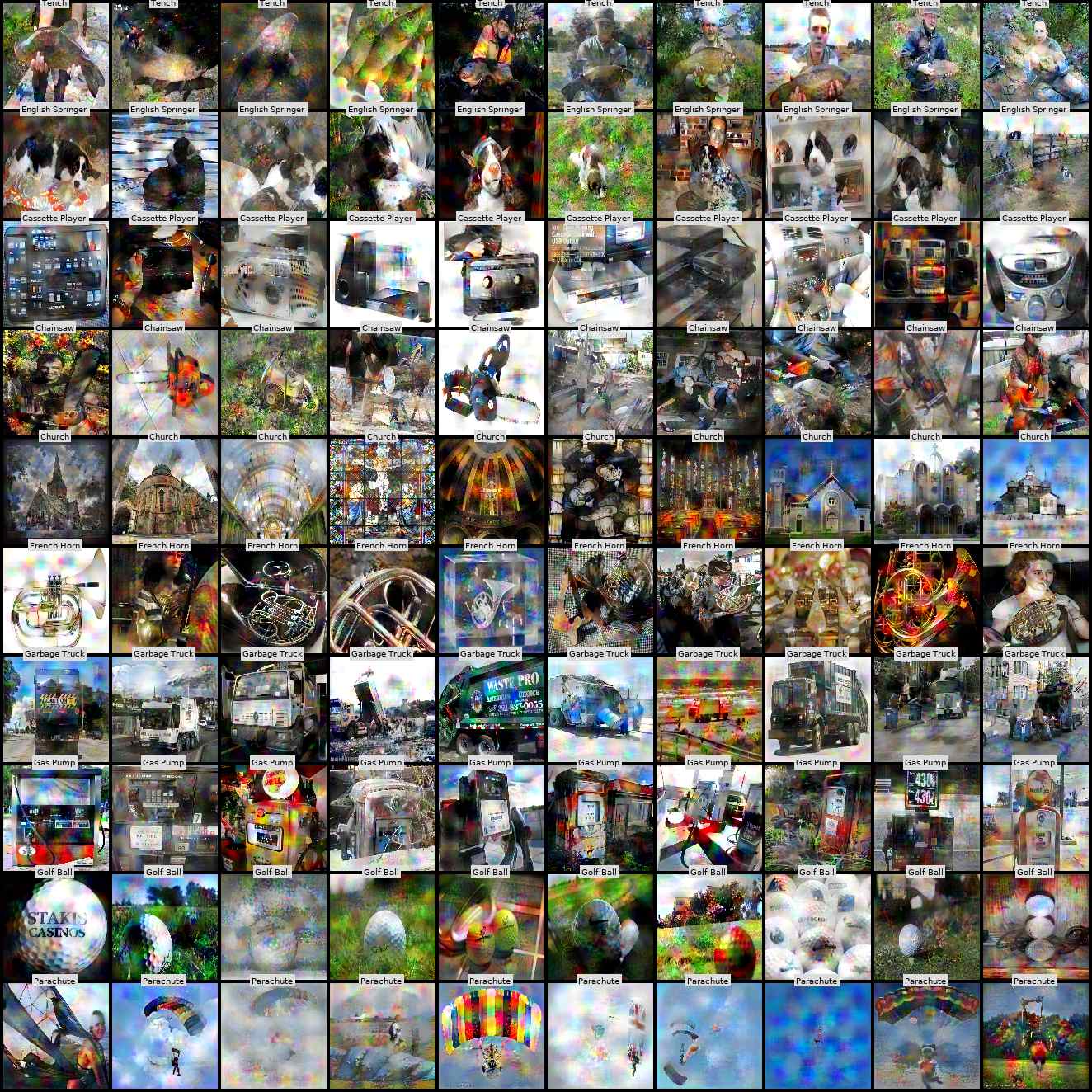}
    \caption{Distilled Image Visualization: ImageNette dataset with IPC 10.}
    \label{fig:imagenetteipc10}
\end{figure*}

\begin{figure*}
    \centering
    \includegraphics[width=\textwidth]{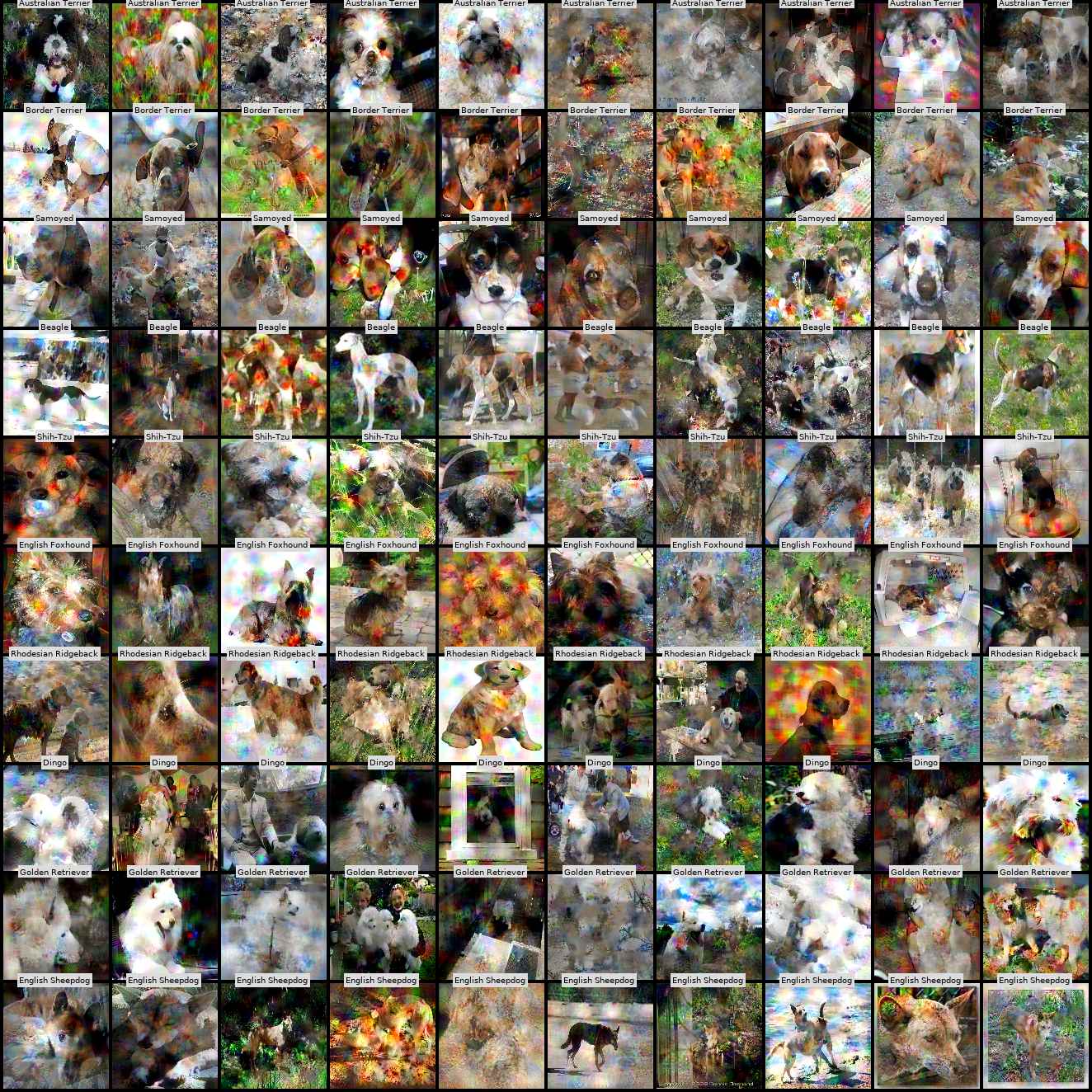}
    \caption{Distilled Image Visualization: ImageWoof dataset with IPC 10.}
    \label{fig:imagenetwoofipc10}
\end{figure*}

\begin{figure*}
    \centering
    \includegraphics[width=\textwidth]{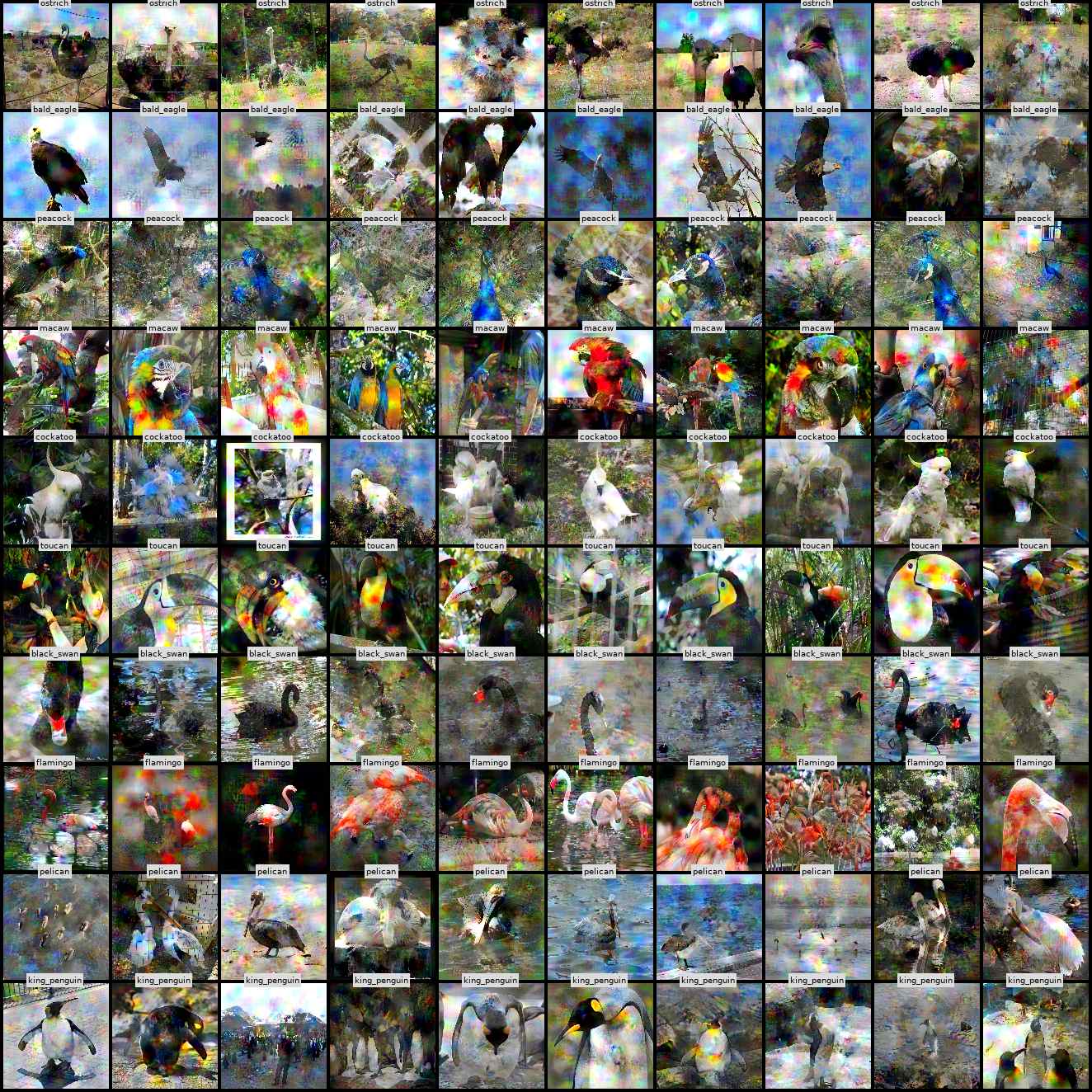}
    \caption{Distilled Image Visualization: ImageSquawk dataset with IPC 10.}
    \label{fig:imagenetsquackipc10}
\end{figure*}

\textbf{More distilled image visualization.}
We provide additional visualizations of the distilled images for all five datasets used in this work, namely CIFAR10 (Figures \ref{fig:cifar10ipc10}, \ref{fig:cifar10ipc50}), CIFAR100 (Figures \ref{fig:cifar100ipc10}, \ref{fig:cifar100ipc50}), TinyImageNet (Figure \ref{fig:tinyimagenetipc1}), ImageNet-1K (Figure \ref{fig:imagenetipc1}), ImageNette (Figure \ref{fig:imagenetteipc10}), ImageWoof (Figure \ref{fig:imagenetwoofipc10}), and ImageSquawk (Figure \ref{fig:imagenetsquackipc10}).



\clearpage


\end{document}